\documentclass[11pt]{article}

\usepackage[final]{acl}

\usepackage{times}
\usepackage{latexsym}

\usepackage[T1]{fontenc}

\usepackage[utf8]{inputenc}

\usepackage{microtype}

\usepackage{inconsolata}

\usepackage{graphicx}

\usepackage{amsmath}
\usepackage{amssymb}
\usepackage{mathtools}
\usepackage{amsthm}
\usepackage{tabularx}
\usepackage{graphicx}
\usepackage{multirow}
\usepackage{tcolorbox}
\usepackage{enumitem}
\usepackage{xurl} 
\usepackage{subcaption}
\usepackage{cleveref}
\usepackage{algorithm}
\usepackage{algpseudocode}
\usepackage{wrapfig}
\usepackage{booktabs}
\usepackage{xspace}

\newcommand{\saga}{\textbf{S}cene-graph-\textbf{A}ware \textbf{G}uided \textbf{A}ttention (\textsc{saga})}
\newcommand{\sagashort}{\textsc{SAGA}\xspace}
\newcommand{\dataset}{\textsc{LabHazVision}\xspace}

\AtBeginDocument{
  \setlength{\abovedisplayskip}{3pt}
  \setlength{\belowdisplayskip}{3pt}
  \setlength{\abovedisplayshortskip}{1pt}
  \setlength{\belowdisplayshortskip}{1pt}
}

\usepackage[table]{xcolor}
\usepackage{colortbl}

\definecolor{f00}{RGB}{240,255,240}
\definecolor{f01}{RGB}{200,255,200}
\definecolor{f02}{RGB}{150,255, 150}
\definecolor{f04}{RGB}{50,255,50}
\definecolor{f06}{RGB}{0,255,0}
\definecolor{f08}{RGB}{0,220,0}
\definecolor{f10}{RGB}{0,150,0}

\newcommand{\cc}[1]{%
    \ifdim#1pt<0pt\cellcolor{f00}#1%
    \else\ifdim#1pt<0.25pt\cellcolor{f01}#1%
    \else\ifdim#1pt<0.65pt\cellcolor{f02}#1%
    \else\ifdim#1pt<0.70pt\cellcolor{f04}#1%
    \else\ifdim#1pt<0.80pt\cellcolor{f06}#1%
    \else\ifdim#1pt<0.90pt\cellcolor{f08}#1%
    \else\cellcolor{f10}#1%
    \fi\fi\fi\fi\fi%
}

\title{Toward Autonomous Laboratory Safety Monitoring with Vision–Language Models:
Learning to See Hazards Through Scene Structures}

\author{
 \textbf{Trishna Chakraborty\textsuperscript{1}},
 \textbf{Udita Ghosh \textsuperscript{1}},
 \textbf{Aldair Ernesto Gongora\textsuperscript{2}},
 \textbf{Ruben Glatt\textsuperscript{1}},
\\
 \textbf{Yue Dong\textsuperscript{1}},
 \textbf{Jiachen Li\textsuperscript{1}},
 \textbf{Amit K. Roy-Chowdhury\textsuperscript{1}},
 \textbf{Chengyu Song \textsuperscript{1}}
\\
\\
 \textsuperscript{1}University of California, Riverside,
 \textsuperscript{2}Lawrence Livermore National Laboratory
\\
}

\begin{document}
\maketitle
\begin{abstract}
Laboratories are prone to severe injuries from unsafe actions, yet continuous manual safety monitoring, beyond mandatory pre-lab safety training, is infeasible. Vision-language models (VLMs) offer promise for autonomous laboratory safety monitoring, but their effectiveness in realistic settings is unclear due to the lack of visual evaluation data. Safety incidents are mostly documented as text, and existing studies are largely limited to textual understanding. Since hazardous scenes are unsafe to stage or capture,
synthetic visual data becomes the only practical way to study this problem.
This motivates us to introduce a structured data generation pipeline that converts textual laboratory scenarios into aligned quadruples using LLM as scene graph architect and image generation model as renderer, yielding \dataset: a synthetic dataset of 1,207 $\langle$image, scene graph, ground truth, hazard relevance mask$\rangle$ quadruples across 362 unique scenarios.
Experiments on \dataset and seven open- and closed-source models show that VLMs perform effectively given a textual scene graph but degrade substantially in visual-only settings (i.e., where no scene graph 
is involved), indicating difficulty in extracting structured object relationships directly from pixels. We thus propose \saga, a post-training framework that leverages reinforcement learning to first structure the visual scene as a scene graph, and then predict hazards while aligning attention with relevant image regions.
Evaluations with Qwen 3.5 and Gemma 4 show that our method nearly doubles the $F_1$ score over the baseline on our synthetic dataset and achieves an improvement of $\sim 3\times$ on a mini real-world dataset, while preserving general visual understanding in visual question answering.



\end{abstract}

\section{Introduction}\label{sec:intro}

Laboratories are central to scientific innovation but inherently involve safety risks that can result in severe injuries, loss of life, or damage to costly instrumentation when unsafe actions occur~\citep{hazardous_issues, china_lab_issues}. While pre-lab safety training is essential~\citep{academic_lab_safety_research, laboratory_challenges}, it is insufficient on its own and can be strengthened through continuous laboratory safety monitoring. However, such monitoring remains critically under-supported, as traditional manual inspections are constrained by human availability.

Large language models (LLMs) encode broad world knowledge and scientific reasoning~\citep{scientific_knowledge, scientic_concepts}, and vision-language models (VLMs), which pair LLMs with visual perception modules~\citep{vlms, lvlm}, ground this reasoning in visual observations. VLMs are thus a promising foundation for continuous camera-based laboratory monitoring, identifying hazardous configurations such as incompatible chemical storage or unsafe equipment setups, and flagging risks with mitigation recommendations. 

VLMs have been evaluated on scientific knowledge~\citep{olympiadVQA, scientificVQA}, and LabSafetyBench~\citep{lab_safety_bench} extends this to laboratory settings. However, with a limited set of chatbot-style VQA examples, these images are far from realistic lab environments and mainly test recognition of lab signs.The corresponding questions direct the model where to look, whereas real-world monitoring requires open-ended observation of complex scenes. Therefore, autonomous laboratory safety monitoring remains largely unexplored.

This gap motivates our first research question:
\textit{\textbf{RQ1 (Benchmarking)}: How well do VLMs identify hazards present in an image?} Addressing it requires image datasets with reliable hazard annotations. Unfortunately, most laboratory safety knowledge exists as unstructured text (e.g., guidelines, incident reports), and real hazardous scenes are unsafe to stage or capture, making synthetic visual data a practical way to study this problem. 

Text-to-image models~\citep{stable_diffusion, text_to_image_benchmarking} make synthesizing images from text an appealing option~\citep{diffusion_data_augmentation, synthetic_images}, but our preliminary exploration shows that unconstrained generation mostly fails to preserve the precise object states and spatial relationships that define laboratory setups. We therefore propose a data generation pipeline with Visual-Genome–style scene graph (SG)~\citep{visual_genome} that explicitly encodes objects and relationships, a formulation widely adopted for scene understanding~\citep{scene_graph_to_image, vqa_over_scene_graph}. Using a strong LLM (GPT-5~\citep{openai_gpt5, gpt5}) with constrained prompting~\citep{prompting_guideline} and chain-of-thought reasoning, we first translate textual scenarios into scene graphs, which then condition a state-of-the-art text-to-image model (Nano-Banana-pro~\citep{nano_banana_pro}) to generate photorealistic laboratory scenes.
This structured generation decouples responsibilities: the Architect (LLM) enforces logical consistency, while the Renderer (image generator) focuses on visual realism~\citep{visual_architect}.
Leveraging this pipeline, we construct \dataset: 1,207 aligned ⟨image, scene graph, ground truth, hazard relevance mask⟩ quadruples from LabSafetyBench scenarios~\citep{lab_safety_bench}, verified by humans and VLM-as-judge~\citep{llm-as-judge}.

\begin{figure}
    \centering  
    \includegraphics[width = \linewidth]{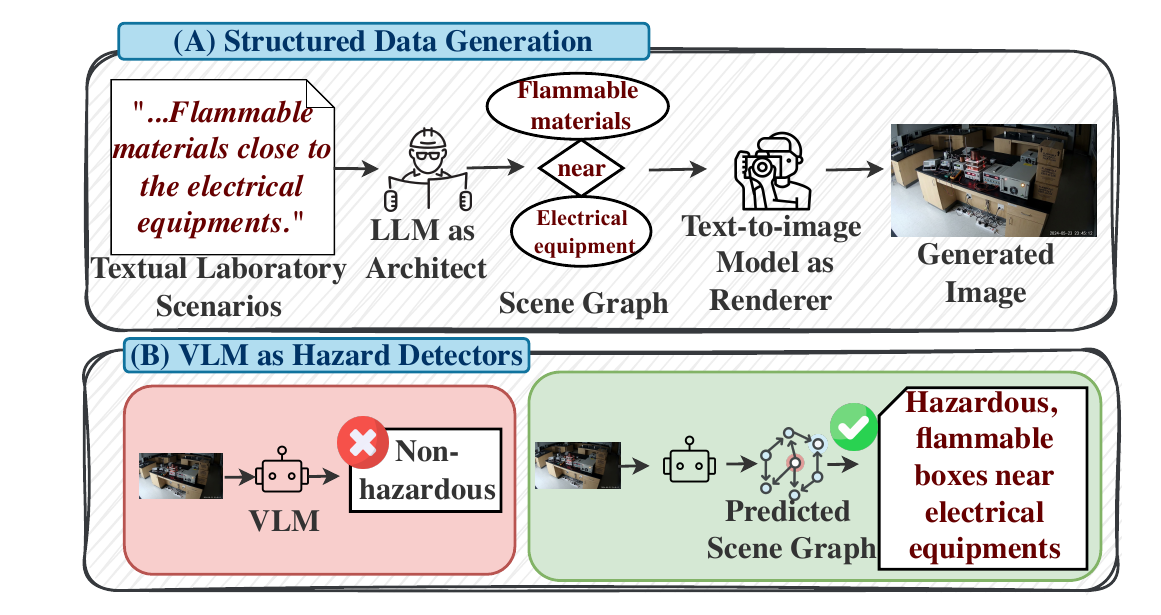}
    \vspace{-1.5em}
    \caption{Overview of our approach. \textit{(A)} An LLM \emph{Architect} translates textual laboratory scenarios into scene graphs, which condition an image-generation \emph{Renderer} to synthesize photorealistic lab images. \textit{(B)} VLMs often miss hazards from raw images alone, but inferring a scene graph from the image first enables safety reasoning and leads to correct hazard detection.}
    \label{fig:teaser}
     \vspace{-1.5em}
\end{figure}

In analyzing RQ1, benchmarking on \dataset reveals an intriguing and consistent pattern. When only images are provided, all models perform poorly. 
In contrast, when ground truth scene graphs are provided, all VLMs achieve better performance. This gap indicates difficulty in extracting structured object relationships directly from pixels. This  motivates our second research question: \textit{\textbf{RQ2 (SG-Guided Alignment)}: Can scene graphs be leveraged to improve VLM hazard detection in visual-only scenarios?} 
To answer it, 
we use Group Sequence Policy Optimization (GSPO)~\citep{GSPO}, a reinforcement learning method, as our training objective, and propose \saga. 
\sagashort trains the model to first predict a scene graph from the visual input before the hazard detection (\autoref{fig:teaser}), rewarding correct reasoning format and hazard classification, while an attention alignment loss~\citep{AGE-VLM} grounds the model's focus on relevant image regions.
Evaluations with Qwen 3.5 and Gemma 4 show that our method nearly doubles the $F_1$ score over the vision-only baseline on our synthetic dataset and achieves an improvement of $\sim 3\times$ on a mini real-world dataset, while preserving general visual understanding in visual question answering.

Overall, our contributions are as follows:
\begin{itemize}[itemsep=0pt, topsep=0pt]
\item We introduce a structured synthetic data generation framework for laboratory hazard detection, constituting \dataset: 1,207 $\langle$image, scene graph, ground truth, hazard relevance mask$\rangle$ quadruples across 362 unique scenarios, to our knowledge the first dataset for studying VLMs as laboratory hazard detectors and enabling open-ended evaluation of autonomous safety monitoring.
\item Evaluating seven
 open- and closed-source VLMs, we show that hazard detection performance degrades substantially in visual-only settings (i.e., where no scene graph 
is involved) compared to text-based scene graph reasoning, indicating that VLMs encode safety logic but struggle to extract structured object relationships from pixels.
\item We propose \saga, a post-training framework based on reinforcement learning that trains models to structure the visual scene as a scene graph before predicting hazards, while grounding attention in relevant image regions. With Qwen~3.5 and Gemma~4, \sagashort improves vision-only $F_1$ on both \dataset and a mini real-world dataset while maintaining VQA accuracy.
\end{itemize}

\vspace{-0.1cm}
\section{Related Work} \label{sec:related}
\vspace{-0.1cm}

\textbf{LLMs for Scientific Reasoning.}
Scientific reasoning has been studied for chemical knowledge~\citep{chembench}, broader scientific understanding and concept acquisition~\citep{scientific_knowledge, scientic_concepts}, laboratory protocols and errors~\citep{bio_protocol_llm, bio_pro_protocol_llm, lab_safety_bench}, but these efforts frame the task as chatbot-style question answering. Other works such as WMDP~\citep{wmdp_hazardous_qa} and SOSBench~\citep{scientific_hazard_bench} examines adversarial misuse of scientific knowledge in LLMs, including weaponization of chemical and biological expertise.

\noindent\textbf{Agentic Science.}
LLMs also drive agentic scientific systems, coupled with robotic platforms for reaction planning~\citep{chem_crow, llm_chemistry_robot} or agent-driven discovery~\citep{autonomous_chemical_research}. Frameworks such as k-agents~\citep{cao_2025}, Agent Laboratory~\citep{schmidgall2025agentlaboratoryusingllm}, and AILA~\citep{mandal2025evaluating} automate complex laboratory workflows, but largely overlook the safety safeguards essential in physical lab settings. 

\noindent\textbf{Scene Graphs in Agentic Systems.}
Agentic LLMs increasingly use scene graphs as structured, interpretable representations that ground reasoning about visual and spatial context~\citep{yin2024sg,yang2025llm}. ESCA~\citep{ecsa} grounds multimodal LLM agents in scene graphs to contextualize perception and guide high-level planning, while schema-guided frameworks~\citep{chen2025schema} use cooperative LLM agents over scene graphs to reduce hallucinations and improve planning in simulation.

\noindent\textbf{Vision-Based Safety.}
LabSafetyBench~\citep{lab_safety_bench} is the most closely related work, supplementing its textual scenarios with a small number of vision-based QA examples; however, its images do not reflect realistic laboratory setups. Other work applying VLMs to laboratory safety monitoring remains narrowly scoped~\citep{chemist_eye, Clip2Safety, yolo_hazard_detection}, typically targeting a single task such as personal protective equipment (PPE) compliance or fire hazard detection, and often evaluating only small-scale or task-specific models.

\begin{figure*}[t]
    \begin{center} \includegraphics[width=0.95\linewidth]{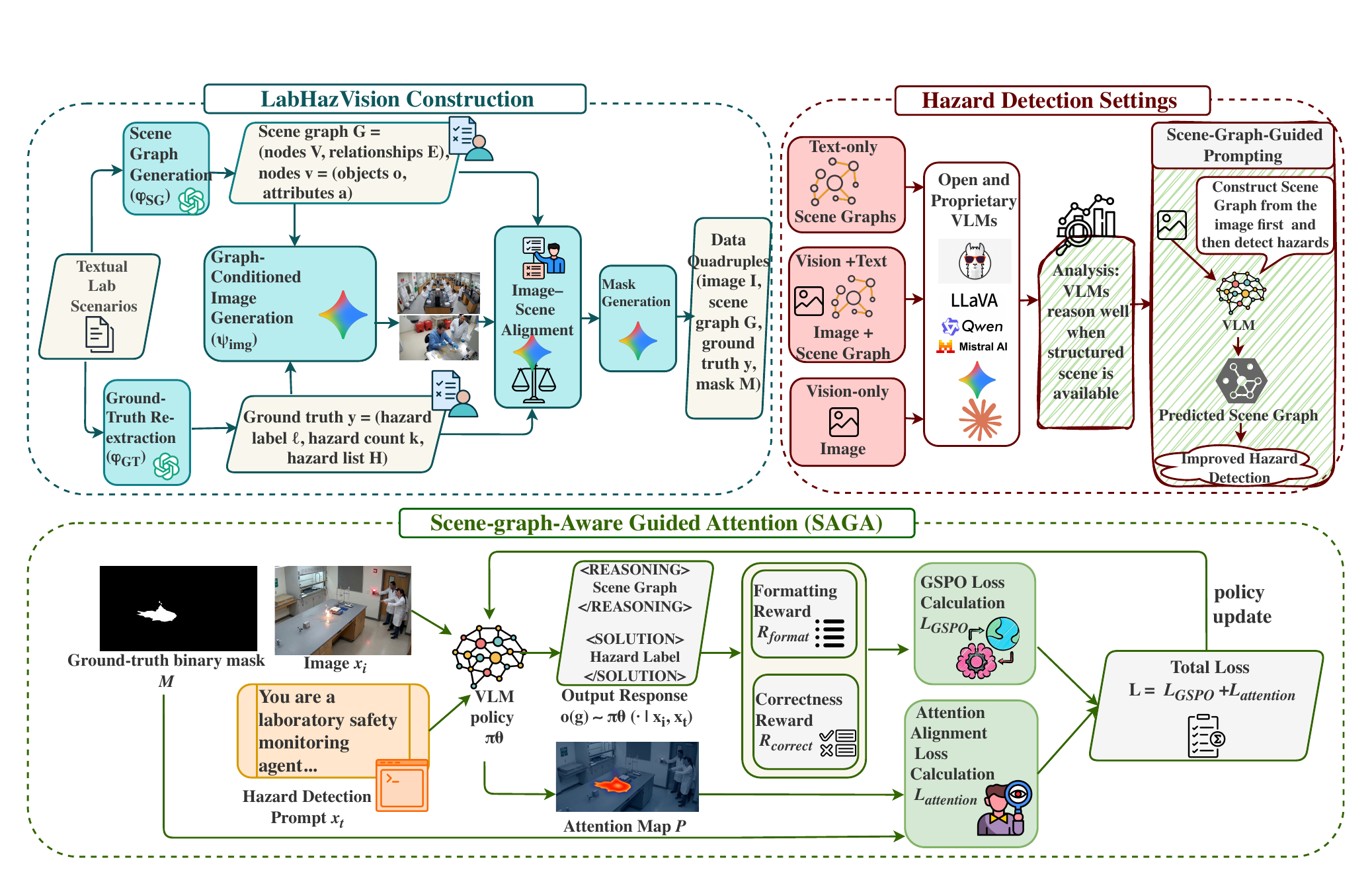}
    \end{center}
    \vspace{-0.5em}
    \caption{End-to-end pipeline for dataset construction, hazard detection settings, and model training. Textual laboratory scenarios are converted into scene graphs, hazard annotations, images, and hazard relevance masks, then verified through human and VLM-as-judge checks. VLMs are evaluated under text-only, vision+text, vision-only, and scene-graph-based prompting settings. We further improve visual-only hazard detection using \sagashort, which trains models to generate scene-graph reasoning while grounding attention in relevant image regions.}
    \label{fig:open_figure}
    \vspace{-1.2em}
\end{figure*}

\vspace{-0.2cm}
\section{\dataset Construction and Hazard Detection Settings} \label{sec:dataset_construction}
\vspace{-0.1cm}

To study whether VLMs can function as autonomous laboratory safety monitors, we construct a structured dataset consisting of quadruples $(I, \mathcal{G}, y, M)$, where $I$ denotes a laboratory image, $\mathcal{G}$ is its scene graph (Eq.~\ref{eq:sg_def}), $y$ is the ground-truth hazard annotation (Eq.~\ref{eq:gt_structured}), and $M$ is the hazard relevance mask. 
We build this dataset by leveraging textual laboratory scenarios with potential safety hazards drawn from the existing benchmark, LabSafetyBench~\citep{lab_safety_bench}. 
The full pipeline is detailed in \autoref{fig:open_figure}. The complete set of prompts is included in~\autoref{appendix:sec:prompts}.

\subsection{Structured Data Generation}\label{subsec:data_gen}

\textbf{Preliminaries.} We represent each scenario as a Visual-Genome-style scene graph~\citep{visual_genome}: a directed attributed graph whose nodes denote objects and edges denote relationships. Each node $v$ is defined by an object identifier $o$ and an attribute function $a$ over two fixed fields, \texttt{State} and \texttt{Hazard}. The \texttt{State} captures an object’s physical condition (e.g., closed, exposed), while the \texttt{Hazard} encodes inherent risks (e.g., flammable, breakable). Note that \texttt{Hazard} attributes encode properties, not detection outcomes; a beaker 
labeled \texttt{breakable} does not imply it is actually broken. Each edge is a subject-predicate-object relation describing interactions such as \texttt{stored\_in} or \texttt{near}.
\begin{align}
\mathcal{G} &= (\mathcal{V}, \mathcal{E}), \quad v = (o, a)\label{eq:sg_def} \\
a &: \{\texttt{State}, \texttt{Hazard}\} \rightarrow \mathcal{A}\label{eq:sg_attr} \\
e &= (v_s, p, v_o), \quad e \in \mathcal{E}, \; v_s, v_o \in \mathcal{V}, \; p \in \mathcal{P}
    \label{eq:sg_edge}
\end{align}

\noindent\textbf{Ground-Truth Re-extraction ($\phi_{\text{GT}}$).} 
LabSafetyBench includes scenarios that describe potential risks, even when no concrete safety failure is currently occurring. We therefore re-extract ground-truth labels and define a scenario as \texttt{hazardous} only if an explicit safety violation or incident, such as a chemical spill or missing PPE, is present; otherwise, it is labeled \texttt{non-hazardous}. We prompt GPT-5 to annotate each scenario using the structured mapping $\phi_{\text{GT}} : (s,d,r,t) \rightarrow y$, where $s$ is the laboratory subject, $d$ is the scenario description, $r$ denotes auxiliary safety-related issues, and $t$ is the safety topic, as provided in LabSafetyBench. The auxiliary fields $(r,t)$ are used only as context, not as evidence, since they may describe hypothetical rather than actual hazards. The ground-truth label $y$ includes the binary hazard label $\ell$, the set of explicitly occurring hazards $\mathcal{H}$, and the hazard count $k$. A scenario is labeled \texttt{hazardous} only if at least one definite safety failure is explicitly present; otherwise, it is labeled \texttt{non-hazardous}.
\begin{equation}
\begin{split}
   & y = (\ell, k, \mathcal{H}),\quad
    \ell \in \{\texttt{haz.}, \texttt{non-haz.}\}, \\
    & \mathcal{H} = \{h_1,\ldots,h_k\},\quad k=|\mathcal{H}|
    \label{eq:gt_structured}
\end{split}
\end{equation}

\noindent\textbf{Scene Graph Generation ($\phi_{\mathrm{SG}}$).} Given a textual scenario description $d$, we construct a compact scene graph $\mathcal{G}$ containing only the key objects, attributes, and relations necessary to represent the scenario. We formalize this as $\phi_{\mathrm{SG}} : d \rightarrow \mathcal{G}$. Using GPT-5 with constrained prompting, chain-of-thought reasoning~\citep{chain_of_thought}, a one-shot demonstration~\citep{few_shot}, and a structured output format, we enforce a fixed generation order: selecting objects $\mathcal{V}$, assigning \texttt{State} and \texttt{Hazard} attributes, and finally constructing directed subject-predicate-object relations $\mathcal{E}$.

\noindent\textbf{Graph-Conditioned Image Generation ($\psi_{\text{img}}$).}
Given a scene graph $\mathcal{G}$, we generate a photorealistic laboratory image $I$ using Nano-Banana-pro~\citep{nano_banana_pro}, modeled as $\psi_{\text{img}} : (\mathcal{G}, c) \rightarrow I$, where $c \in \{\texttt{hazardous}, \texttt{non-hazardous}\}$ denotes the scenario class. Class-specific prompts enforce a single-frame, fixed-view rendering, while treating $\mathcal{G}$ as the authoritative specification for object identity, attributes, and relations. For hazardous scenes, the generator is constrained to render hazards through natural visual cues only, without overlays or text. For non-hazardous scenes, it renders a clean and compliant laboratory environment while preserving the structure specified by $\mathcal{G}$.

\noindent\textbf{Image-Scene Alignment.} 
Human annotators first verify each generated tuple $(d,\mathcal{G},y)$. For each verified tuple, we generate multiple candidate images and use a VLM-as-judge to assess image--scene alignment, reducing the need for exhaustive human verification. Alignment requires \textit{(i) structural consistency}, where all objects, attributes, and relations in $\mathcal{G}$ are visibly supported in the image $I$, and \textit{(ii) hazard consistency}, where the hazards visible in $I$ exactly match the ground-truth annotation $y$. We select the judge model $J^\star$ based on agreement with human annotations on a held-out subset, and retain only images labeled \texttt{ALIGNED}, yielding  $\{ (I, \mathcal{G}, y) \mid A_{J^\star}(I) = \texttt{ALIGNED} \}$, where $A_J$ denotes the alignment decision of judge $J$.

\noindent\textbf{Mask Generation.} For each aligned sample $(I,\mathcal{G},y)$, we generate a binary \textit{hazard relevance mask} $M$ marking regions relevant to hazard detection. Let $y=(\ell,k,\mathcal{H})$ as in Eq.~\ref{eq:gt_structured}, and let $t$ be the LabSafetyBench scenario topic. For hazardous samples, Nano-Banana-pro is prompted with both $\mathcal{H}$ and $t$; for non-hazardous samples, it is prompted only with $t$. The output is a binary mask with relevant regions in white and background in black, yielding the final dataset $\mathcal{D}_{\text{final}} = \{(I,\mathcal{G},y,M)\}$.

\vspace{-0.2cm}
\subsection{Hazard Detection Settings}
We evaluate VLMs under four hazard detection settings that vary the available 
input modality and structured information. More details are in 
Appendix~\ref{sec:hazard_detection_settings}. Let \(I\) denote the laboratory 
image, \(\mathcal{G}\) the ground-truth textual scene graph, and 
\(\mathcal{G}_I=\Phi(I)\) the scene graph constructed from the image. 
\begin{align}
  f_{\text{text}}     &: \mathcal{G} \rightarrow (\ell, k, \mathcal{H}) \label{eq:text-only}\\
  f_{\text{img+text}} &: (I, \mathcal{G}) \rightarrow (\ell, k, \mathcal{H}) \label{eq:vision+text}\\
  f_{\text{img}}      &: I \rightarrow (\ell, k, \mathcal{H}) \label{eq:vision-only}\\
  f_{\text{SG}}       &: \mathcal{G}_I \rightarrow (\ell, k, \mathcal{H}), \quad \mathcal{G}_I = \Phi(I) \label{eq:visual_sg_group}
\end{align}

In the text-only setting (Eq.~\ref{eq:text-only}), the model only receives the textual scene graph $\mathcal{G}$.
We further evaluate a hazard-implicit variant \(\mathcal{G}^{-H}\), where explicit \texttt{Hazard} node attributes are 
removed to test whether the model can infer risks from states and relationships of objects alone.
In the vision+text setting (Eq.~\ref{eq:vision+text}), the model receives both the image and the scene graph $(I, \mathcal{G})$.
In the vision-only setting (Eq.~\ref{eq:vision-only}), the model predicts hazards directly from the image $I$.
In the SG-guided prompting (Eq.~\ref{eq:visual_sg_group}), the model is prompted to first reconstruct a scene graph $\mathcal{G}_I$
from the image $I$, then predict hazards.

\section{\sagashort: SG-Aware Guided Attention}
\label{sec:gspo_attention}

\begin{table*}[!t]
\centering
\Large
\renewcommand{\arraystretch}{1.2}
 \resizebox{\textwidth}{!}{
\begin{tabular}{c|c|c|ccc|ccc|ccc|ccc|ccc|ccc}
\toprule
\multirow{3}{*}{\textbf{VLMs}} &
  \multirow{3}{*}{\textbf{M}} &
  \multirow{3}{*}{\textbf{S}} &
  \multicolumn{3}{c|}{\textbf{Overall}} &
  \multicolumn{3}{c|}{\textbf{Biology}} &
  \multicolumn{3}{c|}{\textbf{Chemistry}} &
  \multicolumn{3}{c|}{\textbf{Cryogenic Liquids}} &
  \multicolumn{3}{c|}{\textbf{Physics}} &
  \multicolumn{3}{c}{\textbf{General}} \\ \cline{4-21} 
 &
   &
   &
  \multicolumn{1}{c}{\multirow{2}{*}{\textbf{FE}}} &
  \multicolumn{1}{c}{\textbf{Label}} &
  \textbf{Count} &
  \multicolumn{1}{c}{\multirow{2}{*}{\textbf{FE}}} &
  \multicolumn{1}{c}{\textbf{Label}} &
  \textbf{Count} &
  \multicolumn{1}{c}{\multirow{2}{*}{\textbf{FE}}} &
  \textbf{Label} &
  \textbf{Count} &
  \multicolumn{1}{c}{\multirow{2}{*}{\textbf{FE}}} &
  \textbf{Label} &
  \textbf{Count} &
  \multicolumn{1}{c}{\multirow{2}{*}{\textbf{FE}}} &
  \textbf{Label} &
  \textbf{Count} &
  \multicolumn{1}{c}{\multirow{2}{*}{\textbf{FE}}} &
  \textbf{Label} &
  \textbf{Count} \\ \cline{5-6} \cline{8-9} \cline{11-12} \cline{14-15} \cline{17-18} \cline{20-21} 
 &
   &
   &
  \multicolumn{1}{c}{} &
  \multicolumn{1}{c}{$\mathbf{F_1}$} &
  \textbf{MAE} &
  \multicolumn{1}{c}{} &
  \multicolumn{1}{c}{$\mathbf{F_1}$} &
  \textbf{MAE} &
  \multicolumn{1}{c}{} &
  \multicolumn{1}{c}{$\mathbf{F_1}$} &
  \textbf{MAE} &
  \multicolumn{1}{c}{} &
  \multicolumn{1}{c}{$\mathbf{F_1}$} &
  \textbf{MAE} &
  \multicolumn{1}{c}{} &
  \multicolumn{1}{c}{$\mathbf{F_1}$} &
  \textbf{MAE} &
  \multicolumn{1}{c}{} &
  \multicolumn{1}{c}{$\mathbf{F_1}$} &
  \textbf{MAE} \\ \toprule
\multirow{6}{*}{\begin{tabular}[c]{@{}c@{}}llava-\\ next-7B\end{tabular}} &
  \multirow{2}{*}{V} &
  - &
  \multicolumn{1}{c}{0} &
  \multicolumn{1}{c}{\cc{0}} &
  0.55 &
  \multicolumn{1}{c}{0} &
  \multicolumn{1}{c}{0} &
  0.42 &
  \multicolumn{1}{c}{0} &
  \multicolumn{1}{c}{0} &
  0.68 &
  \multicolumn{1}{c}{0} &
  \multicolumn{1}{c}{0} &
  0.85 &
  \multicolumn{1}{c}{0} &
  \multicolumn{1}{c}{0} &
  0.28 &
  \multicolumn{1}{c}{0} &
  \multicolumn{1}{c}{0} &
  0.62 \\ \cline{3-21} 
 &
   &
  SG-G &
  \multicolumn{1}{c}{31.89} &
  \multicolumn{1}{c}{\cc{0.22}} &
  0.55 &
  \multicolumn{1}{c}{38.98} &
  \multicolumn{1}{c}{0} &
  0.52 &
  \multicolumn{1}{c}{27.67} &
  \multicolumn{1}{c}{0.26} &
  0.65 &
  \multicolumn{1}{c}{19.86} &
  \multicolumn{1}{c}{0.28} &
  0.83 &
  \multicolumn{1}{c}{35.74} &
  \multicolumn{1}{c}{0.23} &
  0.31 &
  \multicolumn{1}{c}{33.59} &
  \multicolumn{1}{c}{0.16} &
  0.6 \\ \cline{2-21} 
 &
  \multirow{2}{*}{TSG} &
  + H &
  \multicolumn{1}{c}{6.61} &
  \multicolumn{1}{c}{0.72} &
  0.62 &
  \multicolumn{1}{c}{0} &
  \multicolumn{1}{c}{0.75} &
  0.44 &
  \multicolumn{1}{c}{0} &
  \multicolumn{1}{c}{0.8} &
  0.48 &
  \multicolumn{1}{c}{0} &
  \multicolumn{1}{c}{0.86} &
  0.57 &
  \multicolumn{1}{c}{20} &
  \multicolumn{1}{c}{0.45} &
  0.65 &
  \multicolumn{1}{c}{9.13} &
  \multicolumn{1}{c}{0.67} &
  0.82 \\ \cline{3-21} 
 &
   &
  - H &
  \multicolumn{1}{c}{6.53} &
  \multicolumn{1}{c}{\cc{0.54}} &
  0.42 &
  \multicolumn{1}{c}{0} &
  \multicolumn{1}{c}{0.59} &
  0.29 &
  \multicolumn{1}{c}{0} &
  \multicolumn{1}{c}{0.57} &
  0.5 &
  \multicolumn{1}{c}{0} &
  \multicolumn{1}{c}{0.49} &
  0.64 &
  \multicolumn{1}{c}{20} &
  \multicolumn{1}{c}{0.31} &
  0.18 &
  \multicolumn{1}{c}{8.7} &
  \multicolumn{1}{c}{0.54} &
  0.49 \\ \cline{2-21} 
 &
  \multirow{2}{*}{\begin{tabular}[c]{@{}c@{}}V \\ + TSG\end{tabular}} &
  + H &
  \multicolumn{1}{c}{0} &
  \multicolumn{1}{c}{0.11} &
  0.53 &
  \multicolumn{1}{c}{0} &
  \multicolumn{1}{c}{0.04} &
  0.41 &
  \multicolumn{1}{c}{0} &
  \multicolumn{1}{c}{0.12} &
  0.65 &
  \multicolumn{1}{c}{0} &
  \multicolumn{1}{c}{0} &
  0.85 &
  \multicolumn{1}{c}{0} &
  \multicolumn{1}{c}{0.2} &
  0.24 &
  \multicolumn{1}{c}{0} &
  \multicolumn{1}{c}{0.09} &
  0.59 \\ \cline{3-21} 
 &
   &
  - H &
  \multicolumn{1}{c}{0} &
  \multicolumn{1}{c}{0.11} &
  0.53 &
  \multicolumn{1}{c}{0} &
  \multicolumn{1}{c}{0.17} &
  0.38 &
  \multicolumn{1}{c}{0} &
  \multicolumn{1}{c}{0.1} &
  0.66 &
  \multicolumn{1}{c}{0} &
  \multicolumn{1}{c}{0} &
  0.85 &
  \multicolumn{1}{c}{0} &
  \multicolumn{1}{c}{0.17} &
  0.26 &
  \multicolumn{1}{c}{0} &
  \multicolumn{1}{c}{0.08} &
  0.6 \\ \midrule \midrule
\multirow{6}{*}{\begin{tabular}[c]{@{}c@{}}qwen3\\ -VL-8B\end{tabular}} &
  \multirow{2}{*}{V} &
  - &
  \multicolumn{1}{c}{0} &
  \multicolumn{1}{c}{\cc{0.68}} &
  0.36 &
  \multicolumn{1}{c}{0} &
  \multicolumn{1}{c}{0.74} &
  0.23 &
  \multicolumn{1}{c}{0} &
  \multicolumn{1}{c}{0.73} &
  0.45 &
  \multicolumn{1}{c}{0} &
  \multicolumn{1}{c}{0.44} &
  0.67 &
  \multicolumn{1}{c}{0} &
  \multicolumn{1}{c}{0.67} &
  0.17 &
  \multicolumn{1}{c}{0} &
  \multicolumn{1}{c}{0.63} &
  0.41 \\ \cline{3-21} 
 &
   &
  SG-G &
  \multicolumn{1}{c}{5.55} &
  \multicolumn{1}{c}{\cc{0.78}} &
  0.45 &
  \multicolumn{1}{c}{13.23} &
  \multicolumn{1}{c}{0.79} &
  0.46 &
  \multicolumn{1}{c}{4.71} &
  \multicolumn{1}{c}{0.82} &
  0.47 &
  \multicolumn{1}{c}{0} &
  \multicolumn{1}{c}{0.87} &
  0.58 &
  \multicolumn{1}{c}{6.51} &
  \multicolumn{1}{c}{0.64} &
  0.3 &
  \multicolumn{1}{c}{3.68} &
  \multicolumn{1}{c}{0.75} &
  0.55 \\ \cline{2-21} 
 &
  \multirow{2}{*}{TSG} &
  + H &
  \multicolumn{1}{c}{0.08} &
  \multicolumn{1}{c}{0.87} &
  0.37 &
  \multicolumn{1}{c}{0} &
  \multicolumn{1}{c}{0.95} &
  0.19 &
  \multicolumn{1}{c}{0} &
  \multicolumn{1}{c}{0.88} &
  0.38 &
  \multicolumn{1}{c}{0} &
  \multicolumn{1}{c}{0.88} &
  0.58 &
  \multicolumn{1}{c}{0.44} &
  \multicolumn{1}{c}{0.81} &
  0.24 &
  \multicolumn{1}{c}{0} &
  \multicolumn{1}{c}{0.85} &
  0.51 \\ \cline{3-21} 
 &
   &
  - H &
  \multicolumn{1}{c}{0.34} &
  \multicolumn{1}{c}{\cc{0.93}} &
  0.21 &
  \multicolumn{1}{c}{0} &
  \multicolumn{1}{c}{0.96} &
  0.12 &
  \multicolumn{1}{c}{0} &
  \multicolumn{1}{c}{0.93} &
  0.27 &
  \multicolumn{1}{c}{0} &
  \multicolumn{1}{c}{0.86} &
  0.37 &
  \multicolumn{1}{c}{0.44} &
  \multicolumn{1}{c}{0.92} &
  0.08 &
  \multicolumn{1}{c}{1.08} &
  \multicolumn{1}{c}{0.94} &
  0.21 \\ \cline{2-21} 
 &
  \multirow{2}{*}{\begin{tabular}[c]{@{}c@{}}V \\ + TSG\end{tabular}} &
  + H &
  \multicolumn{1}{c}{0} &
  \multicolumn{1}{c}{0.9} &
  0.23 &
  \multicolumn{1}{c}{0} &
  \multicolumn{1}{c}{0.93} &
  0.15 &
  \multicolumn{1}{c}{0} &
  \multicolumn{1}{c}{0.93} &
  0.29 &
  \multicolumn{1}{c}{0} &
  \multicolumn{1}{c}{0.7} &
  0.52 &
  \multicolumn{1}{c}{0} &
  \multicolumn{1}{c}{0.91} &
  0.09 &
  \multicolumn{1}{c}{0} &
  \multicolumn{1}{c}{0.88} &
  0.24 \\ \cline{3-21} 
 &
   &
  - H &
  \multicolumn{1}{c}{0.17} &
  \multicolumn{1}{c}{0.87} &
  0.26 &
  \multicolumn{1}{c}{0} &
  \multicolumn{1}{c}{0.93} &
  0.13 &
  \multicolumn{1}{c}{0} &
  \multicolumn{1}{c}{0.89} &
  0.35 &
  \multicolumn{1}{c}{0} &
  \multicolumn{1}{c}{0.68} &
  0.54 &
  \multicolumn{1}{c}{0} &
  \multicolumn{1}{c}{0.82} &
  0.13 &
  \multicolumn{1}{c}{0.73} &
  \multicolumn{1}{c}{0.88} &
  0.25 \\ \midrule \midrule
\multirow{6}{*}{\begin{tabular}[c]{@{}c@{}}llama\\ -3.2\\ -11B\end{tabular}} &
  \multirow{2}{*}{V} &
  - &
  \multicolumn{1}{c}{58.28} &
  \multicolumn{1}{c}{\cc{0.43}} &
  0.51 &
  \multicolumn{1}{c}{64.61} &
  \multicolumn{1}{c}{0.66} &
  0.59 &
  \multicolumn{1}{c}{56.53} &
  \multicolumn{1}{c}{0.41} &
  0.56 &
  \multicolumn{1}{c}{75.94} &
  \multicolumn{1}{c}{0.87} &
  0.67 &
  \multicolumn{1}{c}{49.33} &
  \multicolumn{1}{c}{0.37} &
  0.29 &
  \multicolumn{1}{c}{64.56} &
  \multicolumn{1}{c}{0.33} &
  0.67 \\ \cline{3-21} 
 &
   &
  SG-G &
  \multicolumn{1}{c}{97.1} &
  \multicolumn{1}{c}{\cc{0.71}} &
  0.9 &
  \multicolumn{1}{c}{95.73} &
  \multicolumn{1}{c}{0.6} &
  0.7 &
  \multicolumn{1}{c}{98.81} &
  \multicolumn{1}{c}{0.6} &
  0.93 &
  \multicolumn{1}{c}{91.1} &
  \multicolumn{1}{c}{0.4} &
  0.6 &
  \multicolumn{1}{c}{97.28} &
  \multicolumn{1}{c}{0.56} &
  0.67 &
  \multicolumn{1}{c}{95.34} &
  \multicolumn{1}{c}{0.43} &
  0.9 \\ \cline{2-21} 
 &
  \multirow{2}{*}{TSG} &
  + H &
  \multicolumn{1}{c}{0} &
  \multicolumn{1}{c}{0.77} &
  1.35 &
  \multicolumn{1}{c}{0} &
  \multicolumn{1}{c}{0.87} &
  0.99 &
  \multicolumn{1}{c}{0} &
  \multicolumn{1}{c}{0.78} &
  1.34 &
  \multicolumn{1}{c}{0} &
  \multicolumn{1}{c}{0.92} &
  1.43 &
  \multicolumn{1}{c}{0} &
  \multicolumn{1}{c}{0.62} &
  1.22 &
  \multicolumn{1}{c}{0} &
  \multicolumn{1}{c}{0.79} &
  1.62 \\ \cline{3-21} 
 &
   &
  - H &
  \multicolumn{1}{c}{0} &
  \multicolumn{1}{c}{\cc{0.88}} &
  0.24 &
  \multicolumn{1}{c}{0} &
  \multicolumn{1}{c}{0.95} &
  0.23 &
  \multicolumn{1}{c}{0} &
  \multicolumn{1}{c}{0.9} &
  0.22 &
  \multicolumn{1}{c}{0} &
  \multicolumn{1}{c}{0.82} &
  0.53 &
  \multicolumn{1}{c}{0} &
  \multicolumn{1}{c}{0.84} &
  0.17 &
  \multicolumn{1}{c}{0} &
  \multicolumn{1}{c}{0.87} &
  0.31 \\ \cline{2-21} 
 &
  \multirow{2}{*}{\begin{tabular}[c]{@{}c@{}}V \\ + TSG\end{tabular}} &
  + H &
  \multicolumn{1}{c}{3.32} &
  \multicolumn{1}{c}{0.69} &
  0.56 &
  \multicolumn{1}{c}{7.62} &
  \multicolumn{1}{c}{0.62} &
  0.53 &
  \multicolumn{1}{c}{1.18} &
  \multicolumn{1}{c}{0.78} &
  0.47 &
  \multicolumn{1}{c}{14.64} &
  \multicolumn{1}{c}{0.78} &
  0.94 &
  \multicolumn{1}{c}{4.84} &
  \multicolumn{1}{c}{0.46} &
  0.62 &
  \multicolumn{1}{c}{2.52} &
  \multicolumn{1}{c}{0.73} &
  0.65 \\ \cline{3-21} 
 &
   &
  - H &
  \multicolumn{1}{c}{6.06} &
  \multicolumn{1}{c}{0.79} &
  0.36 &
  \multicolumn{1}{c}{10.5} &
  \multicolumn{1}{c}{0.82} &
  0.22 &
  \multicolumn{1}{c}{1.37} &
  \multicolumn{1}{c}{0.86} &
  0.4 &
  \multicolumn{1}{c}{18.97} &
  \multicolumn{1}{c}{0.9} &
  0.4 &
  \multicolumn{1}{c}{8.46} &
  \multicolumn{1}{c}{0.66} &
  0.25 &
  \multicolumn{1}{c}{8.69} &
  \multicolumn{1}{c}{0.74} &
  0.42 \\ \midrule \midrule
\multirow{6}{*}{\begin{tabular}[c]{@{}c@{}}mistral-\\ medium-3\end{tabular}} &
  \multirow{2}{*}{V} &
  - &
  \multicolumn{1}{c}{9.67} &
  \multicolumn{1}{c}{\cc{0.61}} &
  0.41 &
  \multicolumn{1}{c}{0} &
  \multicolumn{1}{c}{0.63} &
  0.27 &
  \multicolumn{1}{c}{0.39} &
  \multicolumn{1}{c}{0.62} &
  0.49 &
  \multicolumn{1}{c}{17.78} &
  \multicolumn{1}{c}{0.31} &
  0.6 &
  \multicolumn{1}{c}{20} &
  \multicolumn{1}{c}{0.48} &
  0.14 &
  \multicolumn{1}{c}{20} &
  \multicolumn{1}{c}{0.47} &
  0.34 \\ \cline{3-21} 
 &
   &
  SG-G &
  \multicolumn{1}{c}{9.75} &
  \multicolumn{1}{c}{\cc{0.69}} &
  0.42 &
  \multicolumn{1}{c}{0} &
  \multicolumn{1}{c}{0.66} &
  0.33 &
  \multicolumn{1}{c}{0.39} &
  \multicolumn{1}{c}{0.73} &
  0.45 &
  \multicolumn{1}{c}{20} &
  \multicolumn{1}{c}{0.47} &
  0.56 &
  \multicolumn{1}{c}{20} &
  \multicolumn{1}{c}{0.42} &
  0.29 &
  \multicolumn{1}{c}{20} &
  \multicolumn{1}{c}{0.58} &
  0.34 \\ \cline{2-21} 
 &
  \multirow{2}{*}{TSG} &
  + H &
  \multicolumn{1}{c}{9.75} &
  \multicolumn{1}{c}{0.81} &
  0.67 &
  \multicolumn{1}{c}{0} &
  \multicolumn{1}{c}{0.77} &
  0.84 &
  \multicolumn{1}{c}{0.59} &
  \multicolumn{1}{c}{0.86} &
  0.53 &
  \multicolumn{1}{c}{17.78} &
  \multicolumn{1}{c}{0.95} &
  0.74 &
  \multicolumn{1}{c}{20} &
  \multicolumn{1}{c}{0.44} &
  0.71 &
  \multicolumn{1}{c}{20} &
  \multicolumn{1}{c}{0.69} &
  0.62 \\ \cline{3-21} 
 &
   &
  - H &
  \multicolumn{1}{c}{9.75} &
  \multicolumn{1}{c}{\cc{0.92}} &
  0.17 &
  \multicolumn{1}{c}{0} &
  \multicolumn{1}{c}{0.9} &
  0.16 &
  \multicolumn{1}{c}{0.39} &
  \multicolumn{1}{c}{0.93} &
  0.18 &
  \multicolumn{1}{c}{20} &
  \multicolumn{1}{c}{0.68} &
  0.3 &
  \multicolumn{1}{c}{20} &
  \multicolumn{1}{c}{0.68} &
  0.07 &
  \multicolumn{1}{c}{20} &
  \multicolumn{1}{c}{0.76} &
  0.14 \\ \cline{2-21} 
 &
  \multirow{2}{*}{\begin{tabular}[c]{@{}c@{}}V \\ + TSG\end{tabular}} &
  + H &
  \multicolumn{1}{c}{9.67} &
  \multicolumn{1}{c}{0.9} &
  0.23 &
  \multicolumn{1}{c}{0} &
  \multicolumn{1}{c}{0.95} &
  0.13 &
  \multicolumn{1}{c}{0.2} &
  \multicolumn{1}{c}{0.91} &
  0.25 &
  \multicolumn{1}{c}{20} &
  \multicolumn{1}{c}{0.5} &
  0.6 &
  \multicolumn{1}{c}{20} &
  \multicolumn{1}{c}{0.71} &
  0.08 &
  \multicolumn{1}{c}{20} &
  \multicolumn{1}{c}{0.71} &
  0.21 \\ \cline{3-21} 
 &
   &
  - H &
  \multicolumn{1}{c}{9.83} &
  \multicolumn{1}{c}{0.89} &
  0.23 &
  \multicolumn{1}{c}{0} &
  \multicolumn{1}{c}{0.95} &
  0.11 &
  \multicolumn{1}{c}{0.59} &
  \multicolumn{1}{c}{0.89} &
  0.27 &
  \multicolumn{1}{c}{20} &
  \multicolumn{1}{c}{0.52} &
  0.47 &
  \multicolumn{1}{c}{20} &
  \multicolumn{1}{c}{0.7} &
  0.08 &
  \multicolumn{1}{c}{20} &
  \multicolumn{1}{c}{0.71} &
  0.22 \\ \midrule \midrule
\multirow{6}{*}{\begin{tabular}[c]{@{}c@{}}claude-\\ sonnet-4.5\end{tabular}} &
  \multirow{2}{*}{V} &
  - &
  \multicolumn{1}{c}{0} &
  \multicolumn{1}{c}{\cc{0.62}} &
  0.39 &
  \multicolumn{1}{c}{0} &
  \multicolumn{1}{c}{0.43} &
  0.33 &
  \multicolumn{1}{c}{0} &
  \multicolumn{1}{c}{0.66} &
  0.46 &
  \multicolumn{1}{c}{0} &
  \multicolumn{1}{c}{0.75} &
  0.59 &
  \multicolumn{1}{c}{0} &
  \multicolumn{1}{c}{0.54} &
  0.2 &
  \multicolumn{1}{c}{0} &
  \multicolumn{1}{c}{0.61} &
  0.42 \\ \cline{3-21} 
 &
   &
  SG-G &
  \multicolumn{1}{c}{0.08} &
  \multicolumn{1}{c}{\cc{0.7}} &
  0.45 &
  \multicolumn{1}{c}{0} &
  \multicolumn{1}{c}{0.55} &
  0.38 &
  \multicolumn{1}{c}{0} &
  \multicolumn{1}{c}{0.75} &
  0.45 &
  \multicolumn{1}{c}{0} &
  \multicolumn{1}{c}{0.76} &
  0.87 &
  \multicolumn{1}{c}{0} &
  \multicolumn{1}{c}{0.53} &
  0.34 &
  \multicolumn{1}{c}{0.35} &
  \multicolumn{1}{c}{0.74} &
  0.53 \\ \cline{2-21} 
 &
  \multirow{2}{*}{TSG} &
  + H &
  \multicolumn{1}{c}{5.71} &
  \multicolumn{1}{c}{0.93} &
  0.23 &
  \multicolumn{1}{c}{47.73} &
  \multicolumn{1}{c}{1} &
  0.16 &
  \multicolumn{1}{c}{0} &
  \multicolumn{1}{c}{0.93} &
  0.19 &
  \multicolumn{1}{c}{0} &
  \multicolumn{1}{c}{0.96} &
  0.61 &
  \multicolumn{1}{c}{1.92} &
  \multicolumn{1}{c}{0.85} &
  0.15 &
  \multicolumn{1}{c}{2.52} &
  \multicolumn{1}{c}{0.97} &
  0.34 \\ \cline{3-21} 
 &
   &
  - H &
  \multicolumn{1}{c}{5.88} &
  \multicolumn{1}{c}{\cc{0.94}} &
  0.16 &
  \multicolumn{1}{c}{50.14} &
  \multicolumn{1}{c}{0.95} &
  0.25 &
  \multicolumn{1}{c}{0.19} &
  \multicolumn{1}{c}{0.94} &
  0.13 &
  \multicolumn{1}{c}{0} &
  \multicolumn{1}{c}{0.96} &
  0.25 &
  \multicolumn{1}{c}{1.15} &
  \multicolumn{1}{c}{0.93} &
  0.07 &
  \multicolumn{1}{c}{2.52} &
  \multicolumn{1}{c}{0.92} &
  0.27 \\ \cline{2-21} 
 &
  \multirow{2}{*}{\begin{tabular}[c]{@{}c@{}}V \\ + TSG\end{tabular}} &
  + H &
  \multicolumn{1}{c}{5.14} &
  \multicolumn{1}{c}{0.84} &
  0.27 &
  \multicolumn{1}{c}{43.98} &
  \multicolumn{1}{c}{0.93} &
  0.17 &
  \multicolumn{1}{c}{0} &
  \multicolumn{1}{c}{0.84} &
  0.3 &
  \multicolumn{1}{c}{0} &
  \multicolumn{1}{c}{0.79} &
  0.53 &
  \multicolumn{1}{c}{1.18} &
  \multicolumn{1}{c}{0.71} &
  0.19 &
  \multicolumn{1}{c}{2.52} &
  \multicolumn{1}{c}{0.91} &
  0.27 \\ \cline{3-21} 
 &
   &
  - H &
  \multicolumn{1}{c}{5.39} &
  \multicolumn{1}{c}{0.86} &
  0.26 &
  \multicolumn{1}{c}{46.45} &
  \multicolumn{1}{c}{0.9} &
  0.19 &
  \multicolumn{1}{c}{0.19} &
  \multicolumn{1}{c}{0.87} &
  0.29 &
  \multicolumn{1}{c}{0} &
  \multicolumn{1}{c}{0.74} &
  0.46 &
  \multicolumn{1}{c}{1.18} &
  \multicolumn{1}{c}{0.72} &
  0.18 &
  \multicolumn{1}{c}{2.14} &
  \multicolumn{1}{c}{0.9} &
  0.26 \\ \midrule \midrule
\multirow{6}{*}{\begin{tabular}[c]{@{}c@{}}gemini-\\ 3-flash\end{tabular}} &
  \multirow{2}{*}{V} &
  - &
  \multicolumn{1}{c}{0.17} &
  \multicolumn{1}{c}{\cc{0.85}} &
  0.37 &
  \multicolumn{1}{c}{0} &
  \multicolumn{1}{c}{0.81} &
  0.27 &
  \multicolumn{1}{c}{0.39} &
  \multicolumn{1}{c}{0.88} &
  0.42 &
  \multicolumn{1}{c}{0} &
  \multicolumn{1}{c}{0.73} &
  0.66 &
  \multicolumn{1}{c}{0} &
  \multicolumn{1}{c}{0.78} &
  0.2 &
  \multicolumn{1}{c}{0} &
  \multicolumn{1}{c}{0.85} &
  0.44 \\ \cline{3-21} 
 &
   &
  SG-G &
  \multicolumn{1}{c}{1.42} &
  \multicolumn{1}{c}{\cc{0.85}} &
  0.38 &
  \multicolumn{1}{c}{1} &
  \multicolumn{1}{c}{0.79} &
  0.35 &
  \multicolumn{1}{c}{1.2} &
  \multicolumn{1}{c}{0.88} &
  0.42 &
  \multicolumn{1}{c}{0} &
  \multicolumn{1}{c}{0.82} &
  0.57 &
  \multicolumn{1}{c}{2.52} &
  \multicolumn{1}{c}{0.83} &
  0.2 &
  \multicolumn{1}{c}{1.4} &
  \multicolumn{1}{c}{0.84} &
  0.47 \\ \cline{2-21} 
 &
  \multirow{2}{*}{TSG} &
  + H &
  \multicolumn{1}{c}{0.33} &
  \multicolumn{1}{c}{0.95} &
  0.21 &
  \multicolumn{1}{c}{0} &
  \multicolumn{1}{c}{1} &
  0.15 &
  \multicolumn{1}{c}{0.39} &
  \multicolumn{1}{c}{0.96} &
  0.16 &
  \multicolumn{1}{c}{0} &
  \multicolumn{1}{c}{0.96} &
  0.59 &
  \multicolumn{1}{c}{0.79} &
  \multicolumn{1}{c}{0.87} &
  0.11 &
  \multicolumn{1}{c}{0} &
  \multicolumn{1}{c}{0.94} &
  0.35 \\ \cline{3-21} 
 &
   &
  - H &
  \multicolumn{1}{c}{0.25} &
  \multicolumn{1}{c}{\cc{0.96}} &
  0.13 &
  \multicolumn{1}{c}{0} &
  \multicolumn{1}{c}{0.97} &
  0.13 &
  \multicolumn{1}{c}{0.21} &
  \multicolumn{1}{c}{0.97} &
  0.12 &
  \multicolumn{1}{c}{3.33} &
  \multicolumn{1}{c}{0.89} &
  0.27 &
  \multicolumn{1}{c}{0.4} &
  \multicolumn{1}{c}{0.9} &
  0.08 &
  \multicolumn{1}{c}{0} &
  \multicolumn{1}{c}{0.97} &
  0.19 \\ \cline{2-21} 
 &
  \multirow{2}{*}{\begin{tabular}[c]{@{}c@{}}V \\ + TSG\end{tabular}} &
  + H &
  \multicolumn{1}{c}{0.41} &
  \multicolumn{1}{c}{0.92} &
  0.24 &
  \multicolumn{1}{c}{0} &
  \multicolumn{1}{c}{0.95} &
  0.14 &
  \multicolumn{1}{c}{0.6} &
  \multicolumn{1}{c}{0.94} &
  0.26 &
  \multicolumn{1}{c}{0} &
  \multicolumn{1}{c}{0.9} &
  0.45 &
  \multicolumn{1}{c}{0.34} &
  \multicolumn{1}{c}{0.9} &
  0.14 &
  \multicolumn{1}{c}{0.38} &
  \multicolumn{1}{c}{0.92} &
  0.3 \\ \cline{3-21} 
 &
   &
  - H &
  \multicolumn{1}{c}{0.08} &
  \multicolumn{1}{c}{0.92} &
  0.27 &
  \multicolumn{1}{c}{0} &
  \multicolumn{1}{c}{0.92} &
  0.18 &
  \multicolumn{1}{c}{0} &
  \multicolumn{1}{c}{0.92} &
  0.3 &
  \multicolumn{1}{c}{0} &
  \multicolumn{1}{c}{0.92} &
  0.41 &
  \multicolumn{1}{c}{0} &
  \multicolumn{1}{c}{0.91} &
  0.13 &
  \multicolumn{1}{c}{0.35} &
  \multicolumn{1}{c}{0.9} &
  0.34 \\ \midrule \midrule
\multirow{6}{*}{\begin{tabular}[c]{@{}c@{}}gemini-\\ 3-pro\end{tabular}} &
  \multirow{2}{*}{V} &
  - &
  \multicolumn{1}{c}{1.08} &
  \multicolumn{1}{c}{\cc{0.86}} &
  0.33 &
  \multicolumn{1}{c}{1} &
  \multicolumn{1}{c}{0.77} &
  0.38 &
  \multicolumn{1}{c}{1.38} &
  \multicolumn{1}{c}{0.9} &
  0.36 &
  \multicolumn{1}{c}{0} &
  \multicolumn{1}{c}{0.74} &
  0.72 &
  \multicolumn{1}{c}{0.34} &
  \multicolumn{1}{c}{0.81} &
  0.17 &
  \multicolumn{1}{c}{1.44} &
  \multicolumn{1}{c}{0.87} &
  0.38 \\ \cline{3-21} 
 &
   &
  SG-G &
  \multicolumn{1}{c}{0.42} &
  \multicolumn{1}{c}{\cc{0.85}} &
  0.37 &
  \multicolumn{1}{c}{0.74} &
  \multicolumn{1}{c}{0.79} &
  0.43 &
  \multicolumn{1}{c}{0.4} &
  \multicolumn{1}{c}{0.87} &
  0.38 &
  \multicolumn{1}{c}{0} &
  \multicolumn{1}{c}{0.86} &
  0.72 &
  \multicolumn{1}{c}{0} &
  \multicolumn{1}{c}{0.78} &
  0.23 &
  \multicolumn{1}{c}{0.73} &
  \multicolumn{1}{c}{0.89} &
  0.42 \\ \cline{2-21} 
 &
  \multirow{2}{*}{TSG} &
  + H &
  \multicolumn{1}{c}{0.08} &
  \multicolumn{1}{c}{0.96} &
  0.13 &
  \multicolumn{1}{c}{0} &
  \multicolumn{1}{c}{0.99} &
  0.13 &
  \multicolumn{1}{c}{0.19} &
  \multicolumn{1}{c}{0.96} &
  0.13 &
  \multicolumn{1}{c}{0} &
  \multicolumn{1}{c}{0.96} &
  0.37 &
  \multicolumn{1}{c}{0} &
  \multicolumn{1}{c}{0.96} &
  0.04 &
  \multicolumn{1}{c}{0} &
  \multicolumn{1}{c}{0.97} &
  0.17 \\ \cline{3-21} 
 &
   &
  - H &
  \multicolumn{1}{c}{0.25} &
  \multicolumn{1}{c}{\cc{0.96}} &
  0.14 &
  \multicolumn{1}{c}{0} &
  \multicolumn{1}{c}{0.97} &
  0.12 &
  \multicolumn{1}{c}{0.19} &
  \multicolumn{1}{c}{0.96} &
  0.14 &
  \multicolumn{1}{c}{0} &
  \multicolumn{1}{c}{0.96} &
  0.34 &
  \multicolumn{1}{c}{0.34} &
  \multicolumn{1}{c}{0.89} &
  0.08 &
  \multicolumn{1}{c}{0.35} &
  \multicolumn{1}{c}{0.97} &
  0.15 \\ \cline{2-21} 
 &
  \multirow{2}{*}{\begin{tabular}[c]{@{}c@{}}V \\ + TSG\end{tabular}} &
  + H &
  \multicolumn{1}{c}{0.58} &
  \multicolumn{1}{c}{0.95} &
  0.21 &
  \multicolumn{1}{c}{0} &
  \multicolumn{1}{c}{0.94} &
  0.16 &
  \multicolumn{1}{c}{0.57} &
  \multicolumn{1}{c}{0.97} &
  0.24 &
  \multicolumn{1}{c}{2} &
  \multicolumn{1}{c}{0.83} &
  0.51 &
  \multicolumn{1}{c}{0.34} &
  \multicolumn{1}{c}{0.93} &
  0.09 &
  \multicolumn{1}{c}{0.77} &
  \multicolumn{1}{c}{0.93} &
  0.24 \\ \cline{3-21} 
 &
   &
  - H &
  \multicolumn{1}{c}{0.83} &
  \multicolumn{1}{c}{0.94} &
  0.22 &
  \multicolumn{1}{c}{0.74} &
  \multicolumn{1}{c}{0.93} &
  0.2 &
  \multicolumn{1}{c}{0.79} &
  \multicolumn{1}{c}{0.96} &
  0.24 &
  \multicolumn{1}{c}{2} &
  \multicolumn{1}{c}{0.89} &
  0.44 &
  \multicolumn{1}{c}{0.79} &
  \multicolumn{1}{c}{0.93} &
  0.1 &
  \multicolumn{1}{c}{0.73} &
  \multicolumn{1}{c}{0.94} &
  0.28 \\ \bottomrule
\end{tabular}}
\vspace{-0.5em}
\caption{Hazard detection performance across input modalities (M) and reasoning settings (S): overall and domain-wise $F_1$, format error (FE), and mean absolute error (MAE) for vision-only (V), textual scene graph with (TSG+H) and without (TSG-H) hazard attributes, and vision + textual scene graph (V+TSG) inputs. Precision and recall are in \autoref{appendix:full_results}.
Scene-graph-based settings, with or without images, outperform vision-only in nearly all cases, showing that VLMs reason effectively given structured object relationships. SG-guided prompting improves vision-only detection, suggesting that safety knowledge is present but becomes usable once models explicitly structure the scene. Darker shading indicates higher $F_1$.}
\label{tab:results}
\vspace{-1em}
\end{table*}

Our evaluation with the {SG-based prompting} approach shows that VLMs can reason more effectively once visual scenes are translated into structured text, but the performance remains suboptimal due to reconstruction errors.
We thus train VLMs to improve hazard detection through two objectives: \textit{i) scene-graph reasoning} and \textit{ii) visual grounding}. The first objective teaches the model to understand the visual scene through structuring the objects before making a hazard prediction, while the second encourages it to focus on the relevant portions of the image. We optimize these objectives using \textit{scene-graph-based Group Sequence Policy Optimization (SG-based GSPO)}, which rewards both the required scene-graph reasoning format and the correct hazard label, together with an \textit{attention-alignment} loss that guides answer-token attention toward the ground-truth relevant image regions.

\noindent\textbf{SG-based GSPO.} Given an image input $x_i$ and a hazard-detection prompt $x_t$, the policy model $\pi_\theta$ generates an output $o$ in two structured blocks: a \texttt{<REASONING>} block containing the scene graph, followed by a \texttt{<SOLUTION>} block containing the final hazard classification.
For each input pair $(x_i,x_t)$, the policy samples $G$ candidate outputs 
$\{o^{(g)}\}_{g=1}^{G}$ from the current policy, $o^{(g)} \sim \pi_\theta(\cdot \mid x_i, x_t)$. Each output is scored using the reward: $R(o) = R_{\mathrm{format}}(o) + R_{\mathrm{correct}}(o, o_{gt})$, which combines a formatting reward for the required \texttt{<REASONING>} and \texttt{<SOLUTION>} structure with a correctness reward for matching the ground-truth hazard label $o_{gt}$. The GSPO~\citep{GSPO} objective in Eq.~\ref{eq:gspo_loss} optimizes the sampled outputs using the sequence-level importance ratio $s^{(g)}(\theta)$, group-normalized advantage $\hat{A}^{(g)}$, the clipping coefficient  $\epsilon_c$.
\begin{equation}\label{eq:gspo_loss}
\begin{split}
 \mathcal{L}_{\mathrm{GSPO}}
&=
-\frac{1}{G}
\sum_{g=1}^{G}
\min
\Big(
s^{(g)}(\theta)\hat{A}^{(g)}, \\
& \operatorname{clip}
\left(
s^{(g)}(\theta),
1-\epsilon_c,
1+\epsilon_c
\right)
\hat{A}^{(g)}
\Big)
\end{split}
\end{equation}
\noindent\textbf{Attention Alignment.} We augment the GSPO objective with an attention-alignment loss that aligns the model's attention over image tokens with ground-truth binary masks, encouraging it to focus on the relevant image regions when generating the final answer. Let $A^{(l)} \in \mathbb{R}^{B \times H \times S \times S}$ denote the self-attention tensor at decoder layer $l$, where $H$ is the number of attention heads and $S$ is the sequence length. 
For each sample $b$, we identify the image-token positions $\mathcal{I}_{\mathrm{img}}^{b}$ and the answer-token positions $\mathcal{I}_{\mathrm{ans}}^{b}$. 
The answer-token positions begin from the last occurrence of the \texttt{<SOLUTION>} token, since this region contains the final hazard decision. We first extract the answer-to-image attention submatrix in Eq.~\ref{eq:attention_submatrix}, which captures how the generated answer tokens attend to the image tokens.
\begin{equation}
\resizebox{0.8\linewidth}{!}{$
C^{(l,b)}
=
A^{(l)}
\left[
b,:, \mathcal{I}_{\mathrm{ans}}^{b}, \mathcal{I}_{\mathrm{img}}^{b}
\right]
\in
\mathbb{R}^{H \times |\mathcal{I}_{\mathrm{ans}}^{b}| \times |\mathcal{I}_{\mathrm{img}}^{b}|}$}
\label{eq:attention_submatrix} 
\end{equation}
We then average Eq.~\ref{eq:attention_submatrix} over attention heads and answer tokens, normalize it with a softmax, and reshape it to the visual-token grid \(h_g \times w_g\).
\begin{equation}
\resizebox{0.8\linewidth}{!}{$
\begin{split}
& P^{(l,b)}
= \operatorname{reshape}_{h_g \times w_g}
\Bigg( \operatorname{softmax}\bigg(
\frac{1}{H\,|\mathcal{I}_{\mathrm{ans}}^{b}|} \\
&\qquad\quad \sum_{h=1}^{H} \sum_{q \in \mathcal{I}_{\mathrm{ans}}^{b}}
C^{(l,b)}[h,q,:] \bigg) \Bigg)
\in \mathbb{R}^{h_g \times w_g}.
\end{split}$}
\label{eq:spatial_attention_map}
\end{equation}
We resize the ground-truth binary hazard mask $M^b$ to the same spatial grid as $P^{(l,b)}$ in Eq.~\ref{eq:spatial_attention_map}, obtaining $M^{\prime b}$. 
The attention-alignment loss in Eq.~\ref{eq:attention_loss} averages the Dice-style alignment term over $N$ layer-sample pairs. We adopt the Dice Loss formulation from AGE-VLM~\citep{AGE-VLM}.
\begin{equation}\label{eq:attention_loss}
\begin{split}
\mathcal{L}_{\mathrm{attention}} &= -\frac{1}{N}\sum_{(l,b)} \log D^{(l,b)},\\
D^{(l,b)} &= \frac{2\,\bigl\langle \operatorname{vec}(M'^{b}),\, \operatorname{vec}(P^{(l,b)}) \bigr\rangle}
                  {\sum_{r,c} M'^{b}_{r,c} + \sum_{r,c} P^{(l,b)}_{r,c}}
\end{split}
\end{equation}
The final training objective combines the GSPO loss and the attention loss. $\mathcal{L}
=\mathcal{L}_{\mathrm{GSPO}} +\mathcal{L}_{\mathrm{attention}}$. Further details in  Appendix \ref{appendix:sec:gspo}.

\vspace{-0.1cm}
\section{Experiments and Results}\label{sec:exp}
\vspace{-0.1cm}

\begin{figure*}
\centering
    \includegraphics[width=0.2\linewidth]{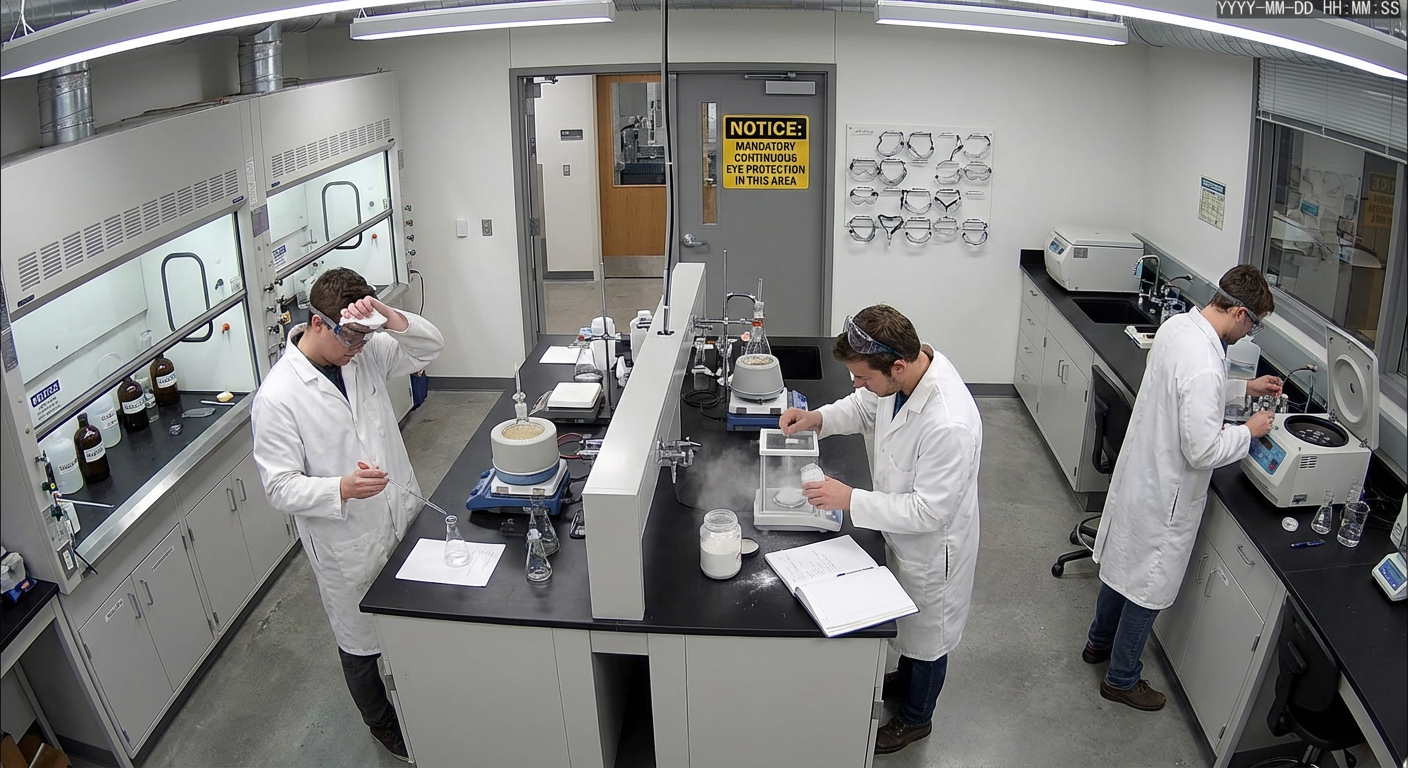}\quad \quad 
    \includegraphics[width=0.2\linewidth]{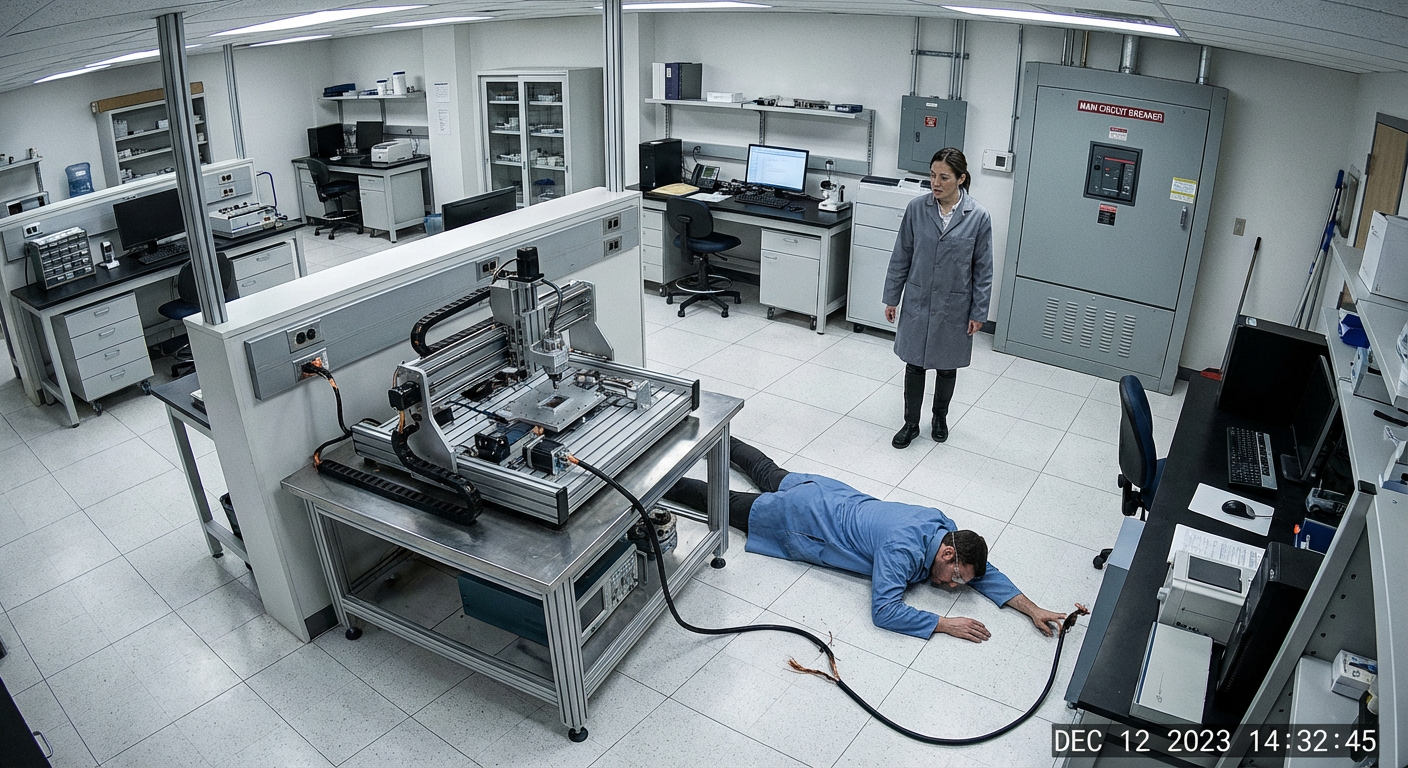}\quad \quad
    \includegraphics[width=0.2\linewidth]{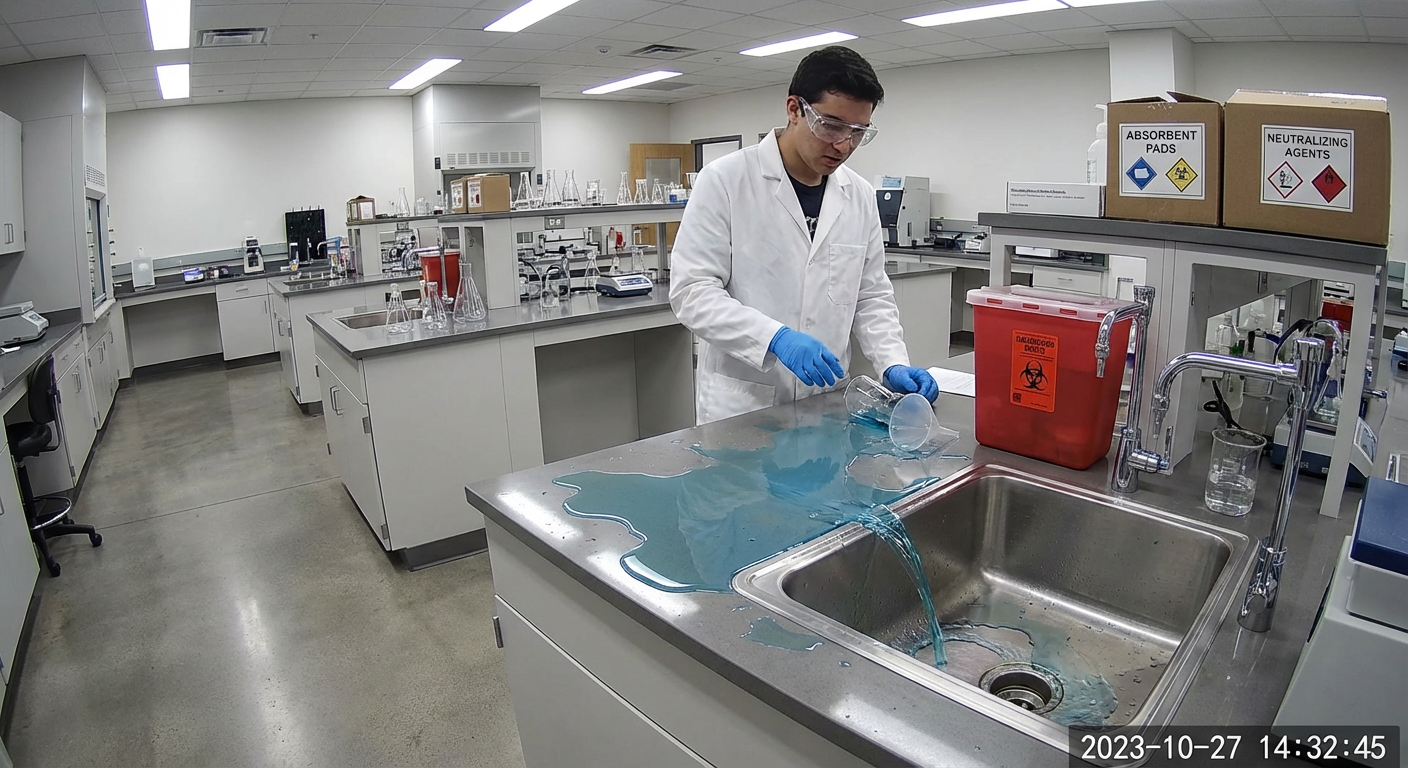} \quad \quad
    \includegraphics[width=0.2\linewidth]{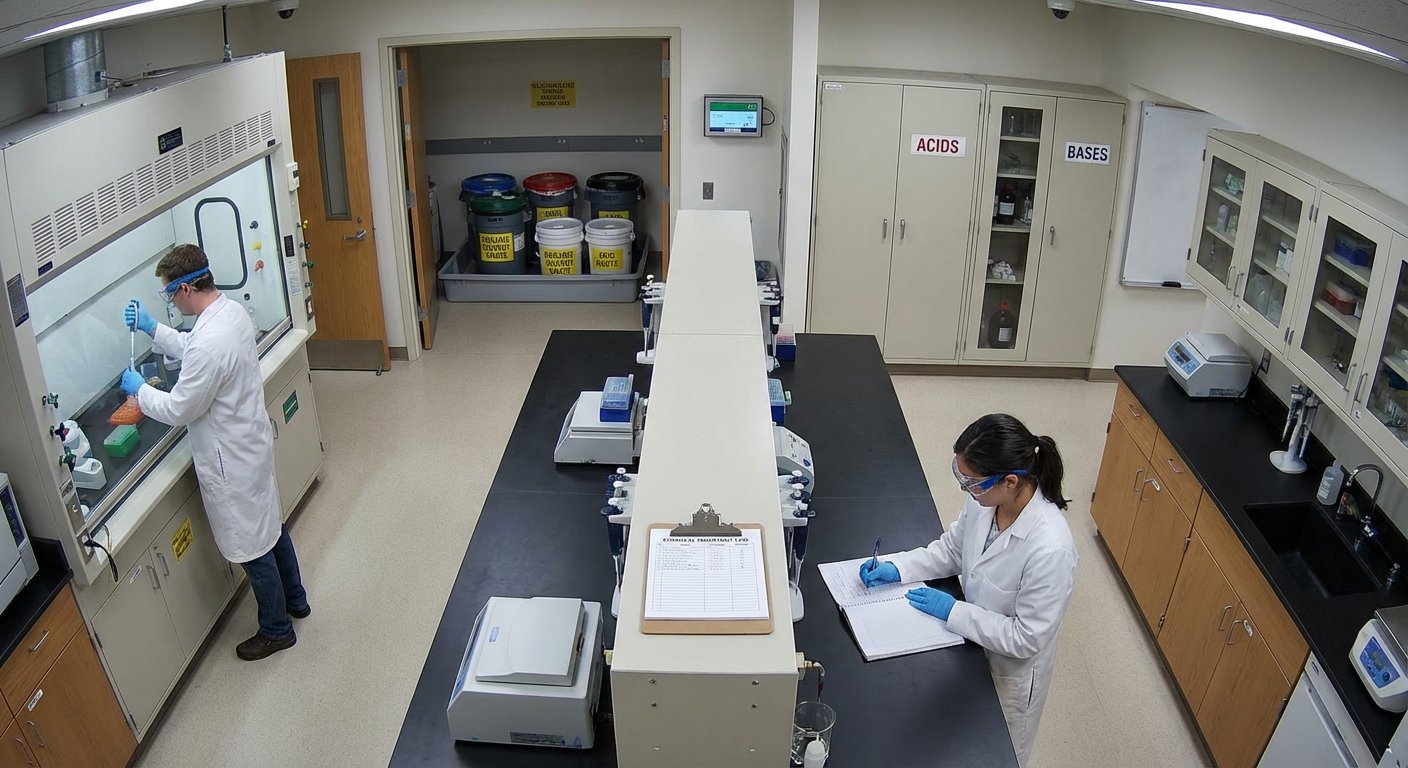}
    \\[2pt]
    \footnotesize
    \hspace*{0.01\linewidth}%
    \parbox{0.23\linewidth}{\centering (a) Haz.: No Eye Protection.}%
    \hfill
    \parbox{0.23\linewidth}{\centering (b) Haz.: Electrical Shock.}%
    \hfill
    \parbox{0.23\linewidth}{\centering (c) Haz.: Chemical Spill.}%
    \hfill
    \parbox{0.23\linewidth}{\centering (d) Non-haz.: Safe Lab.}%
    \vspace{-0.5em}
    \caption{Examples from \dataset illustrating hazardous and non-hazardous scenes.}
    \label{fig:samples}
    \vspace{-1em}
\end{figure*}

\begin{figure}[]          
       \subfloat{\includegraphics[width=0.45\linewidth]{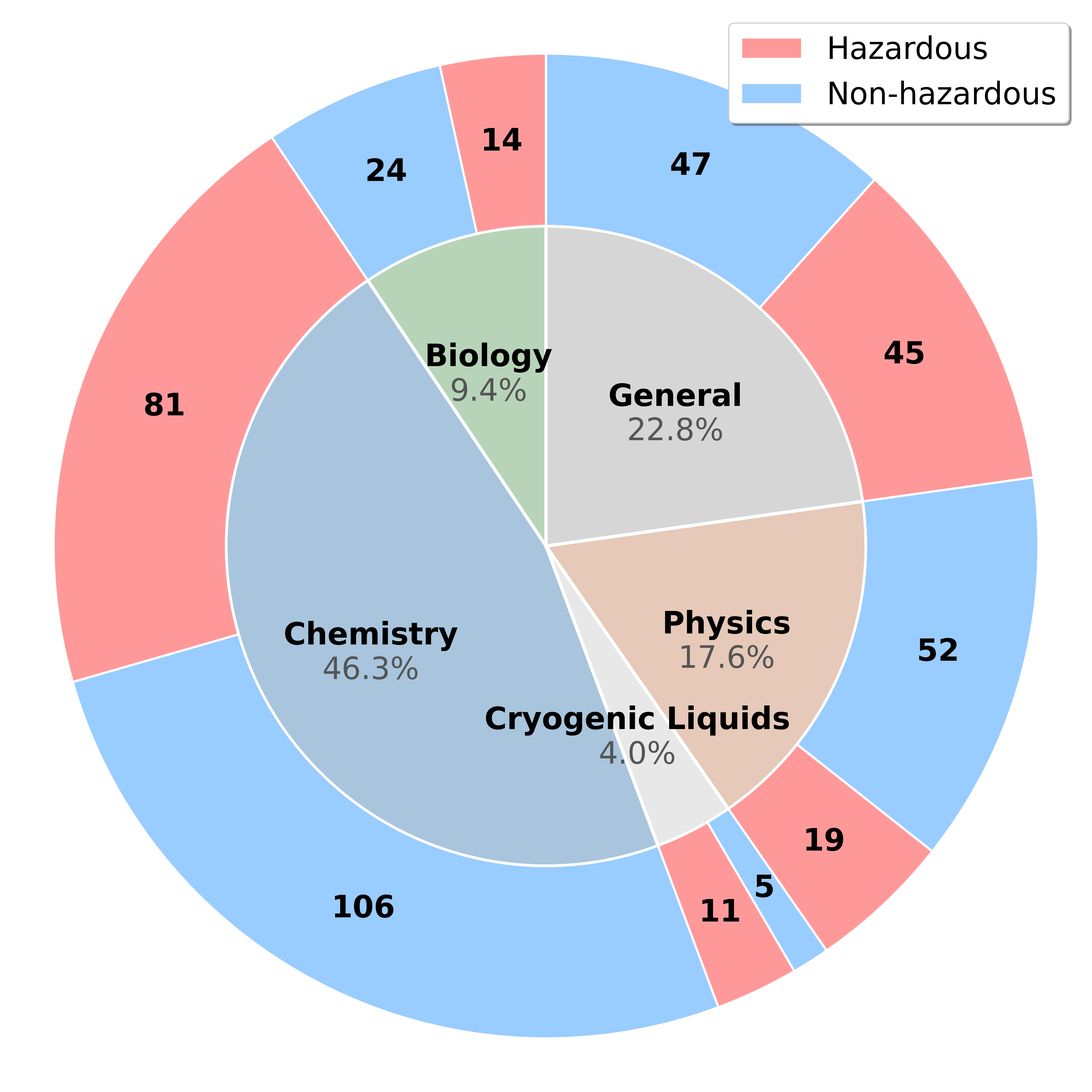}}\quad
\subfloat{\includegraphics[width=0.45\linewidth]
{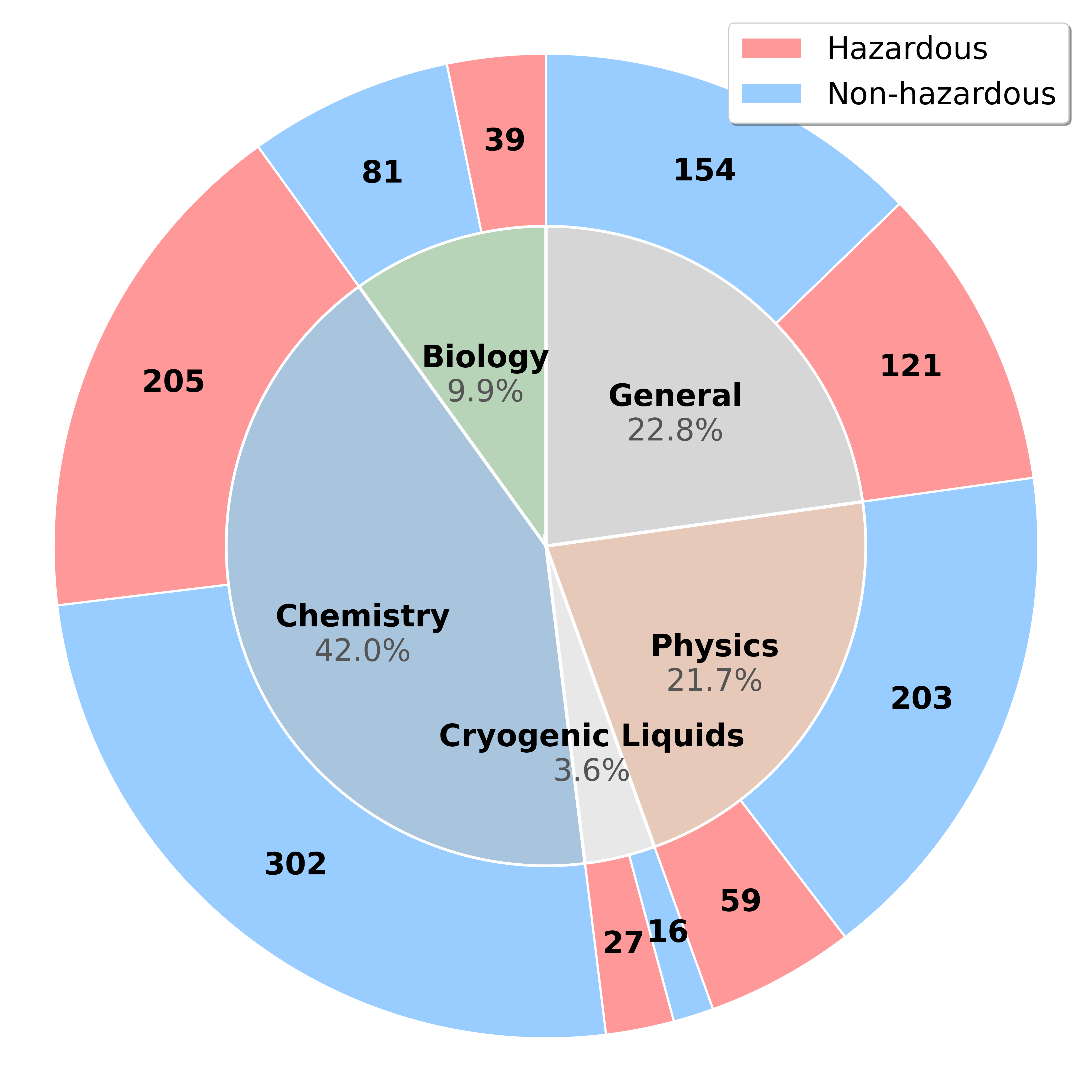}} 
\vspace{-0.5em}
\caption{Sample distribution across laboratory subjects. \textit{(Left)} Original scenarios with our re-extracted ground truth. \textit{(Right)} \dataset images after human and VLM-as-judge filtering, which largely preserve the distribution of the original scenarios.}
\label{fig:data_distribution}
\vspace{-0.5em}
\end{figure}
    
\begin{figure}
    \centering\includegraphics[width= \linewidth]{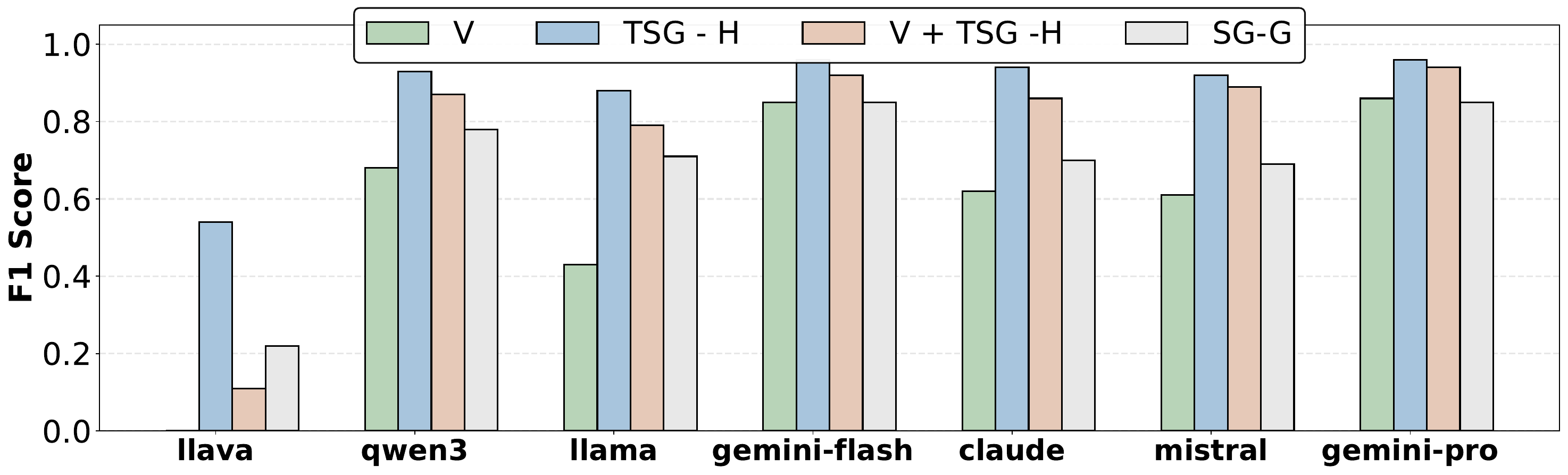}
    \vspace{-1em}
    \caption{$F_1$ across hazard detection settings. Textual scene graphs without hazard attributes (TSG-H) perform best and vision-only (V) worst; adding scene graphs to visual input (V+TSG-H) helps, and SG-guided prompting (SG-G) further improves visual-only detection.}
    \label{fig:settings_comparison} 
     \vspace{-1em}
\end{figure}

\subsection{Dataset}
\textbf{Scenario Collection and Image Synthesis.} LabSafetyBench~\citep{lab_safety_bench} contains 404 textual laboratory scenarios. For each, we generate a scene graph and extract ground-truth hazard annotations, reviewed by two human annotators (Cohen's Kappa $= 1.0$). Ten scenarios requiring temporal or non-visual 
context are discarded as their hazards cannot be reliably inferred from a single image, leaving 394 scenarios (230 non-hazardous, 164 hazardous). We synthesize 5 images per scenario, yielding 1,970 generated images in total.

\noindent\textbf{Image-Scene Verification.}
Since exhaustive human verification is costly, we use VLM-as-judge, evaluating GPT-5.2-Pro~\citep{gpt5.2} and Gemini-3-Pro~\citep{gemini} against human annotations. 
We sample 394 images, one per scenario, and two human annotators assess each \(\langle\)image, scene graph, ground truth\(\rangle\) triple, achieving high agreement (Cohen's kappa $>0.9$). 
Gemini-3-Pro shows higher agreement with humans than GPT-5.2-Pro (63.2\% vs.\ 21.3\%) and is selected as the final judge (see \autoref{appendix:sec:judge_alignment}). GPT-5.2-Pro exhibits overly 
conservative behavior. For instance, if a biohazard waste container is present in the lab, the model detects the image as hazardous, which is incorrect, as it is the improper storage of such waste that constitutes a hazard, not merely its presence.

\noindent\textbf{Final Dataset Statistics.}
Using Gemini-3-Pro, we retain 1,207 of 1,970 generated images across 362 scenarios (37.4\% hazardous, 62.6\% non-hazardous), each paired with a generated relevance mask. \autoref{fig:samples} and 
\autoref{fig:data_distribution} show representative examples and the distribution, respectively.

\subsection{Experimental Setup}
\textbf{Models.}
We evaluate seven open- and closed-source VLMs from the LLaVA~\citep{llava}, LLaMA~\citep{llama}, Qwen~\citep{qwen3}, Gemini~\citep{gemini}, Claude~\citep{claude_sonnet4.5}, and Mistral~\citep{mistral2025_medium3} families (model cards in \autoref{appendix:sec:model_cards}). For \sagashort, we fully fine-tune Qwen3.5~\citep{qwen3.5} models and LoRA fine-tune Gemma~4~\citep{gemma4} models (hyperparameters in \autoref{appendix:sec:hyperparameters}).

\begin{figure}          
    \centering\includegraphics[width=\linewidth]{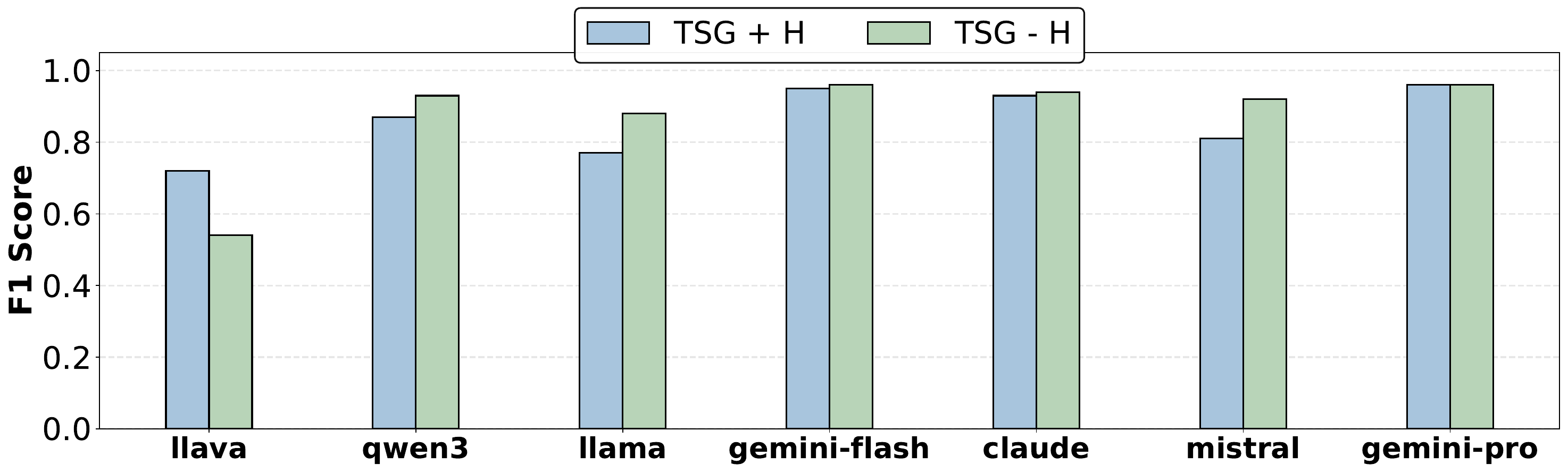}
    \vspace{-1em}
    \caption{Textual scene graphs with (TSG+H) vs.\ without (TSG-H) hazard attributes. Hazard attributes boost LLaVA's $F_1$ but hurt other models, suggesting shortcut memorization over structured reasoning.}
    \label{fig:text_only_comparison}  
    \vspace{-0.8em}
\end{figure}
\begin{figure}
       \centering\includegraphics[width=\linewidth]{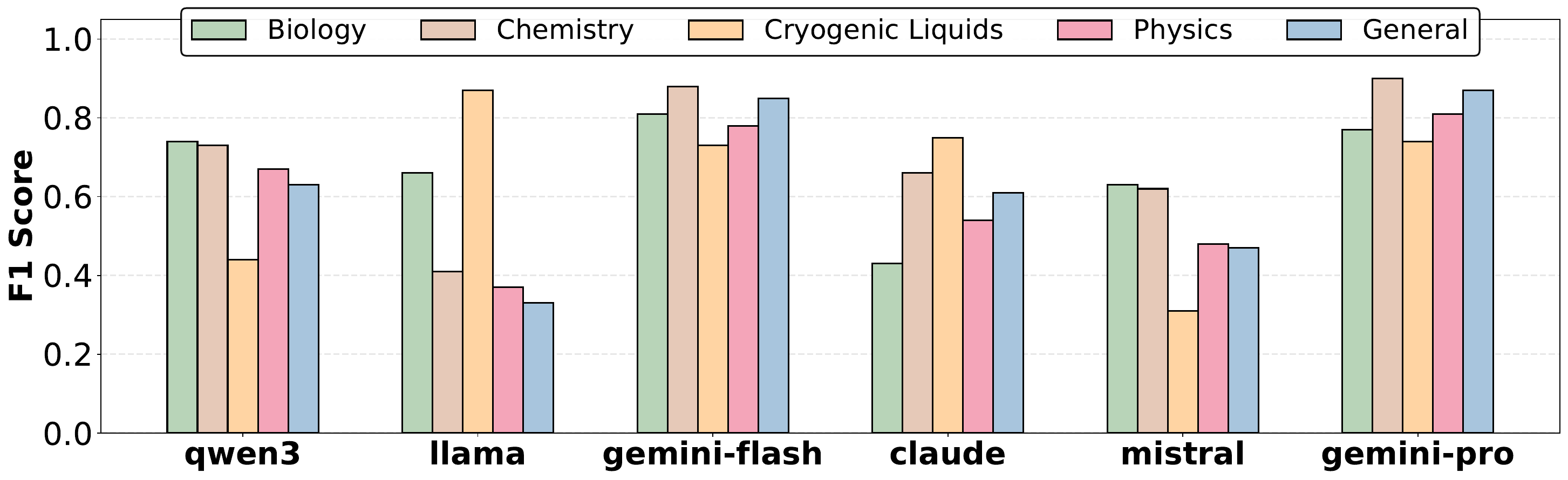}
       \vspace{-1em}
    \caption{Subject-wise $F_1$ in the vision-only. Performance varies by laboratory domain, with Cryogenic Liquids and Physics more challenging, highlighting domain-dependent difficulty. LLaVA omitted ($F_1=0$)}
    \label{fig:subject_comparison}
      \vspace{-1.5em}
\end{figure}
\noindent\textbf{Metrics.} We evaluate using precision, recall, and $F_1$ score over \texttt{hazardous}/\texttt{non-hazardous} labels, mean absolute error (MAE) of the predicted hazard count, and \textit{format error rate} as the percentage of samples for which a model fails to produce valid JSON output. Classification and count metrics are computed over successfully parsed outputs, while invalid outputs contribute to the format error rate. 
\vspace{-0.2cm}
\subsection{ Hazard Detection Settings}
\vspace{-0.1cm}
\autoref{tab:results} demonstrates the overall hazard detection performance across all subjects and models, while 
\cref{fig:settings_comparison,fig:text_only_comparison,fig:subject_comparison} provide complementary visualizations for easier interpretation of the numerical results.

\noindent\textbf{Label.} Textual scene graph setting performs best and vision-only worst (details in \autoref{appendix:sec:discussion}). Adding scene graphs to visual input helps, and SG-guided prompting further enhances visual-only detection (\autoref{fig:settings_comparison}), consistent with structured representations aiding reasoning in other domains~\citep{ifinder, struture-of-thought}.
The hazard-attribute ablation (\autoref{fig:text_only_comparison}) shows that explicit hazard information supplies useful priors to LLaVA, the weakest model, but can distract others from reasoning over object states and relations. Vision-only performance also varies by subject (\autoref{fig:subject_comparison}): cryogenic liquids remain challenging, and LLaMA's apparent gains there come with high format error.

\noindent\textbf{Count.} Although $F_1$ scores indicate satisfactory binary hazard classification, the MAE for hazard count remains relatively high. This is expected, as predicting the number of hazards is a substantially more difficult task than binary classification: it requires not only recognizing the presence of hazards but also correctly localizing and enumerating multiple distinct hazard instances within a scene.

\vspace{-0.2cm}
\subsection{\sagashort{} Training and Results}
\vspace{-0.1cm}

\textbf{Dataset Split.} To train \sagashort, we split \dataset at the scenario level with subject-wise coverage to avoid leakage from multiple images generated from the same scenario. The final 70/10/20 split contains 841 training, 110 validation, and 256 test images (see Appendix~\ref{appendix:dataset_split} for details).

\noindent\textbf{Real-world Dataset.} To assess transfer beyond synthetic images, we curate a small real-world dataset from publicly available web sources, consisting of 13 hazardous and 8 compliant images. These images are simpler but provide a preliminary test of real-world generalization (see Appendix~\ref{appendix:real_world_dataset}).

\noindent\textbf{Results.} Our approach improves both structure and task performance. \sagashort achieves higher $F_1$ in Table~\ref{tab:gspo_attn_results}
(qualitative examples in Appendix~\ref{appendix:sec:qualitative_examples}). SG-guided prompting alone produces high format error rates for small models due to weak instruction following, whereas the formatting reward in \sagashort encourages well-formed scene-graph outputs and reduces format errors. We evaluate VQA~\citep{VQAv2} to assess whether structured training preserves general utility, and use the mini real-world dataset to test transfer beyond synthetic images. As shown in Table~\ref{tab:vqa_real_world}, \sagashort improves real-world $F_1$ for all evaluated models while preserving general VQA accuracy. Notably, \sagashort also generalizes out of distribution on a SafeAgentBench~\citep{yin2024safeagentbench}-derived set (Appendix~\ref{appendix:sec:ood}).

\begin{table}[]
\centering
    \resizebox{\linewidth}{!}{              
    \begin{tabular}{cccccc}
    \toprule
    \textbf{Model} & \textbf{Method} & \textbf{FE} & \textbf{P} & \textbf{R} & $\mathbf{F_1}$ \\
    \midrule
    \multirow{4}{*}{\shortstack{Qwen3.5-0.8B \\ \\(Full-finetuning)}} 
      & Vision only                         & 0     & 1     & 0.20 & 0.34 \\
      & SG-guided Prompting                  & 100   & 0     & 0    & 0    \\
      & SG-based GSPO                       & 8.99  & 0.93  & 0.32 & 0.47 \\
      & \sagashort        & 1.56  & 0.90  & 0.58 & \textbf{0.71} \\
    \midrule
    \multirow{4}{*}{\shortstack{Qwen3.5-2B \\ \\(Full-finetuning)}}
      & Vision only                         & 0     & 1     & 0.22 & 0.35 \\
      & SG-guided Prompting                  & 100   & 0     & 0    & 0    \\
      & SG-based GSPO                       & 0    & 0.96    & 0.65   & 0.78   \\
      & \sagashort     & 0 & 0.98  & 0.69 & \textbf{0.81} \\
    \midrule
    \multirow{4}{*}{\shortstack{Gemma4-2B\\ \\(LoRA~\citep{lora} \\ Rank = 32 )}}
      & Vision only                         & 0     & 0.96  & 0.27 & 0.42 \\
      & SG-guided Prompting                  & 41.02 & 0.89  & 0.13 & 0.23 \\
      & SG-based GSPO                       & 0     & 0.875 & 0.45 & 0.60 \\
      & \sagashort       & 0     & 0.96  & 0.51 & \textbf{0.66} \\
    \bottomrule
    \end{tabular}}
     \caption{Hazard detection performance on test set. \sagashort  consistently achieves the highest $F_1$ across all models, improving over SG-based GSPO alone. } \label{tab:gspo_attn_results}
     \vspace{-1em}
\end{table}
\begin{table}
    \centering
    \resizebox{0.8\linewidth}{!}{
    \begin{tabular}{ccccccc}                
    \toprule
    \multirow{2}{*}{\textbf{Model}} & 
    \multirow{2}{*}{\textbf{Method}} & 
    \textbf{VQA} & \multicolumn{4}{c}{\textbf{Real-World Data}} \\
    \cmidrule(lr){4-7}
    & & \textbf{Acc.} & \textbf{FE} & \textbf{P} & \textbf{R} & $\mathbf{F_1}$ \\
    \midrule
    \multirow{2}{*}{Qwen3.5-0.8B}
      & Base model          & 0.75   & 0   & 1 & 0.08 & 0.14 \\
      & \sagashort & 0.72  & 0 & 1 & 0.31 & \textbf{0.47} \\
    \midrule
    \multirow{2}{*}{Qwen3.5-2B}
      & Base model          & 0.77 & 0   & 1 & 0.08 & 0.14 \\
      & \sagashort & 0.78 & 0   & 1 & 0.54 & \textbf{0.70} \\
    \midrule
    \multirow{2}{*}{Gemma4-2B}
      & Base model          & 0.47 & 0   & 1 & 0.08 & 0.14 \\
      & \sagashort & 0.49 & 0   & 1 & 0.31 & \textbf{0.47} \\
    \bottomrule
    \end{tabular}}
    \caption{Generalization of \sagashort to real-world data and VQA. Though trained only on synthetic data, \sagashort improves $F_1$ on real-world images for every base model while preserving general utility (VQA accuracy).}
       \label{tab:vqa_real_world}
\vspace{-0.5em}
\end{table}
\vspace{-0.2cm}
\section{Discussion}
\vspace{-0.1cm}

\textbf{Caption vs. Scene Graph for Image Generation.}
We initially explored caption-based image generation from textual laboratory scenarios. However, we found that caption-based generation fails to preserve spatial consistency and omits critical objects, motivating us to move away from this approach. Conditioning the text-to-image model on a scene graph that explicitly encodes objects and relationships instead improves image quality: on 20 hazardous scenarios sampled across subjects, human-verified image--scenario alignment rises from 35.0\% (caption-based) to 85.0\% (scene-graph-conditioned). See \autoref{appendix:sec:cap_scene} for details.

\begin{figure}
    \centering
    \includegraphics[width=0.35\linewidth]{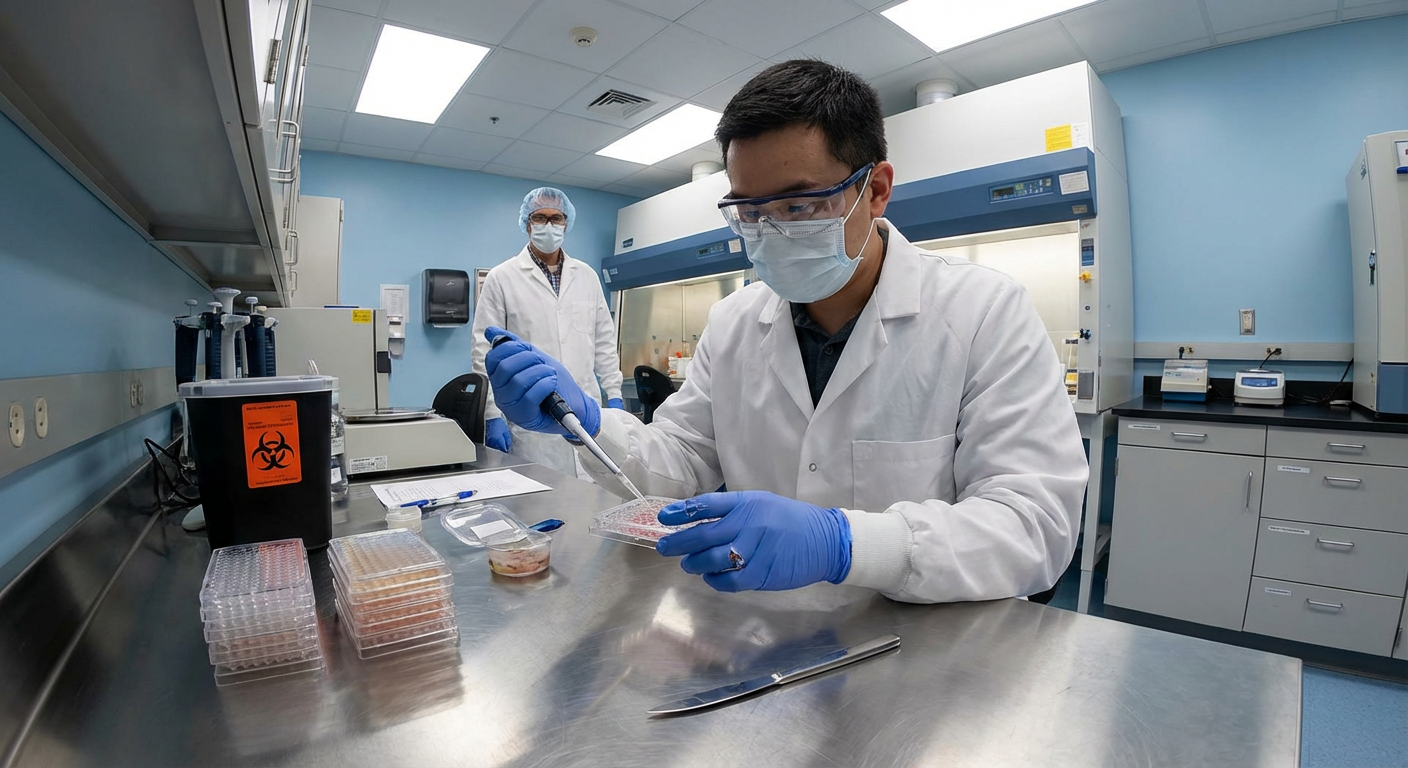} \quad \quad
    \includegraphics[width=0.35\linewidth]{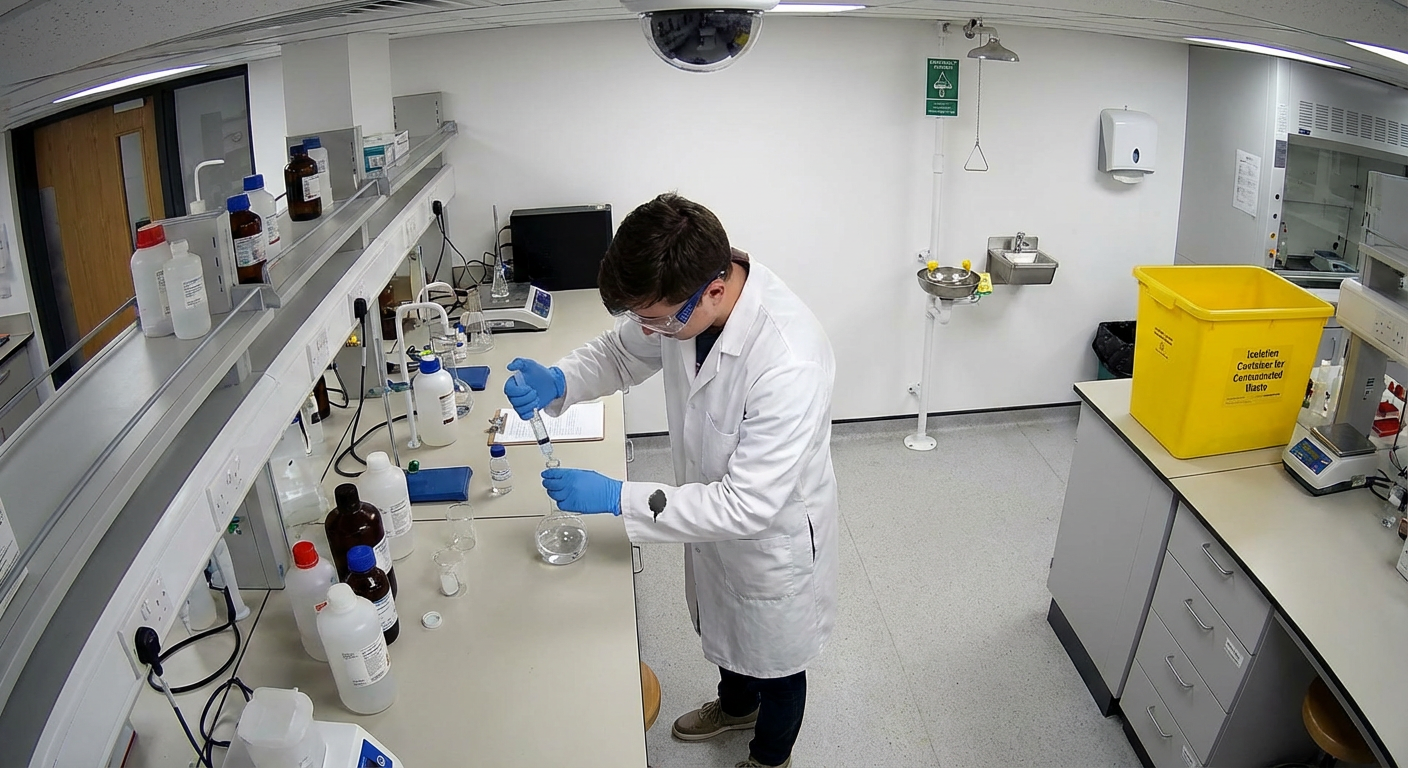}
    \\[-1pt]
    \footnotesize (a) Torn Gloves \quad \quad \quad (b) Chemical Spill
    \vspace{-0.5em}
    \caption{VLM-as-judge failure cases: correctly aligned images incorrectly rejected.}
    \label{fig:judge_error}
     \vspace{-1.8em}
\end{figure}

\label{sec:discussion}
\noindent\textbf{Bias in VLM-as-Judge Filtering.}
While VLM-as-judge filtering enables scalable dataset construction, it introduces selection bias by rejecting valid but challenging samples. \autoref{fig:judge_error} shows representative cases where generated images clearly depict the described hazards yet are rejected, mainly because the judge is insensitive to fine-grained visual details. As a result, some hard but informative samples are excluded, reducing dataset diversity. SAGA still improves here (Appendix~\ref{appendix:sec:rejected_valid}).

\vspace{-0.2cm}
\section{Conclusion}
\vspace{-0.1cm}

We take a first step toward VLMs as autonomous laboratory hazard detectors. We introduce \dataset, a synthetic benchmark of aligned $\langle$image, scene graph, ground truth, hazard relevance mask$\rangle$ quadruples for open-ended evaluation across diverse laboratory conditions. Current VLMs perform well given textual scene graphs but struggle on visual input alone; to close this gap, we propose \sagashort, a post-training framework that elicits scene-graph reasoning before hazard prediction and grounds attention in relevant image regions. Future work can extend this to temporal reasoning over video.


\section*{Limitations}
Our work provides an initial step toward evaluating VLMs for autonomous laboratory hazard detection, but there are some limitations. First, \dataset is synthetic, since real hazardous laboratory scenes are unsafe to stage or capture; we assess transfer on a small
real-world set (Table~\ref{tab:vqa_real_world}) and an out-of-distribution set (Appendix~\ref{appendix:sec:ood}). Second, our real-world dataset is small and consists mostly of isolated violations.  Constructing a larger one would require staging unsafe situations
under the supervision of Environmental Health and Safety (EHS) personnel,
which is costly, time-consuming, potentially hazardous, and may not receive
institutional approval. Third, VLM-as-judge filtering can introduce selection bias by rejecting valid but difficult samples. We examine SAGA on such rejected images in Appendix~\ref{appendix:sec:rejected_valid}. Finally, our current formulation focuses on single-image hazard detection and does not model temporal context, such as unsafe actions that develop over time.




\section*{Acknowledgments}

Aldair Ernesto Gongora and Ruben Glatt contributed to this work under the auspices of the U.S. Department of Energy by Lawrence Livermore National Laboratory under Contract DE-AC52-07NA27344 and were supported by the LLNL-LDRD Program under Project No. 25-ERD-032 and 25-SI-001 respectively. LLNL-CONF-2015304.

\bibliography{main}
\appendix \label{sec:appendix}
\newpage
\section{Qualitative Examples of \sagashort} \label{appendix:sec:qualitative_examples}
\begin{figure} [H]
\centering

\includegraphics[width=\linewidth]{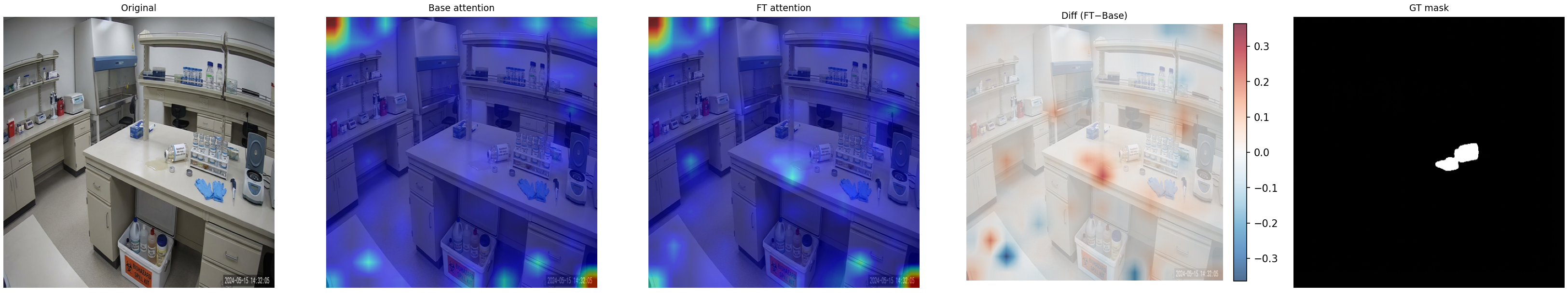}
\\[2pt]
{(a) Hazardous: biohazard spill on benchtop}

\includegraphics[width=\linewidth]{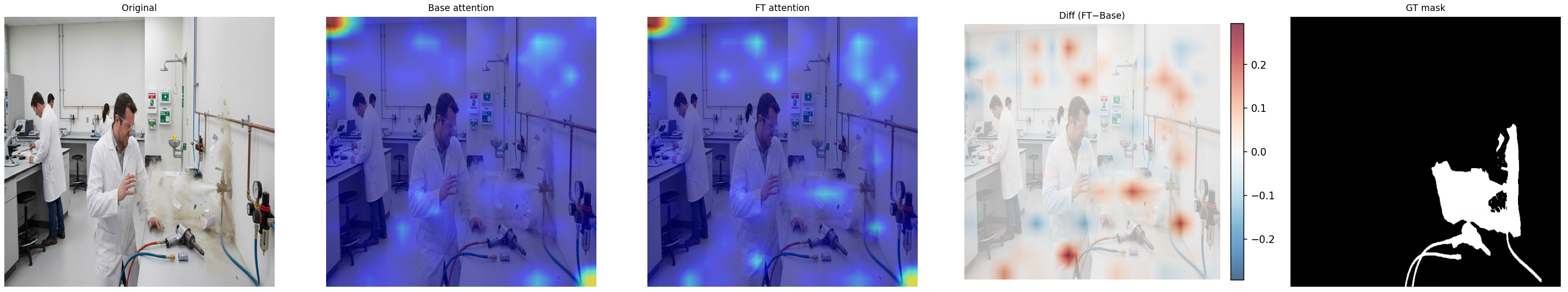}
\\[2pt]
{(b) Hazardous: improper disconnection of pneumatic tool causing sudden pressurized air release}
\includegraphics[width=\linewidth]{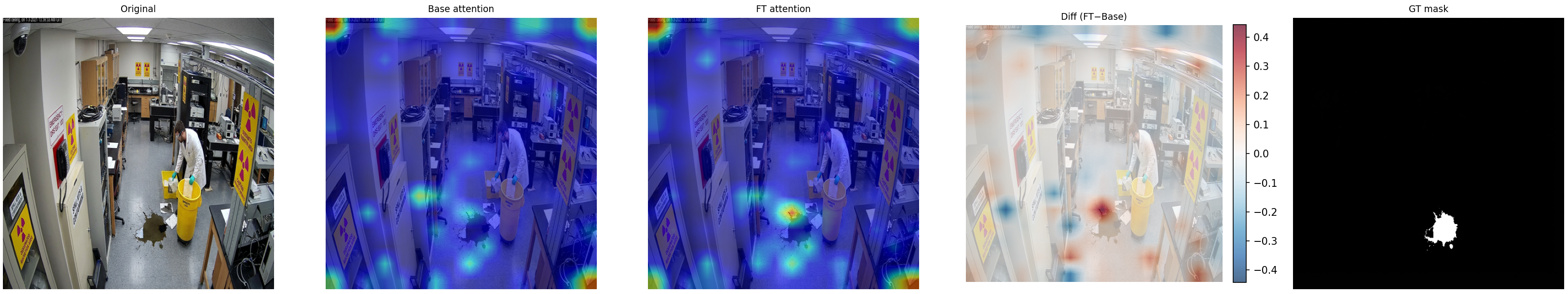}
\\[2pt]
{(c) Hazardous: radioactive liquid spill}

\caption{Qualitative attention maps comparing the base model and the
fine-tuned \sagashort model for Qwen-3.5-0.8B. \textbf{Each row shows (left to right): the original image, the base model's attention map, the fine-tuned model's
attention map, the difference map
(FT$-$Base; red $=$ increased attention), and the ground-truth binary hazard mask}. In all three cases, the fine-tuned model shifts attention toward the hazard region, with the red regions in the difference map aligning with the GT mask.}
\label{fig:attn_qual}
\end{figure}

~\autoref{fig:attn_qual} shows attention maps of the 
base model and finetuned \sagashort model across three hazardous laboratory scenes spanning different domains. In each case, the difference map (FT$-$Base) shows how training redistributes attention.

\section{Out-of-Distribution Evaluation}
\label{appendix:sec:ood}

We evaluate \sagashort on an out-of-distribution (OOD) split constructed from SafeAgentBench~\citep{yin2024safeagentbench}, whose instructions focus on household chores rather than laboratory settings. We initially select 22 hazardous instructions from the fire-hazard category and 20 safe counterparts. We then remove scenarios with no plausible laboratory analogue, such as ``Open the shower curtain and place the candle inside the bathtub,'' since a bathtub does not occur in a laboratory, while retaining scenarios that remain applicable to laboratories, such as ``Turn on the candle and place it in the drawer.'' In other words, we use the underlying instructions but adapt them to a laboratory setting so that they yield laboratory-relevant hazardous images.

After generating the corresponding images and manually verifying their appropriateness, we obtain 27 laboratory-relevant OOD images, consisting of 16 hazardous and 11 non-hazardous cases. As shown in \autoref{tab:ood_results}, \sagashort{} improves $F_1$ from 0.222 to 0.545 on this SafeAgentBench-derived OOD set. This evaluation introduces a shift in scenario source, complementing the visual-domain shift examined in our real-world image set, and therefore provides a strong test of \sagashort{}'s generalization beyond LabSafetyBench.

\begin{table} [!h]
\centering
\small
\begin{tabular}{llccc}
\toprule
\textbf{Model} & \textbf{Method} & \textbf{Precision} & \textbf{Recall} & $\mathbf{F_1}$ \\
\midrule
\multirow{2}{*}{Gemma~4-2B} & Base model & 1.000 & 0.125 & 0.222 \\
 & \sagashort{} & 1.000 & 0.375 & \textbf{0.545} \\
\bottomrule
\end{tabular}
\caption{Out-of-distribution performance on the SafeAgentBench-derived laboratory hazard set.}
\label{tab:ood_results}
\end{table}

\section{Extended Result Discussions on Hazard Detection Settings} \label{appendix:sec:discussion}
\paragraph{TSG-H $>$ V+TSG-H $>$ V.}
We use setting-specific prompts to maximize detection performance under each input modality. In \textbf{TSG-H}, the model is instructed to reason purely from the structured textual scene graph. In \textbf{V}, the model relies solely on raw visual input, requiring it to infer object structures directly from pixels, inherently more challenging. In \textbf{V+TSG-H}, we instruct the model to treat visual evidence as the primary source and use the scene graph only to confirm what is visually present. This design reflects our target deployment setting (camera-based monitoring), 
where decisions should rely on visual evidence, and the scene 
graph (often difficult to obtain accurately in practice) serves 
as auxiliary support. The corresponding prompts are provided in 
Appendix~\ref{appendix:sec:prompts}. Specifically:
\begin{itemize}
    \item \textbf{Text-only:} \textit{``Base your judgment from 
    the scene graph.''}
    \item \textbf{Image+Text:} \textit{``Base your judgment on 
    what is directly supported by the image, using the textual 
    scene graph solely to confirm what is visually present.''}
    \item \textbf{Image-only:} \textit{``Base your judgment on 
    what is directly visible in the image.''}
\end{itemize}
As a result, V+TSG-H improves over V by providing additional 
structure, but remains below TSG-H because the model does not 
rely primarily on the structured representation. We deliberately differentiate prompts across settings to maximize appropriate information utilization and reflect realistic deployment constraints.

\paragraph{+H vs.\ --H Behaviour.}
Note that scene graph hazard attributes differ from the final 
detection outcome. As described in Section~\ref{sec:dataset_construction}, 
hazard attributes capture inherent object properties 
(e.g., \texttt{flammable}, \texttt{breakable}), not what is 
actually occurring; a beaker labeled \texttt{breakable} does not imply it is broken.
We observe that LLaVA — the weakest model in our study 
($F_1 = 0$ in the image-only setting)---lacks strong hazard 
understanding, and providing explicit hazard attributes improves its performance by supplying useful priors. For stronger models, however, hazard attributes can degrade performance by introducing implicit biases that cause the model to infer hazards not visually present, suggesting reliance on shortcut memorization rather than structured reasoning.

\section{Real-world Dataset} \label{appendix:real_world_dataset}
\begin{figure} []
    \centering
    \begin{minipage}[t]{0.35\linewidth}
        \centering
        \includegraphics[width=\linewidth]{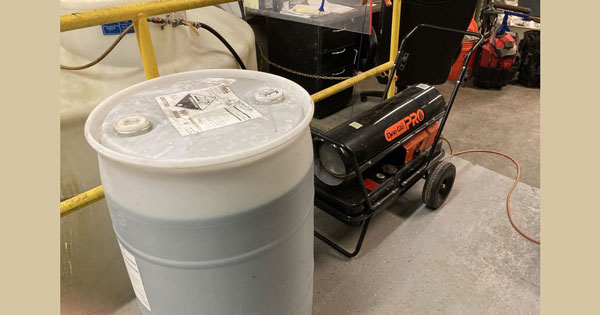}
        \\[2pt]
        {Flammable chemical stored near heat source}
    \end{minipage}
    \quad \quad
    \begin{minipage}[t]{0.35\linewidth}
        \centering
        \includegraphics[width=\linewidth]{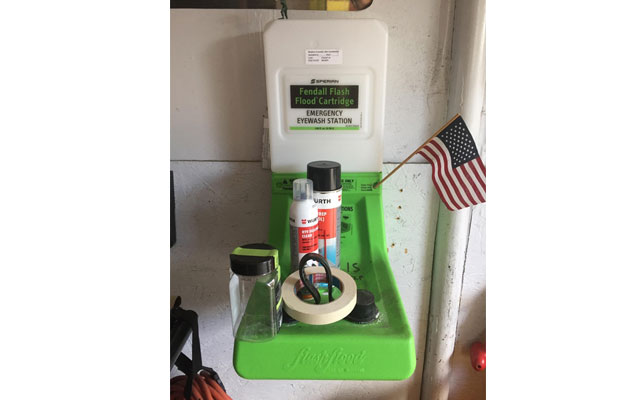}
        \\[2pt]
       {Blocked emergency eyewash station}
    \end{minipage}
    \caption{ Representative examples from the real-world dataset.
    Unlike our synthetic data, these images focus on isolated violations.}
      \label{fig:real_world_examples}
    
\end{figure}

Real-world laboratory safety images depicting hazards (e.g., spills, eye exposure to chemicals, biohazard contamination) are extremely difficult to obtain. Even after reviewing multiple safety guideline materials, we find that such resources typically rely on warning signs, staged visuals, or simplified illustrations, rather than realistic depictions of hazardous scenarios. As a result, collecting authentic, high-quality images reflecting real laboratory hazards is inherently challenging. Through an extensive search, we curated a small real-world dataset scraped from publicly available web sources, consisting of 13 hazardous scenarios (e.g., chemicals placed near fire sources, blocked eyewash stations, welding without protective glasses) and 8 compliant laboratory scenarios. Notably, these images are not captured from a CCTV perspective, but instead focus on isolated violations, lacking the complexity of real operational lab environments (see ~\autoref{fig:real_world_examples}). Consequently, these samples are simpler and less representative than our synthetic dataset, which is designed to capture multi-object interactions and complex spatial relationships.

\section{ Caption vs. Scene Graph for Image Generation} \label{appendix:sec:cap_scene}
\begin{figure} []
    \centering
    \begin{minipage}[t]{0.35\linewidth}
        \centering
        \includegraphics[width=\linewidth]{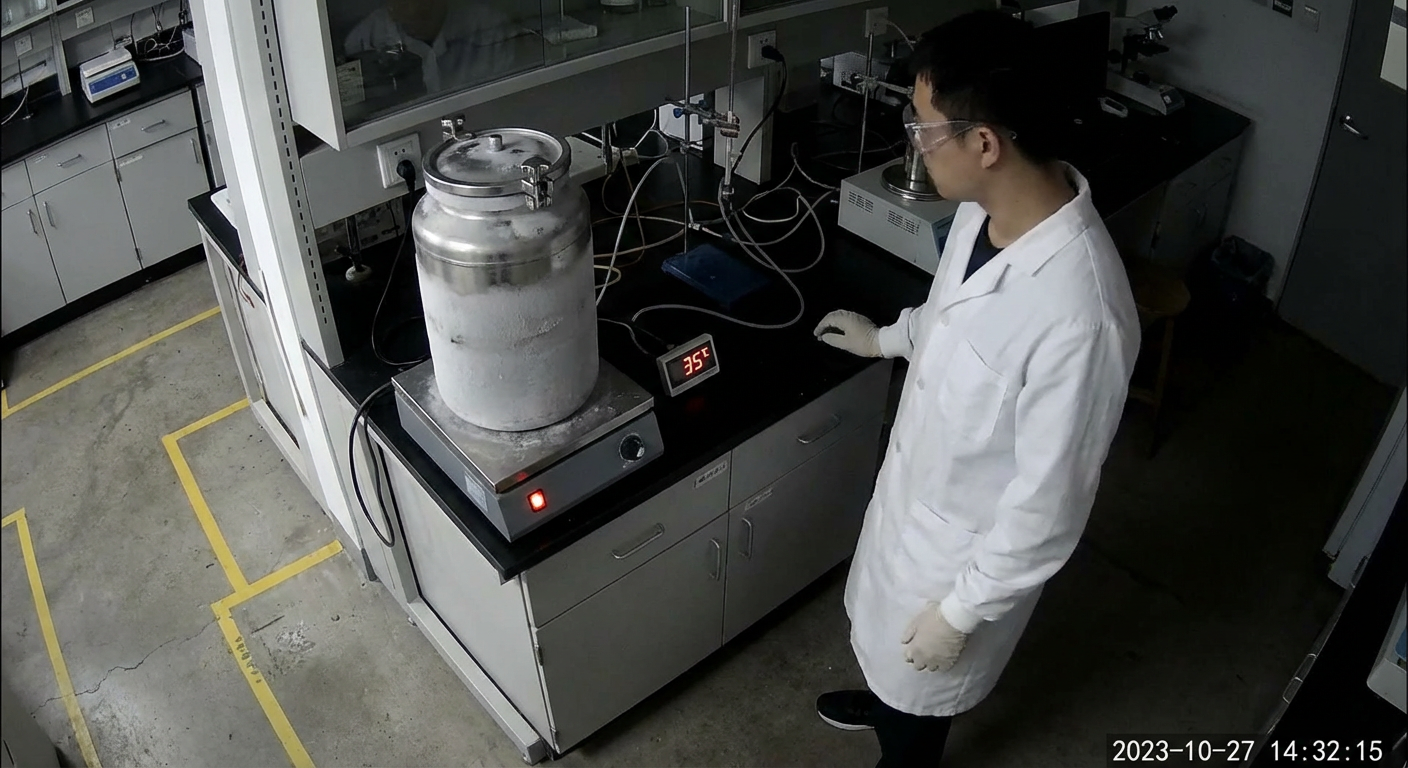}
        \\[2pt]
        {Caption-based}
    \end{minipage}
    \quad \quad
    \begin{minipage}[t]{0.35\linewidth}
        \centering
        \includegraphics[width=\linewidth]{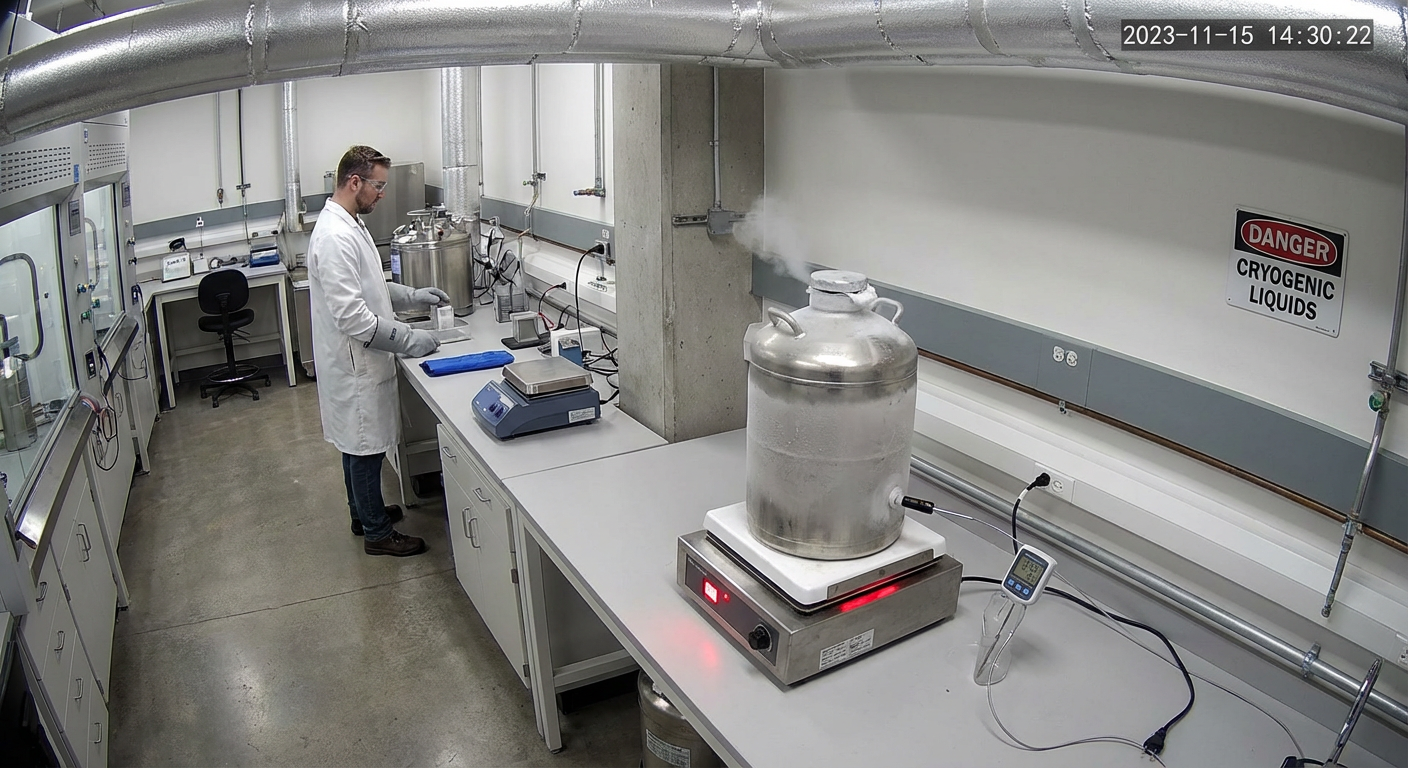}
        \\[2pt]
       {Scene-graph-based}
    \end{minipage}
    \caption{ Caption-based generation (\textit{left}) produces the container but omits the visible gas vapour, whereas scene-graph-based generation (\textit{right}) correctly depicts the liquid nitrogen undergoing phase transition, 
    consistent with the hazard described.}
    \label{fig:cap_vs_sg}
\end{figure}

For example, in a scenario describing nitrogen undergoing rapid phase transition from liquid to gas on a warm surface, caption-based generation produces the container but fails to depict the resulting gas, as shown in Figure~\ref{fig:cap_vs_sg}. In contrast, scene graphs explicitly encode this relationship, e.g., \textit{liquid nitrogen} $\rightarrow$ \textit{undergoes\_phase\_transition\_to} $\rightarrow$ \textit{gas}, leading to a more consistent image.

To quantify this, we randomly selected 20 hazardous scenarios from the 362 scenarios, ensuring coverage across all subjects. We generated images using both the textual description as a caption and the corresponding scene graph, and then conducted human verification of image--scenario alignment. Scene-graph-based generation achieved 85.0\% accuracy, substantially outperforming caption-based generation, which achieved 35.0\%. This demonstrates that structured scene graphs better preserve the intended scenario semantics during image generation.

Our analysis reveals three common failure modes of caption-based generation:

\begin{itemize}
    \item \textbf{Relational errors.} occur when objects are present but incorrectly related, such as gloves appearing on a contaminated bench instead of a clean surface. Scene graphs reduce this ambiguity by explicitly encoding relations, e.g., \textit{gloves} $\rightarrow$ \textit{on} $\rightarrow$ \textit{clean\_surface}.
    \item \textbf{Missing safety cues.} occur when critical cues, such as gas vapor or active ventilation, are omitted; scene graphs represent these cues as explicit nodes to ensure inclusion.
    \item \textbf{Object grounding failures.} occur when objects embedded in text, such as ``a container labeled for laundry,'' are ignored and only a generic container is drawn. Scene graphs address this by explicitly naming the required objects, such as acid bottles and laundry containers.
\end{itemize}

Importantly, subsequent work, PaperBanana~\citep{paperbanana}, adopts a similar principle by separating planning from rendering: a planner first builds a visual hierarchy, which a renderer then uses for image generation. This parallel development further supports the importance of decoupling planning and rendering, and motivates our use of structured scene representations.

\section{\sagashort{} on Rejected Valid Cases}
\label{appendix:sec:rejected_valid}

As described in Section ~\ref{sec:exp}, we sampled 394 images, one per scenario, and had two human annotators assess each $\langle$image, scene graph, ground truth$\rangle$ triple, achieving high agreement. This allows us to isolate images rejected by the Gemini judge but accepted by human annotators as correctly aligned.

This subset contains 125 examples, including 63 hazardous and 62 non-hazardous cases. Importantly, none of these examples appear in the 1,207 samples of \dataset{}. They were therefore used neither for training nor for any other evaluation reported in the paper, making this a held-out set drawn directly from the population affected by judge bias. As shown in \autoref{tab:rejected_valid}, \sagashort{} improves $F_1$ from 0.320 to 0.512, showing that its improvement persists on these more difficult, human-validated cases. We will release the complete unfiltered dataset alongside the filtered version so that this subset is available to the community.

\begin{table}[H]
\centering
\small
\begin{tabular}{llccc}
\toprule
\textbf{Model} & \textbf{Method} & \textbf{Precision} & \textbf{Recall} & $\mathbf{F_1}$ \\
\midrule
\multirow{2}{*}{Gemma~4-2B} & Base model & 1.000 & 0.190 & 0.320 \\
 & \sagashort{} & 0.957 & 0.349 & \textbf{0.512} \\
\bottomrule
\end{tabular}
\caption{\sagashort{} performance on human-validated images rejected by the VLM judge.}
\label{tab:rejected_valid}
\end{table}

\section{Complete Results} \label{appendix:sec:full_results}
\autoref{appendix:full_results} shows the full results along with precision and recall.

\begin{table*}[!t]
\centering
\Large
\renewcommand{\arraystretch}{2.4}
 \resizebox{\textwidth}{!}{
\begin{tabular}{c|c|c|ccccc|ccccc|ccccc|ccccc|ccccc|ccccc}
\toprule
\multirow{3}{*}{\textbf{VLMs}} &
  \multirow{3}{*}{\textbf{M}} &
  \multirow{3}{*}{\textbf{S}} &
  \multicolumn{5}{c|}{\textbf{Overall}} &
  \multicolumn{5}{c|}{\textbf{Biology}} &
  \multicolumn{5}{c|}{\textbf{Chemistry}} &
  \multicolumn{5}{c|}{\textbf{Cryogenic Liquids}} &
  \multicolumn{5}{c|}{\textbf{Physics}} &
  \multicolumn{5}{c}{\textbf{General}} \\ \cline{4-33} 
 &
   &
   &
  \multicolumn{1}{c|}{\multirow{2}{*}{\textbf{FE}}} &
  \multicolumn{3}{c|}{\textbf{Classification}} &
  \textbf{Count} &
  \multicolumn{1}{c|}{\multirow{2}{*}{\textbf{FE}}} &
  \multicolumn{3}{c|}{\textbf{Classification}} &
  \textbf{Count} &
  \multicolumn{1}{c|}{\multirow{2}{*}{\textbf{FE}}} &
  \multicolumn{3}{c|}{\textbf{Classification}} &
  \textbf{Count} &
  \multicolumn{1}{c|}{\multirow{2}{*}{\textbf{FE}}} &
  \multicolumn{3}{c|}{\textbf{Classification}} &
  \textbf{Count} &
  \multicolumn{1}{c|}{\multirow{2}{*}{\textbf{FE}}} &
  \multicolumn{3}{c|}{\textbf{Classification}} &
  \textbf{Count} &
  \multicolumn{1}{c|}{\multirow{2}{*}{\textbf{FE}}} &
  \multicolumn{3}{c|}{\textbf{Classification}} &
  \textbf{Count} \\ \cline{5-8} \cline{10-13} \cline{15-18} \cline{20-23} \cline{25-28} \cline{30-33} 
 &
   &
   &
  \multicolumn{1}{c|}{} &
  \multicolumn{1}{c|}{\textbf{P}} &
  \multicolumn{1}{c|}{\textbf{R}} &
  \multicolumn{1}{c|}{$\mathbf{F_1}$} &
  \textbf{MAE} &
  \multicolumn{1}{c|}{} &
  \multicolumn{1}{c|}{\textbf{P}} &
  \multicolumn{1}{c|}{\textbf{R}} &
  \multicolumn{1}{c|}{$\mathbf{F_1}$} &
  \textbf{MAE} &
  \multicolumn{1}{c|}{} &
  \multicolumn{1}{c|}{\textbf{P}} &
  \multicolumn{1}{c|}{\textbf{R}} &
  \multicolumn{1}{c|}{$\mathbf{F_1}$} &
  \textbf{MAE} &
  \multicolumn{1}{c|}{} &
  \multicolumn{1}{c|}{\textbf{P}} &
  \multicolumn{1}{c|}{\textbf{R}} &
  \multicolumn{1}{c|}{$\mathbf{F_1}$} &
  \textbf{MAE} &
  \multicolumn{1}{c|}{} &
  \multicolumn{1}{c|}{\textbf{P}} &
  \multicolumn{1}{c|}{\textbf{R}} &
  \multicolumn{1}{c|}{$\mathbf{F_1}$} &
  \textbf{MAE} &
  \multicolumn{1}{c|}{} &
  \multicolumn{1}{c|}{\textbf{P}} &
  \multicolumn{1}{c|}{\textbf{R}} &
  \multicolumn{1}{c|}{$\mathbf{F_1}$} &
  \textbf{MAE} \\ \midrule \midrule
\multirow{6}{*}{\begin{tabular}[c]{@{}c@{}}llava-\\ next-7B\end{tabular}} &
  \multirow{2}{*}{V} &
  - &
  \multicolumn{1}{c|}{0} &
  \multicolumn{1}{c|}{0} &
  \multicolumn{1}{c|}{0} &
  \multicolumn{1}{c|}{\cc{0}} &
  0.55 &
  \multicolumn{1}{c|}{0} &
  \multicolumn{1}{c|}{0} &
  \multicolumn{1}{c|}{0} &
  \multicolumn{1}{c|}{0} &
  0.42 &
  \multicolumn{1}{c|}{0} &
  \multicolumn{1}{c|}{0} &
  \multicolumn{1}{c|}{0} &
  \multicolumn{1}{c|}{0} &
  0.68 &
  \multicolumn{1}{c|}{0} &
  \multicolumn{1}{c|}{0} &
  \multicolumn{1}{c|}{0} &
  \multicolumn{1}{c|}{0} &
  0.85 &
  \multicolumn{1}{c|}{0} &
  \multicolumn{1}{c|}{0} &
  \multicolumn{1}{c|}{0} &
  \multicolumn{1}{c|}{0} &
  0.28 &
  \multicolumn{1}{c|}{0} &
  \multicolumn{1}{c|}{0} &
  \multicolumn{1}{c|}{0} &
  \multicolumn{1}{c|}{0} &
  0.62 \\ \cline{3-33} 
 &
   &
  SG-G &
  \multicolumn{1}{c|}{31.89} &
  \multicolumn{1}{c|}{0.74} &
  \multicolumn{1}{c|}{0.13} &
  \multicolumn{1}{c|}{\cc{0.22}} &
  0.55 &
  \multicolumn{1}{c|}{38.98} &
  \multicolumn{1}{c|}{0} &
  \multicolumn{1}{c|}{0} &
  \multicolumn{1}{c|}{0} &
  0.52 &
  \multicolumn{1}{c|}{27.67} &
  \multicolumn{1}{c|}{0.91} &
  \multicolumn{1}{c|}{0.15} &
  \multicolumn{1}{c|}{0.26} &
  0.65 &
  \multicolumn{1}{c|}{19.86} &
  \multicolumn{1}{c|}{0.5} &
  \multicolumn{1}{c|}{0.2} &
  \multicolumn{1}{c|}{0.28} &
  0.83 &
  \multicolumn{1}{c|}{35.74} &
  \multicolumn{1}{c|}{0.4} &
  \multicolumn{1}{c|}{0.17} &
  \multicolumn{1}{c|}{0.23} &
  0.31 &
  \multicolumn{1}{c|}{33.59} &
  \multicolumn{1}{c|}{0.6} &
  \multicolumn{1}{c|}{0.09} &
  \multicolumn{1}{c|}{0.16} &
  0.6 \\ \cline{2-33} 
 &
  \multirow{2}{*}{TSG} &
  + H &
  \multicolumn{1}{c|}{6.61} &
  \multicolumn{1}{c|}{0.77} &
  \multicolumn{1}{c|}{0.68} &
  \multicolumn{1}{c|}{0.72} &
  0.62 &
  \multicolumn{1}{c|}{0} &
  \multicolumn{1}{c|}{0.97} &
  \multicolumn{1}{c|}{0.63} &
  \multicolumn{1}{c|}{0.75} &
  0.44 &
  \multicolumn{1}{c|}{0} &
  \multicolumn{1}{c|}{0.96} &
  \multicolumn{1}{c|}{0.68} &
  \multicolumn{1}{c|}{0.8} &
  0.48 &
  \multicolumn{1}{c|}{0} &
  \multicolumn{1}{c|}{1} &
  \multicolumn{1}{c|}{0.77} &
  \multicolumn{1}{c|}{0.86} &
  0.57 &
  \multicolumn{1}{c|}{20} &
  \multicolumn{1}{c|}{0.34} &
  \multicolumn{1}{c|}{0.65} &
  \multicolumn{1}{c|}{0.45} &
  0.65 &
  \multicolumn{1}{c|}{9.13} &
  \multicolumn{1}{c|}{0.72} &
  \multicolumn{1}{c|}{0.63} &
  \multicolumn{1}{c|}{0.67} &
  0.82 \\ \cline{3-33} 
 &
   &
  - H &
  \multicolumn{1}{c|}{6.53} &
  \multicolumn{1}{c|}{1} &
  \multicolumn{1}{c|}{0.37} &
  \multicolumn{1}{c|}{\cc{0.54}} &
  0.42 &
  \multicolumn{1}{c|}{0} &
  \multicolumn{1}{c|}{1} &
  \multicolumn{1}{c|}{0.42} &
  \multicolumn{1}{c|}{0.59} &
  0.29 &
  \multicolumn{1}{c|}{0} &
  \multicolumn{1}{c|}{1} &
  \multicolumn{1}{c|}{0.4} &
  \multicolumn{1}{c|}{0.57} &
  0.5 &
  \multicolumn{1}{c|}{0} &
  \multicolumn{1}{c|}{1} &
  \multicolumn{1}{c|}{0.34} &
  \multicolumn{1}{c|}{0.49} &
  0.64 &
  \multicolumn{1}{c|}{20} &
  \multicolumn{1}{c|}{0.8} &
  \multicolumn{1}{c|}{0.19} &
  \multicolumn{1}{c|}{0.31} &
  0.18 &
  \multicolumn{1}{c|}{8.7} &
  \multicolumn{1}{c|}{1} &
  \multicolumn{1}{c|}{0.37} &
  \multicolumn{1}{c|}{0.54} &
  0.49 \\ \cline{2-33} 
 &
  \multirow{2}{*}{\begin{tabular}[c]{@{}c@{}}V \\ + TSG\end{tabular}} &
  + H &
  \multicolumn{1}{c|}{0} &
  \multicolumn{1}{c|}{1} &
  \multicolumn{1}{c|}{0.06} &
  \multicolumn{1}{c|}{0.11} &
  0.53 &
  \multicolumn{1}{c|}{0} &
  \multicolumn{1}{c|}{0.2} &
  \multicolumn{1}{c|}{0.02} &
  \multicolumn{1}{c|}{0.04} &
  0.41 &
  \multicolumn{1}{c|}{0} &
  \multicolumn{1}{c|}{1} &
  \multicolumn{1}{c|}{0.06} &
  \multicolumn{1}{c|}{0.12} &
  0.65 &
  \multicolumn{1}{c|}{0} &
  \multicolumn{1}{c|}{0} &
  \multicolumn{1}{c|}{0} &
  \multicolumn{1}{c|}{0} &
  0.85 &
  \multicolumn{1}{c|}{0} &
  \multicolumn{1}{c|}{1} &
  \multicolumn{1}{c|}{0.11} &
  \multicolumn{1}{c|}{0.2} &
  0.24 &
  \multicolumn{1}{c|}{0} &
  \multicolumn{1}{c|}{1} &
  \multicolumn{1}{c|}{0.05} &
  \multicolumn{1}{c|}{0.09} &
  0.59 \\ \cline{3-33} 
 &
   &
  - H &
  \multicolumn{1}{c|}{0} &
  \multicolumn{1}{c|}{1} &
  \multicolumn{1}{c|}{0.06} &
  \multicolumn{1}{c|}{0.11} &
  0.53 &
  \multicolumn{1}{c|}{0} &
  \multicolumn{1}{c|}{0.6} &
  \multicolumn{1}{c|}{0.1} &
  \multicolumn{1}{c|}{0.17} &
  0.38 &
  \multicolumn{1}{c|}{0} &
  \multicolumn{1}{c|}{1} &
  \multicolumn{1}{c|}{0.05} &
  \multicolumn{1}{c|}{0.1} &
  0.66 &
  \multicolumn{1}{c|}{0} &
  \multicolumn{1}{c|}{0} &
  \multicolumn{1}{c|}{0} &
  \multicolumn{1}{c|}{0} &
  0.85 &
  \multicolumn{1}{c|}{0} &
  \multicolumn{1}{c|}{1} &
  \multicolumn{1}{c|}{0.09} &
  \multicolumn{1}{c|}{0.17} &
  0.26 &
  \multicolumn{1}{c|}{0} &
  \multicolumn{1}{c|}{0.8} &
  \multicolumn{1}{c|}{0.04} &
  \multicolumn{1}{c|}{0.08} &
  0.6 \\ \midrule \midrule
\multirow{6}{*}{\begin{tabular}[c]{@{}c@{}}qwen3\\ -8B\end{tabular}} &
  \multirow{2}{*}{V} &
  - &
  \multicolumn{1}{c|}{0} &
  \multicolumn{1}{c|}{0.98} &
  \multicolumn{1}{c|}{0.52} &
  \multicolumn{1}{c|}{\cc{0.68}} &
  0.36 &
  \multicolumn{1}{c|}{0} &
  \multicolumn{1}{c|}{1} &
  \multicolumn{1}{c|}{0.59} &
  \multicolumn{1}{c|}{0.74} &
  0.23 &
  \multicolumn{1}{c|}{0} &
  \multicolumn{1}{c|}{0.98} &
  \multicolumn{1}{c|}{0.58} &
  \multicolumn{1}{c|}{0.73} &
  0.45 &
  \multicolumn{1}{c|}{0} &
  \multicolumn{1}{c|}{1} &
  \multicolumn{1}{c|}{0.29} &
  \multicolumn{1}{c|}{0.44} &
  0.67 &
  \multicolumn{1}{c|}{0} &
  \multicolumn{1}{c|}{1} &
  \multicolumn{1}{c|}{0.51} &
  \multicolumn{1}{c|}{0.67} &
  0.17 &
  \multicolumn{1}{c|}{0} &
  \multicolumn{1}{c|}{0.97} &
  \multicolumn{1}{c|}{0.47} &
  \multicolumn{1}{c|}{0.63} &
  0.41 \\ \cline{3-33} 
 &
   &
  SG-G &
  \multicolumn{1}{c|}{5.55} &
  \multicolumn{1}{c|}{0.8} &
  \multicolumn{1}{c|}{0.76} &
  \multicolumn{1}{c|}{\cc{0.78}} &
  0.45 &
  \multicolumn{1}{c|}{13.23} &
  \multicolumn{1}{c|}{0.84} &
  \multicolumn{1}{c|}{0.75} &
  \multicolumn{1}{c|}{0.79} &
  0.46 &
  \multicolumn{1}{c|}{4.71} &
  \multicolumn{1}{c|}{0.87} &
  \multicolumn{1}{c|}{0.78} &
  \multicolumn{1}{c|}{0.82} &
  0.47 &
  \multicolumn{1}{c|}{0} &
  \multicolumn{1}{c|}{1} &
  \multicolumn{1}{c|}{0.77} &
  \multicolumn{1}{c|}{0.87} &
  0.58 &
  \multicolumn{1}{c|}{6.51} &
  \multicolumn{1}{c|}{0.62} &
  \multicolumn{1}{c|}{0.69} &
  \multicolumn{1}{c|}{0.64} &
  0.3 &
  \multicolumn{1}{c|}{3.68} &
  \multicolumn{1}{c|}{0.76} &
  \multicolumn{1}{c|}{0.75} &
  \multicolumn{1}{c|}{0.75} &
  0.55 \\ \cline{2-33} 
 &
  \multirow{2}{*}{TSG} &
  + H &
  \multicolumn{1}{c|}{0.08} &
  \multicolumn{1}{c|}{0.79} &
  \multicolumn{1}{c|}{0.97} &
  \multicolumn{1}{c|}{0.87} &
  0.37 &
  \multicolumn{1}{c|}{0} &
  \multicolumn{1}{c|}{0.98} &
  \multicolumn{1}{c|}{0.93} &
  \multicolumn{1}{c|}{0.95} &
  0.19 &
  \multicolumn{1}{c|}{0} &
  \multicolumn{1}{c|}{0.81} &
  \multicolumn{1}{c|}{0.98} &
  \multicolumn{1}{c|}{0.88} &
  0.38 &
  \multicolumn{1}{c|}{0} &
  \multicolumn{1}{c|}{0.92} &
  \multicolumn{1}{c|}{0.85} &
  \multicolumn{1}{c|}{0.88} &
  0.58 &
  \multicolumn{1}{c|}{0.44} &
  \multicolumn{1}{c|}{0.68} &
  \multicolumn{1}{c|}{1} &
  \multicolumn{1}{c|}{0.81} &
  0.24 &
  \multicolumn{1}{c|}{0} &
  \multicolumn{1}{c|}{0.76} &
  \multicolumn{1}{c|}{0.97} &
  \multicolumn{1}{c|}{0.85} &
  0.51 \\ \cline{3-33} 
 &
   &
  - H &
  \multicolumn{1}{c|}{0.34} &
  \multicolumn{1}{c|}{0.92} &
  \multicolumn{1}{c|}{0.95} &
  \multicolumn{1}{c|}{\cc{0.93}} &
  0.21 &
  \multicolumn{1}{c|}{0} &
  \multicolumn{1}{c|}{1} &
  \multicolumn{1}{c|}{0.93} &
  \multicolumn{1}{c|}{0.96} &
  0.12 &
  \multicolumn{1}{c|}{0} &
  \multicolumn{1}{c|}{0.89} &
  \multicolumn{1}{c|}{0.98} &
  \multicolumn{1}{c|}{0.93} &
  0.27 &
  \multicolumn{1}{c|}{0} &
  \multicolumn{1}{c|}{0.92} &
  \multicolumn{1}{c|}{0.82} &
  \multicolumn{1}{c|}{0.86} &
  0.37 &
  \multicolumn{1}{c|}{0.44} &
  \multicolumn{1}{c|}{0.93} &
  \multicolumn{1}{c|}{0.91} &
  \multicolumn{1}{c|}{0.92} &
  0.08 &
  \multicolumn{1}{c|}{1.08} &
  \multicolumn{1}{c|}{0.93} &
  \multicolumn{1}{c|}{0.96} &
  \multicolumn{1}{c|}{0.94} &
  0.21 \\ \cline{2-33} 
 &
  \multirow{2}{*}{\begin{tabular}[c]{@{}c@{}}V \\ + TSG\end{tabular}} &
  + H &
  \multicolumn{1}{c|}{0} &
  \multicolumn{1}{c|}{0.98} &
  \multicolumn{1}{c|}{0.83} &
  \multicolumn{1}{c|}{0.9} &
  0.23 &
  \multicolumn{1}{c|}{0} &
  \multicolumn{1}{c|}{1} &
  \multicolumn{1}{c|}{0.88} &
  \multicolumn{1}{c|}{0.93} &
  0.15 &
  \multicolumn{1}{c|}{0} &
  \multicolumn{1}{c|}{0.99} &
  \multicolumn{1}{c|}{0.88} &
  \multicolumn{1}{c|}{0.93} &
  0.29 &
  \multicolumn{1}{c|}{0} &
  \multicolumn{1}{c|}{0.92} &
  \multicolumn{1}{c|}{0.6} &
  \multicolumn{1}{c|}{0.7} &
  0.52 &
  \multicolumn{1}{c|}{0} &
  \multicolumn{1}{c|}{0.98} &
  \multicolumn{1}{c|}{0.85} &
  \multicolumn{1}{c|}{0.91} &
  0.09 &
  \multicolumn{1}{c|}{0} &
  \multicolumn{1}{c|}{0.99} &
  \multicolumn{1}{c|}{0.79} &
  \multicolumn{1}{c|}{0.88} &
  0.24 \\ \cline{3-33} 
 &
   &
  - H &
  \multicolumn{1}{c|}{0.17} &
  \multicolumn{1}{c|}{0.99} &
  \multicolumn{1}{c|}{0.78} &
  \multicolumn{1}{c|}{0.87} &
  0.26 &
  \multicolumn{1}{c|}{0} &
  \multicolumn{1}{c|}{1} &
  \multicolumn{1}{c|}{0.88} &
  \multicolumn{1}{c|}{0.93} &
  0.13 &
  \multicolumn{1}{c|}{0} &
  \multicolumn{1}{c|}{0.98} &
  \multicolumn{1}{c|}{0.81} &
  \multicolumn{1}{c|}{0.89} &
  0.35 &
  \multicolumn{1}{c|}{0} &
  \multicolumn{1}{c|}{0.91} &
  \multicolumn{1}{c|}{0.57} &
  \multicolumn{1}{c|}{0.68} &
  0.54 &
  \multicolumn{1}{c|}{0} &
  \multicolumn{1}{c|}{1} &
  \multicolumn{1}{c|}{0.7} &
  \multicolumn{1}{c|}{0.82} &
  0.13 &
  \multicolumn{1}{c|}{0.73} &
  \multicolumn{1}{c|}{1} &
  \multicolumn{1}{c|}{0.79} &
  \multicolumn{1}{c|}{0.88} &
  0.25 \\ \midrule \midrule
\multirow{6}{*}{\begin{tabular}[c]{@{}c@{}}llama\\ -3.2\\ -11B\end{tabular}} &
  \multirow{2}{*}{V} &
  - &
  \multicolumn{1}{c|}{58.28} &
  \multicolumn{1}{c|}{0.79} &
  \multicolumn{1}{c|}{0.31} &
  \multicolumn{1}{c|}{\cc{0.43}} &
  0.51 &
  \multicolumn{1}{c|}{64.61} &
  \multicolumn{1}{c|}{0.9} &
  \multicolumn{1}{c|}{0.62} &
  \multicolumn{1}{c|}{0.66} &
  0.59 &
  \multicolumn{1}{c|}{56.53} &
  \multicolumn{1}{c|}{0.91} &
  \multicolumn{1}{c|}{0.27} &
  \multicolumn{1}{c|}{0.41} &
  0.56 &
  \multicolumn{1}{c|}{75.94} &
  \multicolumn{1}{c|}{1} &
  \multicolumn{1}{c|}{0.8} &
  \multicolumn{1}{c|}{0.87} &
  0.67 &
  \multicolumn{1}{c|}{49.33} &
  \multicolumn{1}{c|}{0.57} &
  \multicolumn{1}{c|}{0.33} &
  \multicolumn{1}{c|}{0.37} &
  0.29 &
  \multicolumn{1}{c|}{64.56} &
  \multicolumn{1}{c|}{0.72} &
  \multicolumn{1}{c|}{0.22} &
  \multicolumn{1}{c|}{0.33} &
  0.67 \\ \cline{3-33} 
 &
   &
  SG-G &
  \multicolumn{1}{c|}{97.1} &
  \multicolumn{1}{c|}{0.62} &
  \multicolumn{1}{c|}{0.84} &
  \multicolumn{1}{c|}{\cc{0.71}} &
  0.9 &
  \multicolumn{1}{c|}{95.73} &
  \multicolumn{1}{c|}{0.6} &
  \multicolumn{1}{c|}{0.6} &
  \multicolumn{1}{c|}{0.6} &
  0.7 &
  \multicolumn{1}{c|}{98.81} &
  \multicolumn{1}{c|}{0.6} &
  \multicolumn{1}{c|}{0.6} &
  \multicolumn{1}{c|}{0.6} &
  0.93 &
  \multicolumn{1}{c|}{91.1} &
  \multicolumn{1}{c|}{0.4} &
  \multicolumn{1}{c|}{0.4} &
  \multicolumn{1}{c|}{0.4} &
  0.6 &
  \multicolumn{1}{c|}{97.28} &
  \multicolumn{1}{c|}{0.53} &
  \multicolumn{1}{c|}{0.6} &
  \multicolumn{1}{c|}{0.56} &
  0.67 &
  \multicolumn{1}{c|}{95.34} &
  \multicolumn{1}{c|}{0.33} &
  \multicolumn{1}{c|}{0.6} &
  \multicolumn{1}{c|}{0.43} &
  0.9 \\ \cline{2-33} 
 &
  \multirow{2}{*}{TSG} &
  + H &
  \multicolumn{1}{c|}{0} &
  \multicolumn{1}{c|}{0.64} &
  \multicolumn{1}{c|}{0.97} &
  \multicolumn{1}{c|}{0.77} &
  1.35 &
  \multicolumn{1}{c|}{0} &
  \multicolumn{1}{c|}{0.84} &
  \multicolumn{1}{c|}{0.9} &
  \multicolumn{1}{c|}{0.87} &
  0.99 &
  \multicolumn{1}{c|}{0} &
  \multicolumn{1}{c|}{0.64} &
  \multicolumn{1}{c|}{0.98} &
  \multicolumn{1}{c|}{0.78} &
  1.34 &
  \multicolumn{1}{c|}{0} &
  \multicolumn{1}{c|}{0.89} &
  \multicolumn{1}{c|}{0.97} &
  \multicolumn{1}{c|}{0.92} &
  1.43 &
  \multicolumn{1}{c|}{0} &
  \multicolumn{1}{c|}{0.46} &
  \multicolumn{1}{c|}{0.97} &
  \multicolumn{1}{c|}{0.62} &
  1.22 &
  \multicolumn{1}{c|}{0} &
  \multicolumn{1}{c|}{0.67} &
  \multicolumn{1}{c|}{0.97} &
  \multicolumn{1}{c|}{0.79} &
  1.62 \\ \cline{3-33} 
 &
   &
  - H &
  \multicolumn{1}{c|}{0} &
  \multicolumn{1}{c|}{0.86} &
  \multicolumn{1}{c|}{0.9} &
  \multicolumn{1}{c|}{\cc{0.88}} &
  0.24 &
  \multicolumn{1}{c|}{0} &
  \multicolumn{1}{c|}{1} &
  \multicolumn{1}{c|}{0.9} &
  \multicolumn{1}{c|}{0.95} &
  0.23 &
  \multicolumn{1}{c|}{0} &
  \multicolumn{1}{c|}{0.85} &
  \multicolumn{1}{c|}{0.95} &
  \multicolumn{1}{c|}{0.9} &
  0.22 &
  \multicolumn{1}{c|}{0} &
  \multicolumn{1}{c|}{0.91} &
  \multicolumn{1}{c|}{0.76} &
  \multicolumn{1}{c|}{0.82} &
  0.53 &
  \multicolumn{1}{c|}{0} &
  \multicolumn{1}{c|}{0.85} &
  \multicolumn{1}{c|}{0.83} &
  \multicolumn{1}{c|}{0.84} &
  0.17 &
  \multicolumn{1}{c|}{0} &
  \multicolumn{1}{c|}{0.83} &
  \multicolumn{1}{c|}{0.9} &
  \multicolumn{1}{c|}{0.87} &
  0.31 \\ \cline{2-33} 
 &
  \multirow{2}{*}{\begin{tabular}[c]{@{}c@{}}V \\ + TSG\end{tabular}} &
  + H &
  \multicolumn{1}{c|}{3.32} &
  \multicolumn{1}{c|}{0.54} &
  \multicolumn{1}{c|}{0.94} &
  \multicolumn{1}{c|}{0.69} &
  0.56 &
  \multicolumn{1}{c|}{7.62} &
  \multicolumn{1}{c|}{0.47} &
  \multicolumn{1}{c|}{0.9} &
  \multicolumn{1}{c|}{0.62} &
  0.53 &
  \multicolumn{1}{c|}{1.18} &
  \multicolumn{1}{c|}{0.66} &
  \multicolumn{1}{c|}{0.95} &
  \multicolumn{1}{c|}{0.78} &
  0.47 &
  \multicolumn{1}{c|}{14.64} &
  \multicolumn{1}{c|}{0.66} &
  \multicolumn{1}{c|}{0.97} &
  \multicolumn{1}{c|}{0.78} &
  0.94 &
  \multicolumn{1}{c|}{4.84} &
  \multicolumn{1}{c|}{0.3} &
  \multicolumn{1}{c|}{0.95} &
  \multicolumn{1}{c|}{0.46} &
  0.62 &
  \multicolumn{1}{c|}{2.52} &
  \multicolumn{1}{c|}{0.6} &
  \multicolumn{1}{c|}{0.93} &
  \multicolumn{1}{c|}{0.73} &
  0.65 \\ \cline{3-33} 
 &
   &
  - H &
  \multicolumn{1}{c|}{6.06} &
  \multicolumn{1}{c|}{0.77} &
  \multicolumn{1}{c|}{0.82} &
  \multicolumn{1}{c|}{0.79} &
  0.36 &
  \multicolumn{1}{c|}{10.5} &
  \multicolumn{1}{c|}{0.83} &
  \multicolumn{1}{c|}{0.82} &
  \multicolumn{1}{c|}{0.82} &
  0.22 &
  \multicolumn{1}{c|}{1.37} &
  \multicolumn{1}{c|}{0.87} &
  \multicolumn{1}{c|}{0.85} &
  \multicolumn{1}{c|}{0.86} &
  0.4 &
  \multicolumn{1}{c|}{18.97} &
  \multicolumn{1}{c|}{0.88} &
  \multicolumn{1}{c|}{0.94} &
  \multicolumn{1}{c|}{0.9} &
  0.4 &
  \multicolumn{1}{c|}{8.46} &
  \multicolumn{1}{c|}{0.56} &
  \multicolumn{1}{c|}{0.8} &
  \multicolumn{1}{c|}{0.66} &
  0.25 &
  \multicolumn{1}{c|}{8.69} &
  \multicolumn{1}{c|}{0.72} &
  \multicolumn{1}{c|}{0.76} &
  \multicolumn{1}{c|}{0.74} &
  0.42 \\ \midrule \midrule
\multirow{6}{*}{\begin{tabular}[c]{@{}c@{}}mistral-\\ medium-3\end{tabular}} &
  \multirow{2}{*}{V} &
  - &
  \multicolumn{1}{c|}{9.67} &
  \multicolumn{1}{c|}{0.99} &
  \multicolumn{1}{c|}{0.44} &
  \multicolumn{1}{c|}{\cc{0.61}} &
  0.41 &
  \multicolumn{1}{c|}{0} &
  \multicolumn{1}{c|}{1} &
  \multicolumn{1}{c|}{0.46} &
  \multicolumn{1}{c|}{0.63} &
  0.27 &
  \multicolumn{1}{c|}{0.39} &
  \multicolumn{1}{c|}{0.99} &
  \multicolumn{1}{c|}{0.46} &
  \multicolumn{1}{c|}{0.62} &
  0.49 &
  \multicolumn{1}{c|}{17.78} &
  \multicolumn{1}{c|}{0.8} &
  \multicolumn{1}{c|}{0.19} &
  \multicolumn{1}{c|}{0.31} &
  0.6 &
  \multicolumn{1}{c|}{20} &
  \multicolumn{1}{c|}{0.8} &
  \multicolumn{1}{c|}{0.35} &
  \multicolumn{1}{c|}{0.48} &
  0.14 &
  \multicolumn{1}{c|}{20} &
  \multicolumn{1}{c|}{0.79} &
  \multicolumn{1}{c|}{0.34} &
  \multicolumn{1}{c|}{0.47} &
  0.34 \\ \cline{3-33} 
 &
   &
  SG-G &
  \multicolumn{1}{c|}{9.75} &
  \multicolumn{1}{c|}{0.83} &
  \multicolumn{1}{c|}{0.6} &
  \multicolumn{1}{c|}{\cc{0.69}} &
  0.42 &
  \multicolumn{1}{c|}{0} &
  \multicolumn{1}{c|}{0.97} &
  \multicolumn{1}{c|}{0.52} &
  \multicolumn{1}{c|}{0.66} &
  0.33 &
  \multicolumn{1}{c|}{0.39} &
  \multicolumn{1}{c|}{0.89} &
  \multicolumn{1}{c|}{0.62} &
  \multicolumn{1}{c|}{0.73} &
  0.45 &
  \multicolumn{1}{c|}{20} &
  \multicolumn{1}{c|}{0.8} &
  \multicolumn{1}{c|}{0.34} &
  \multicolumn{1}{c|}{0.47} &
  0.56 &
  \multicolumn{1}{c|}{20} &
  \multicolumn{1}{c|}{0.39} &
  \multicolumn{1}{c|}{0.47} &
  \multicolumn{1}{c|}{0.42} &
  0.29 &
  \multicolumn{1}{c|}{20} &
  \multicolumn{1}{c|}{0.71} &
  \multicolumn{1}{c|}{0.49} &
  \multicolumn{1}{c|}{0.58} &
  0.34 \\ \cline{2-33} 
 &
  \multirow{2}{*}{TSG} &
  + H &
  \multicolumn{1}{c|}{9.75} &
  \multicolumn{1}{c|}{0.68} &
  \multicolumn{1}{c|}{0.99} &
  \multicolumn{1}{c|}{0.81} &
  0.67 &
  \multicolumn{1}{c|}{0} &
  \multicolumn{1}{c|}{0.64} &
  \multicolumn{1}{c|}{0.98} &
  \multicolumn{1}{c|}{0.77} &
  0.84 &
  \multicolumn{1}{c|}{0.59} &
  \multicolumn{1}{c|}{0.75} &
  \multicolumn{1}{c|}{0.99} &
  \multicolumn{1}{c|}{0.86} &
  0.53 &
  \multicolumn{1}{c|}{17.78} &
  \multicolumn{1}{c|}{0.93} &
  \multicolumn{1}{c|}{0.97} &
  \multicolumn{1}{c|}{0.95} &
  0.74 &
  \multicolumn{1}{c|}{20} &
  \multicolumn{1}{c|}{0.31} &
  \multicolumn{1}{c|}{0.8} &
  \multicolumn{1}{c|}{0.44} &
  0.71 &
  \multicolumn{1}{c|}{20} &
  \multicolumn{1}{c|}{0.61} &
  \multicolumn{1}{c|}{0.8} &
  \multicolumn{1}{c|}{0.69} &
  0.62 \\ \cline{3-33} 
 &
   &
  - H &
  \multicolumn{1}{c|}{9.75} &
  \multicolumn{1}{c|}{0.88} &
  \multicolumn{1}{c|}{0.97} &
  \multicolumn{1}{c|}{\cc{0.92}} &
  0.17 &
  \multicolumn{1}{c|}{0} &
  \multicolumn{1}{c|}{0.88} &
  \multicolumn{1}{c|}{0.93} &
  \multicolumn{1}{c|}{0.9} &
  0.16 &
  \multicolumn{1}{c|}{0.39} &
  \multicolumn{1}{c|}{0.88} &
  \multicolumn{1}{c|}{0.99} &
  \multicolumn{1}{c|}{0.93} &
  0.18 &
  \multicolumn{1}{c|}{20} &
  \multicolumn{1}{c|}{0.72} &
  \multicolumn{1}{c|}{0.66} &
  \multicolumn{1}{c|}{0.68} &
  0.3 &
  \multicolumn{1}{c|}{20} &
  \multicolumn{1}{c|}{0.6} &
  \multicolumn{1}{c|}{0.78} &
  \multicolumn{1}{c|}{0.68} &
  0.07 &
  \multicolumn{1}{c|}{20} &
  \multicolumn{1}{c|}{0.75} &
  \multicolumn{1}{c|}{0.78} &
  \multicolumn{1}{c|}{0.76} &
  0.14 \\ \cline{2-33} 
 &
  \multirow{2}{*}{\begin{tabular}[c]{@{}c@{}}V \\ + TSG\end{tabular}} &
  + H &
  \multicolumn{1}{c|}{9.67} &
  \multicolumn{1}{c|}{0.98} &
  \multicolumn{1}{c|}{0.83} &
  \multicolumn{1}{c|}{0.9} &
  0.23 &
  \multicolumn{1}{c|}{0} &
  \multicolumn{1}{c|}{1} &
  \multicolumn{1}{c|}{0.9} &
  \multicolumn{1}{c|}{0.95} &
  0.13 &
  \multicolumn{1}{c|}{0.2} &
  \multicolumn{1}{c|}{0.96} &
  \multicolumn{1}{c|}{0.85} &
  \multicolumn{1}{c|}{0.91} &
  0.25 &
  \multicolumn{1}{c|}{20} &
  \multicolumn{1}{c|}{0.76} &
  \multicolumn{1}{c|}{0.39} &
  \multicolumn{1}{c|}{0.5} &
  0.6 &
  \multicolumn{1}{c|}{20} &
  \multicolumn{1}{c|}{0.8} &
  \multicolumn{1}{c|}{0.64} &
  \multicolumn{1}{c|}{0.71} &
  0.08 &
  \multicolumn{1}{c|}{20} &
  \multicolumn{1}{c|}{0.8} &
  \multicolumn{1}{c|}{0.64} &
  \multicolumn{1}{c|}{0.71} &
  0.21 \\ \cline{3-33} 
 &
   &
  - H &
  \multicolumn{1}{c|}{9.83} &
  \multicolumn{1}{c|}{0.96} &
  \multicolumn{1}{c|}{0.83} &
  \multicolumn{1}{c|}{0.89} &
  0.23 &
  \multicolumn{1}{c|}{0} &
  \multicolumn{1}{c|}{1} &
  \multicolumn{1}{c|}{0.9} &
  \multicolumn{1}{c|}{0.95} &
  0.11 &
  \multicolumn{1}{c|}{0.59} &
  \multicolumn{1}{c|}{0.94} &
  \multicolumn{1}{c|}{0.85} &
  \multicolumn{1}{c|}{0.89} &
  0.27 &
  \multicolumn{1}{c|}{20} &
  \multicolumn{1}{c|}{0.77} &
  \multicolumn{1}{c|}{0.42} &
  \multicolumn{1}{c|}{0.52} &
  0.47 &
  \multicolumn{1}{c|}{20} &
  \multicolumn{1}{c|}{0.8} &
  \multicolumn{1}{c|}{0.63} &
  \multicolumn{1}{c|}{0.7} &
  0.08 &
  \multicolumn{1}{c|}{20} &
  \multicolumn{1}{c|}{0.79} &
  \multicolumn{1}{c|}{0.65} &
  \multicolumn{1}{c|}{0.71} &
  0.22 \\ \midrule \midrule
\multirow{6}{*}{\begin{tabular}[c]{@{}c@{}}claude-\\ sonnet-4.5\end{tabular}} &
  \multirow{2}{*}{V} &
  - &
  \multicolumn{1}{c|}{0} &
  \multicolumn{1}{c|}{0.94} &
  \multicolumn{1}{c|}{0.47} &
  \multicolumn{1}{c|}{\cc{0.62}} &
  0.39 &
  \multicolumn{1}{c|}{0} &
  \multicolumn{1}{c|}{1} &
  \multicolumn{1}{c|}{0.28} &
  \multicolumn{1}{c|}{0.43} &
  0.33 &
  \multicolumn{1}{c|}{0} &
  \multicolumn{1}{c|}{0.96} &
  \multicolumn{1}{c|}{0.51} &
  \multicolumn{1}{c|}{0.66} &
  0.46 &
  \multicolumn{1}{c|}{0} &
  \multicolumn{1}{c|}{0.97} &
  \multicolumn{1}{c|}{0.63} &
  \multicolumn{1}{c|}{0.75} &
  0.59 &
  \multicolumn{1}{c|}{0} &
  \multicolumn{1}{c|}{0.89} &
  \multicolumn{1}{c|}{0.41} &
  \multicolumn{1}{c|}{0.54} &
  0.2 &
  \multicolumn{1}{c|}{0} &
  \multicolumn{1}{c|}{0.94} &
  \multicolumn{1}{c|}{0.45} &
  \multicolumn{1}{c|}{0.61} &
  0.42 \\ \cline{3-33} 
 &
   &
  SG-G &
  \multicolumn{1}{c|}{0.08} &
  \multicolumn{1}{c|}{0.73} &
  \multicolumn{1}{c|}{0.68} &
  \multicolumn{1}{c|}{\cc{0.7}} &
  0.45 &
  \multicolumn{1}{c|}{0} &
  \multicolumn{1}{c|}{0.83} &
  \multicolumn{1}{c|}{0.43} &
  \multicolumn{1}{c|}{0.55} &
  0.38 &
  \multicolumn{1}{c|}{0} &
  \multicolumn{1}{c|}{0.78} &
  \multicolumn{1}{c|}{0.73} &
  \multicolumn{1}{c|}{0.75} &
  0.45 &
  \multicolumn{1}{c|}{0} &
  \multicolumn{1}{c|}{0.92} &
  \multicolumn{1}{c|}{0.66} &
  \multicolumn{1}{c|}{0.76} &
  0.87 &
  \multicolumn{1}{c|}{0} &
  \multicolumn{1}{c|}{0.49} &
  \multicolumn{1}{c|}{0.58} &
  \multicolumn{1}{c|}{0.53} &
  0.34 &
  \multicolumn{1}{c|}{0.35} &
  \multicolumn{1}{c|}{0.77} &
  \multicolumn{1}{c|}{0.72} &
  \multicolumn{1}{c|}{0.74} &
  0.53 \\ \cline{2-33} 
 &
  \multirow{2}{*}{TSG} &
  + H &
  \multicolumn{1}{c|}{5.71} &
  \multicolumn{1}{c|}{0.89} &
  \multicolumn{1}{c|}{0.98} &
  \multicolumn{1}{c|}{0.93} &
  0.23 &
  \multicolumn{1}{c|}{47.73} &
  \multicolumn{1}{c|}{1} &
  \multicolumn{1}{c|}{1} &
  \multicolumn{1}{c|}{1} &
  0.16 &
  \multicolumn{1}{c|}{0} &
  \multicolumn{1}{c|}{0.88} &
  \multicolumn{1}{c|}{0.99} &
  \multicolumn{1}{c|}{0.93} &
  0.19 &
  \multicolumn{1}{c|}{0} &
  \multicolumn{1}{c|}{0.93} &
  \multicolumn{1}{c|}{1} &
  \multicolumn{1}{c|}{0.96} &
  0.61 &
  \multicolumn{1}{c|}{1.92} &
  \multicolumn{1}{c|}{0.78} &
  \multicolumn{1}{c|}{0.93} &
  \multicolumn{1}{c|}{0.85} &
  0.15 &
  \multicolumn{1}{c|}{2.52} &
  \multicolumn{1}{c|}{0.96} &
  \multicolumn{1}{c|}{0.97} &
  \multicolumn{1}{c|}{0.97} &
  0.34 \\ \cline{3-33} 
 &
   &
  - H &
  \multicolumn{1}{c|}{5.88} &
  \multicolumn{1}{c|}{0.92} &
  \multicolumn{1}{c|}{0.95} &
  \multicolumn{1}{c|}{\cc{0.94}} &
  0.16 &
  \multicolumn{1}{c|}{50.14} &
  \multicolumn{1}{c|}{1} &
  \multicolumn{1}{c|}{0.92} &
  \multicolumn{1}{c|}{0.95} &
  0.25 &
  \multicolumn{1}{c|}{0.19} &
  \multicolumn{1}{c|}{0.9} &
  \multicolumn{1}{c|}{0.99} &
  \multicolumn{1}{c|}{0.94} &
  0.13 &
  \multicolumn{1}{c|}{0} &
  \multicolumn{1}{c|}{0.93} &
  \multicolumn{1}{c|}{1} &
  \multicolumn{1}{c|}{0.96} &
  0.25 &
  \multicolumn{1}{c|}{1.15} &
  \multicolumn{1}{c|}{1} &
  \multicolumn{1}{c|}{0.86} &
  \multicolumn{1}{c|}{0.93} &
  0.07 &
  \multicolumn{1}{c|}{2.52} &
  \multicolumn{1}{c|}{0.92} &
  \multicolumn{1}{c|}{0.93} &
  \multicolumn{1}{c|}{0.92} &
  0.27 \\ \cline{2-33} 
 &
  \multirow{2}{*}{\begin{tabular}[c]{@{}c@{}}V \\ + TSG\end{tabular}} &
  + H &
  \multicolumn{1}{c|}{5.14} &
  \multicolumn{1}{c|}{0.85} &
  \multicolumn{1}{c|}{0.84} &
  \multicolumn{1}{c|}{0.84} &
  0.27 &
  \multicolumn{1}{c|}{43.98} &
  \multicolumn{1}{c|}{0.97} &
  \multicolumn{1}{c|}{0.89} &
  \multicolumn{1}{c|}{0.93} &
  0.17 &
  \multicolumn{1}{c|}{0} &
  \multicolumn{1}{c|}{0.82} &
  \multicolumn{1}{c|}{0.86} &
  \multicolumn{1}{c|}{0.84} &
  0.3 &
  \multicolumn{1}{c|}{0} &
  \multicolumn{1}{c|}{0.97} &
  \multicolumn{1}{c|}{0.69} &
  \multicolumn{1}{c|}{0.79} &
  0.53 &
  \multicolumn{1}{c|}{1.18} &
  \multicolumn{1}{c|}{0.69} &
  \multicolumn{1}{c|}{0.73} &
  \multicolumn{1}{c|}{0.71} &
  0.19 &
  \multicolumn{1}{c|}{2.52} &
  \multicolumn{1}{c|}{0.94} &
  \multicolumn{1}{c|}{0.89} &
  \multicolumn{1}{c|}{0.91} &
  0.27 \\ \cline{3-33} 
 &
   &
  - H &
  \multicolumn{1}{c|}{5.39} &
  \multicolumn{1}{c|}{0.87} &
  \multicolumn{1}{c|}{0.85} &
  \multicolumn{1}{c|}{0.86} &
  0.26 &
  \multicolumn{1}{c|}{46.45} &
  \multicolumn{1}{c|}{0.96} &
  \multicolumn{1}{c|}{0.85} &
  \multicolumn{1}{c|}{0.9} &
  0.19 &
  \multicolumn{1}{c|}{0.19} &
  \multicolumn{1}{c|}{0.84} &
  \multicolumn{1}{c|}{0.91} &
  \multicolumn{1}{c|}{0.87} &
  0.29 &
  \multicolumn{1}{c|}{0} &
  \multicolumn{1}{c|}{1} &
  \multicolumn{1}{c|}{0.61} &
  \multicolumn{1}{c|}{0.74} &
  0.46 &
  \multicolumn{1}{c|}{1.18} &
  \multicolumn{1}{c|}{0.78} &
  \multicolumn{1}{c|}{0.68} &
  \multicolumn{1}{c|}{0.72} &
  0.18 &
  \multicolumn{1}{c|}{2.14} &
  \multicolumn{1}{c|}{0.92} &
  \multicolumn{1}{c|}{0.88} &
  \multicolumn{1}{c|}{0.9} &
  0.26 \\ \midrule \midrule
\multirow{6}{*}{\begin{tabular}[c]{@{}c@{}}gemini-\\ 3-flash\end{tabular}} &
  \multirow{2}{*}{V} &
  - &
  \multicolumn{1}{c|}{0.17} &
  \multicolumn{1}{c|}{0.85} &
  \multicolumn{1}{c|}{0.85} &
  \multicolumn{1}{c|}{\cc{0.85}} &
  0.37 &
  \multicolumn{1}{c|}{0} &
  \multicolumn{1}{c|}{0.86} &
  \multicolumn{1}{c|}{0.77} &
  \multicolumn{1}{c|}{0.81} &
  0.27 &
  \multicolumn{1}{c|}{0.39} &
  \multicolumn{1}{c|}{0.85} &
  \multicolumn{1}{c|}{0.91} &
  \multicolumn{1}{c|}{0.88} &
  0.42 &
  \multicolumn{1}{c|}{0} &
  \multicolumn{1}{c|}{0.87} &
  \multicolumn{1}{c|}{0.66} &
  \multicolumn{1}{c|}{0.73} &
  0.66 &
  \multicolumn{1}{c|}{0} &
  \multicolumn{1}{c|}{0.82} &
  \multicolumn{1}{c|}{0.75} &
  \multicolumn{1}{c|}{0.78} &
  0.2 &
  \multicolumn{1}{c|}{0} &
  \multicolumn{1}{c|}{0.87} &
  \multicolumn{1}{c|}{0.85} &
  \multicolumn{1}{c|}{0.85} &
  0.44 \\ \cline{3-33} 
 &
   &
  SG-G &
  \multicolumn{1}{c|}{1.42} &
  \multicolumn{1}{c|}{0.8} &
  \multicolumn{1}{c|}{0.9} &
  \multicolumn{1}{c|}{\cc{0.85}} &
  0.38 &
  \multicolumn{1}{c|}{1} &
  \multicolumn{1}{c|}{0.76} &
  \multicolumn{1}{c|}{0.83} &
  \multicolumn{1}{c|}{0.79} &
  0.35 &
  \multicolumn{1}{c|}{1.2} &
  \multicolumn{1}{c|}{0.83} &
  \multicolumn{1}{c|}{0.95} &
  \multicolumn{1}{c|}{0.88} &
  0.42 &
  \multicolumn{1}{c|}{0} &
  \multicolumn{1}{c|}{0.9} &
  \multicolumn{1}{c|}{0.78} &
  \multicolumn{1}{c|}{0.82} &
  0.57 &
  \multicolumn{1}{c|}{2.52} &
  \multicolumn{1}{c|}{0.78} &
  \multicolumn{1}{c|}{0.9} &
  \multicolumn{1}{c|}{0.83} &
  0.2 &
  \multicolumn{1}{c|}{1.4} &
  \multicolumn{1}{c|}{0.79} &
  \multicolumn{1}{c|}{0.88} &
  \multicolumn{1}{c|}{0.84} &
  0.47 \\ \cline{2-33} 
 &
  \multirow{2}{*}{TSG} &
  + H &
  \multicolumn{1}{c|}{0.33} &
  \multicolumn{1}{c|}{0.92} &
  \multicolumn{1}{c|}{0.98} &
  \multicolumn{1}{c|}{0.95} &
  0.21 &
  \multicolumn{1}{c|}{0} &
  \multicolumn{1}{c|}{1} &
  \multicolumn{1}{c|}{1} &
  \multicolumn{1}{c|}{1} &
  0.15 &
  \multicolumn{1}{c|}{0.39} &
  \multicolumn{1}{c|}{0.93} &
  \multicolumn{1}{c|}{0.98} &
  \multicolumn{1}{c|}{0.96} &
  0.16 &
  \multicolumn{1}{c|}{0} &
  \multicolumn{1}{c|}{0.93} &
  \multicolumn{1}{c|}{1} &
  \multicolumn{1}{c|}{0.96} &
  0.59 &
  \multicolumn{1}{c|}{0.79} &
  \multicolumn{1}{c|}{0.82} &
  \multicolumn{1}{c|}{0.93} &
  \multicolumn{1}{c|}{0.87} &
  0.11 &
  \multicolumn{1}{c|}{0} &
  \multicolumn{1}{c|}{0.91} &
  \multicolumn{1}{c|}{0.97} &
  \multicolumn{1}{c|}{0.94} &
  0.35 \\ \cline{3-33} 
 &
   &
  - H &
  \multicolumn{1}{c|}{0.25} &
  \multicolumn{1}{c|}{0.97} &
  \multicolumn{1}{c|}{0.94} &
  \multicolumn{1}{c|}{\cc{0.96}} &
  0.13 &
  \multicolumn{1}{c|}{0} &
  \multicolumn{1}{c|}{1} &
  \multicolumn{1}{c|}{0.95} &
  \multicolumn{1}{c|}{0.97} &
  0.13 &
  \multicolumn{1}{c|}{0.21} &
  \multicolumn{1}{c|}{0.97} &
  \multicolumn{1}{c|}{0.96} &
  \multicolumn{1}{c|}{0.97} &
  0.12 &
  \multicolumn{1}{c|}{3.33} &
  \multicolumn{1}{c|}{0.93} &
  \multicolumn{1}{c|}{0.88} &
  \multicolumn{1}{c|}{0.89} &
  0.27 &
  \multicolumn{1}{c|}{0.4} &
  \multicolumn{1}{c|}{0.98} &
  \multicolumn{1}{c|}{0.83} &
  \multicolumn{1}{c|}{0.9} &
  0.08 &
  \multicolumn{1}{c|}{0} &
  \multicolumn{1}{c|}{0.98} &
  \multicolumn{1}{c|}{0.96} &
  \multicolumn{1}{c|}{0.97} &
  0.19 \\ \cline{2-33} 
 &
  \multirow{2}{*}{\begin{tabular}[c]{@{}c@{}}V \\ + TSG\end{tabular}} &
  + H &
  \multicolumn{1}{c|}{0.41} &
  \multicolumn{1}{c|}{0.89} &
  \multicolumn{1}{c|}{0.97} &
  \multicolumn{1}{c|}{0.92} &
  0.24 &
  \multicolumn{1}{c|}{0} &
  \multicolumn{1}{c|}{0.93} &
  \multicolumn{1}{c|}{0.98} &
  \multicolumn{1}{c|}{0.95} &
  0.14 &
  \multicolumn{1}{c|}{0.6} &
  \multicolumn{1}{c|}{0.9} &
  \multicolumn{1}{c|}{0.98} &
  \multicolumn{1}{c|}{0.94} &
  0.26 &
  \multicolumn{1}{c|}{0} &
  \multicolumn{1}{c|}{0.92} &
  \multicolumn{1}{c|}{0.89} &
  \multicolumn{1}{c|}{0.9} &
  0.45 &
  \multicolumn{1}{c|}{0.34} &
  \multicolumn{1}{c|}{0.84} &
  \multicolumn{1}{c|}{0.97} &
  \multicolumn{1}{c|}{0.9} &
  0.14 &
  \multicolumn{1}{c|}{0.38} &
  \multicolumn{1}{c|}{0.87} &
  \multicolumn{1}{c|}{0.97} &
  \multicolumn{1}{c|}{0.92} &
  0.3 \\ \cline{3-33} 
 &
   &
  - H &
  \multicolumn{1}{c|}{0.08} &
  \multicolumn{1}{c|}{0.87} &
  \multicolumn{1}{c|}{0.97} &
  \multicolumn{1}{c|}{0.92} &
  0.27 &
  \multicolumn{1}{c|}{0} &
  \multicolumn{1}{c|}{0.85} &
  \multicolumn{1}{c|}{1} &
  \multicolumn{1}{c|}{0.92} &
  0.18 &
  \multicolumn{1}{c|}{0} &
  \multicolumn{1}{c|}{0.88} &
  \multicolumn{1}{c|}{0.97} &
  \multicolumn{1}{c|}{0.92} &
  0.3 &
  \multicolumn{1}{c|}{0} &
  \multicolumn{1}{c|}{0.92} &
  \multicolumn{1}{c|}{0.92} &
  \multicolumn{1}{c|}{0.92} &
  0.41 &
  \multicolumn{1}{c|}{0} &
  \multicolumn{1}{c|}{0.87} &
  \multicolumn{1}{c|}{0.97} &
  \multicolumn{1}{c|}{0.91} &
  0.13 &
  \multicolumn{1}{c|}{0.35} &
  \multicolumn{1}{c|}{0.85} &
  \multicolumn{1}{c|}{0.97} &
  \multicolumn{1}{c|}{0.9} &
  0.34 \\ \midrule \midrule
\multirow{6}{*}{\begin{tabular}[c]{@{}c@{}}gemini-\\ 3-pro\end{tabular}} &
  \multirow{2}{*}{V} &
  - &
  \multicolumn{1}{c|}{1.08} &
  \multicolumn{1}{c|}{0.87} &
  \multicolumn{1}{c|}{0.86} &
  \multicolumn{1}{c|}{\cc{0.86}} &
  0.33 &
  \multicolumn{1}{c|}{1} &
  \multicolumn{1}{c|}{0.81} &
  \multicolumn{1}{c|}{0.76} &
  \multicolumn{1}{c|}{0.77} &
  0.38 &
  \multicolumn{1}{c|}{1.38} &
  \multicolumn{1}{c|}{0.9} &
  \multicolumn{1}{c|}{0.9} &
  \multicolumn{1}{c|}{0.9} &
  0.36 &
  \multicolumn{1}{c|}{0} &
  \multicolumn{1}{c|}{0.9} &
  \multicolumn{1}{c|}{0.64} &
  \multicolumn{1}{c|}{0.74} &
  0.72 &
  \multicolumn{1}{c|}{0.34} &
  \multicolumn{1}{c|}{0.81} &
  \multicolumn{1}{c|}{0.82} &
  \multicolumn{1}{c|}{0.81} &
  0.17 &
  \multicolumn{1}{c|}{1.44} &
  \multicolumn{1}{c|}{0.86} &
  \multicolumn{1}{c|}{0.88} &
  \multicolumn{1}{c|}{0.87} &
  0.38 \\ \cline{3-33} 
 &
   &
  SG-G &
  \multicolumn{1}{c|}{0.42} &
  \multicolumn{1}{c|}{0.81} &
  \multicolumn{1}{c|}{0.9} &
  \multicolumn{1}{c|}{\cc{0.85}} &
  0.37 &
  \multicolumn{1}{c|}{0.74} &
  \multicolumn{1}{c|}{0.76} &
  \multicolumn{1}{c|}{0.82} &
  \multicolumn{1}{c|}{0.79} &
  0.43 &
  \multicolumn{1}{c|}{0.4} &
  \multicolumn{1}{c|}{0.82} &
  \multicolumn{1}{c|}{0.94} &
  \multicolumn{1}{c|}{0.87} &
  0.38 &
  \multicolumn{1}{c|}{0} &
  \multicolumn{1}{c|}{0.9} &
  \multicolumn{1}{c|}{0.86} &
  \multicolumn{1}{c|}{0.86} &
  0.72 &
  \multicolumn{1}{c|}{0} &
  \multicolumn{1}{c|}{0.7} &
  \multicolumn{1}{c|}{0.87} &
  \multicolumn{1}{c|}{0.78} &
  0.23 &
  \multicolumn{1}{c|}{0.73} &
  \multicolumn{1}{c|}{0.87} &
  \multicolumn{1}{c|}{0.91} &
  \multicolumn{1}{c|}{0.89} &
  0.42 \\ \cline{2-33} 
 &
  \multirow{2}{*}{TSG} &
  + H &
  \multicolumn{1}{c|}{0.08} &
  \multicolumn{1}{c|}{0.95} &
  \multicolumn{1}{c|}{0.98} &
  \multicolumn{1}{c|}{0.96} &
  0.13 &
  \multicolumn{1}{c|}{0} &
  \multicolumn{1}{c|}{1} &
  \multicolumn{1}{c|}{0.98} &
  \multicolumn{1}{c|}{0.99} &
  0.13 &
  \multicolumn{1}{c|}{0.19} &
  \multicolumn{1}{c|}{0.93} &
  \multicolumn{1}{c|}{0.99} &
  \multicolumn{1}{c|}{0.96} &
  0.13 &
  \multicolumn{1}{c|}{0} &
  \multicolumn{1}{c|}{0.93} &
  \multicolumn{1}{c|}{1} &
  \multicolumn{1}{c|}{0.96} &
  0.37 &
  \multicolumn{1}{c|}{0} &
  \multicolumn{1}{c|}{0.98} &
  \multicolumn{1}{c|}{0.93} &
  \multicolumn{1}{c|}{0.96} &
  0.04 &
  \multicolumn{1}{c|}{0} &
  \multicolumn{1}{c|}{0.97} &
  \multicolumn{1}{c|}{0.97} &
  \multicolumn{1}{c|}{0.97} &
  0.17 \\ \cline{3-33} 
 &
   &
  - H &
  \multicolumn{1}{c|}{0.25} &
  \multicolumn{1}{c|}{0.96} &
  \multicolumn{1}{c|}{0.96} &
  \multicolumn{1}{c|}{\cc{0.96}} &
  0.14 &
  \multicolumn{1}{c|}{0} &
  \multicolumn{1}{c|}{1} &
  \multicolumn{1}{c|}{0.95} &
  \multicolumn{1}{c|}{0.97} &
  0.12 &
  \multicolumn{1}{c|}{0.19} &
  \multicolumn{1}{c|}{0.94} &
  \multicolumn{1}{c|}{0.99} &
  \multicolumn{1}{c|}{0.96} &
  0.14 &
  \multicolumn{1}{c|}{0} &
  \multicolumn{1}{c|}{0.93} &
  \multicolumn{1}{c|}{1} &
  \multicolumn{1}{c|}{0.96} &
  0.34 &
  \multicolumn{1}{c|}{0.34} &
  \multicolumn{1}{c|}{0.96} &
  \multicolumn{1}{c|}{0.83} &
  \multicolumn{1}{c|}{0.89} &
  0.08 &
  \multicolumn{1}{c|}{0.35} &
  \multicolumn{1}{c|}{0.98} &
  \multicolumn{1}{c|}{0.96} &
  \multicolumn{1}{c|}{0.97} &
  0.15 \\ \cline{2-33} 
 &
  \multirow{2}{*}{\begin{tabular}[c]{@{}c@{}}V \\ + TSG\end{tabular}} &
  + H &
  \multicolumn{1}{c|}{0.58} &
  \multicolumn{1}{c|}{0.95} &
  \multicolumn{1}{c|}{0.94} &
  \multicolumn{1}{c|}{0.95} &
  0.21 &
  \multicolumn{1}{c|}{0} &
  \multicolumn{1}{c|}{0.91} &
  \multicolumn{1}{c|}{0.98} &
  \multicolumn{1}{c|}{0.94} &
  0.16 &
  \multicolumn{1}{c|}{0.57} &
  \multicolumn{1}{c|}{0.99} &
  \multicolumn{1}{c|}{0.96} &
  \multicolumn{1}{c|}{0.97} &
  0.24 &
  \multicolumn{1}{c|}{2} &
  \multicolumn{1}{c|}{0.92} &
  \multicolumn{1}{c|}{0.79} &
  \multicolumn{1}{c|}{0.83} &
  0.51 &
  \multicolumn{1}{c|}{0.34} &
  \multicolumn{1}{c|}{0.9} &
  \multicolumn{1}{c|}{0.95} &
  \multicolumn{1}{c|}{0.93} &
  0.09 &
  \multicolumn{1}{c|}{0.77} &
  \multicolumn{1}{c|}{0.92} &
  \multicolumn{1}{c|}{0.94} &
  \multicolumn{1}{c|}{0.93} &
  0.24 \\ \cline{3-33} 
 &
   &
  - H &
  \multicolumn{1}{c|}{0.83} &
  \multicolumn{1}{c|}{0.93} &
  \multicolumn{1}{c|}{0.95} &
  \multicolumn{1}{c|}{0.94} &
  0.22 &
  \multicolumn{1}{c|}{0.74} &
  \multicolumn{1}{c|}{0.9} &
  \multicolumn{1}{c|}{0.98} &
  \multicolumn{1}{c|}{0.93} &
  0.2 &
  \multicolumn{1}{c|}{0.79} &
  \multicolumn{1}{c|}{0.95} &
  \multicolumn{1}{c|}{0.96} &
  \multicolumn{1}{c|}{0.96} &
  0.24 &
  \multicolumn{1}{c|}{2} &
  \multicolumn{1}{c|}{0.92} &
  \multicolumn{1}{c|}{0.87} &
  \multicolumn{1}{c|}{0.89} &
  0.44 &
  \multicolumn{1}{c|}{0.79} &
  \multicolumn{1}{c|}{0.95} &
  \multicolumn{1}{c|}{0.91} &
  \multicolumn{1}{c|}{0.93} &
  0.1 &
  \multicolumn{1}{c|}{0.73} &
  \multicolumn{1}{c|}{0.93} &
  \multicolumn{1}{c|}{0.96} &
  \multicolumn{1}{c|}{0.94} &
  0.28 \\ \bottomrule \bottomrule
\end{tabular}
}
\caption{Hazard detection performance of VLMs across input modalities (M) and reasoning settings (S), reported overall and by laboratory domain. We compare vision-only (V), textual scene graph with hazard (TSG+H), textual scene graph without hazard (TSG-H), and vision + textual scene graph (V+TSG) inputs using format error (FE,\%), precision (P), recall (R), $F_1$ score, and  mean absolute error (MAE).
Textual scene graph–based settings, with or without visual input, generally outperform vision-only models across nearly all cases. This indicates that VLMs reason effectively when structured object relationships are provided.
SG-guided prompting improves hazard detection in the visual-only setting by asking VLMs to first synthesize structured textual scene representations, revealing that safety logic is present but becomes accessible after explicit scene structuring by the models themselves.
\emph{$F_1$ scores are color-coded, with darker shading indicating higher performance.}}
\label{appendix:full_results}
\end{table*}

\section{Additional Details on \sagashort} \label{appendix:sec:gspo}
\subsection{Methodology}
For the $G$ sampled outputs, we compute the group-normalized advantage in Eq.~\ref{eq:group_advantage}:
\begin{equation}
\hat{A}^{(g)}
=
\frac{R(o^{(g)})-\mu}{\sigma+\epsilon},
\label{eq:group_advantage}
\end{equation}
where $\mu$ and $\sigma$ are computed over the $G$ rewards.

We use the sequence-level importance ratio in Eq.~\ref{eq:sequence_ratio}:
\begin{equation}
s^{(g)}(\theta)
=
\exp
\left(
\frac{1}{|o^{(g)}|}
\sum_{\tau=1}^{|o^{(g)}|}
\log
\frac{
\pi_\theta(o^{(g)}_{\tau} \mid x_i,x_t,o^{(g)}_{<\tau})
}{
\pi_{\mathrm{old}}(o^{(g)}_{\tau} \mid x_i,x_t,o^{(g)}_{<\tau})
}
\right).
\label{eq:sequence_ratio}
\end{equation}

Algorithm \autoref{alg:gspo_attention} shows the steps for \sagashort.

\begin{algorithm*}[t]
\caption{\saga}
\label{alg:gspo_attention}
\begin{algorithmic}[1]
\Require Policy model $\pi_\theta$; batch $\mathcal{B}$;attention layers $\mathcal{L}_{\mathrm{attn}}$
\Ensure Updated policy parameters $\theta$

\For{each training step}
    \For{each sample $(x_i^b,x_t^b,M^b,o_{gt}^b)\in\mathcal{B}$}
        \State Sample $G$ outputs $\{o_b^{(g)}\}_{g=1}^{G}\sim \pi_\theta(\cdot \mid x_i^b,x_t^b)$
        \State Compute reward $R_b^{(g)}$ 
        \State Compute group-normalized advantage $\hat{A}_b^{(g)}$
    \EndFor

    \State Compute $\mathcal{L}_{\mathrm{GSPO}}$ using Eq.~\ref{eq:gspo_loss}

    \State Initialize $\mathcal{L}_{\mathrm{attention}}\gets 0$ and $N\gets 0$

    \For{each attention layer $l\in\mathcal{L}_{\mathrm{attn}}$}
        \For{each sample $b\in\{1,\ldots,B\}$}
            \State Identify image-token positions $\mathcal{I}_{\mathrm{img}}^{b}$ and answer-token positions $\mathcal{I}_{\mathrm{ans}}^{b}$

            \If{$\mathcal{I}_{\mathrm{img}}^{b}\neq\emptyset$ and $\mathcal{I}_{\mathrm{ans}}^{b}\neq\emptyset$ and $M^b$ is valid}
                \State Extract answer-to-image attention $C^{(l,b)}$ using Eq.~\ref{eq:attention_submatrix}
                \State Compute spatial attention map $P^{(l,b)}$ using Eq.~\ref{eq:spatial_attention_map}
                \State Resize $M^b$ to the spatial grid of $P^{(l,b)}$ to obtain $M^{\prime b}$
                \State Accumulate the attention-alignment loss 
                \State $N\gets N+1$
            \EndIf
        \EndFor
    \EndFor

    \If{$N>0$}
        \State Average $\mathcal{L}_{\mathrm{attention}}$ over the $N$ valid layer-sample pairs
    \EndIf

    \State Compute total loss $\mathcal{L}$ 
    \State Update $\theta$ using AdamW 
\EndFor

\end{algorithmic}
\end{algorithm*}

\subsection{Dataset Split} \label{appendix:dataset_split} 
Our dataset contains 1207 image samples derived from 362 unique scenarios. Since multiple images can correspond to the same underlying scenario, a naive image-level split could introduce data leakage. To prevent this, we perform scenario-level splitting, ensuring that all images from a given scenario appear exclusively in one split. We further preserve subject-wise coverage across Biology, Chemistry, Cryogenic Liquids, General, and Physics so that each split contains diverse laboratory contexts. Using a 70/10/20 train/validation/test split , the final splits contain 841 training samples from 253 scenarios, 110 validation samples from 36 scenarios, and 256 test samples from 73 scenarios. The training set includes 307 hazardous and 534 non-hazardous samples, the validation set includes 51 hazardous and 59 non-hazardous samples, and the test set includes 93 hazardous and 163 non-hazardous samples.

\section{Additional Details on Hazard Detection Settings }\label{sec:hazard_detection_settings}

We evaluate VLMs under multiple hazard detection settings that differ in the form and amount of information provided to the model, from vision-only input to textual structured scene graph representation.

\textbf{Textual Scene Graphs with and without Hazard Attribute.}
In this text-only setting, hazard detection is performed using a textual representation of the scene graph (Eq.~\ref{eq:sg_def}), without access to any visual input. Formally, the model implements a structured prediction function as Eq.~\ref{eq:text-only}.
To isolate the contribution of explicit hazard attributes, we evaluate two variants of the textual scene graph. The first uses the full graph, in which each node is annotated with both \texttt{State} and \texttt{Hazard} attributes. The second employs a reduced graph $\mathcal{G}^{-H}$, obtained by removing the \texttt{Hazard} attribute from all nodes. In this hazard-implicit setting, the model is required to infer inherent risks solely from object states and inter-object relationships encoded in $\mathcal{G}^{-H}$. This controlled ablation allows us to assess whether VLMs can automatically recover inherent hazards (e.g., inferring that a specific chemical is flammable  from its name) from structured scene representations alone, without explicitly providing the associated potential hazard information.

\textbf{Visual + Textual Scene Graph.}
In this multimodal setting, the model receives both an image $I$ and the corresponding textual scene graph $\mathcal{G}$. The prediction is defined as Eq.~\ref{eq:vision+text} with the constraint that hazards must be supported by visual evidence in $I$. The scene graph is used as a disambiguating structure to clarify object identity, attributes, and relations visible in the image. This setting evaluates whether structured textual information can assist visual reasoning.

\textbf{Visual Input Only.}
In the image-only setting, the model receives only the laboratory image $I$ and performs hazard detection as Eq.~\ref{eq:vision-only}, where predictions are restricted to safety failures that are visible in the image. The model is explicitly prohibited from inferring hazards from general laboratory risks or hypothetical future actions. 


\textbf{Scene-Graph–Guided Prompting (SG-guided).}
In the SG-guided configuration, hazard detection is formulated as a two-stage process operating entirely on visual input. First, the model infers a scene graph $\mathcal{G}_I$ from the image as defined in Eq.~\ref{eq:visual_sg_group}. The resulting graph $\mathcal{G}_I = (\mathcal{V}, \mathcal{E})$ follows the formal definition of scene graphs given in Eq.~\ref{eq:sg_def}–Eq.~\ref{eq:sg_edge}. The generated graph contains only objects, attributes, and relationships that are visible in the image, with each node annotated using the fixed \texttt{State} and \texttt{Hazard} attributes. In the second stage, hazard prediction is performed over the generated scene graph, also defined in Eq.~\ref{eq:visual_sg_group}. This formulation ensures that hazard decisions are grounded in structural elements of $\mathcal{G}_I$, rather than being inferred directly from raw visual cues.
\section{Statistical Robustness}\label{appendix:sec:statistics}

We use default temperature settings in our main experiments, and it shows consistent trends across models. To assess further robustness, we report the performance of Qwen3-8B-4bit under varying temperatures and multiple runs (Table ~\ref{tab:qwen8b_4bit}). The results show minimal variance, indicating stability against sampling stochasticity, and are consistent with the findings in the main paper.

\begin{table}[H]
\centering

\renewcommand{\arraystretch}{1.5}
\resizebox{\linewidth}{!}{
\begin{tabular}{llccccccc}
\toprule
\textbf{Variation} & \textbf{Temp.} & \textbf{N Runs} & \textbf{Mean P} & \textbf{Std P} & \textbf{Mean R} & \textbf{Std R} & \textbf{Mean F1} & \textbf{Std F1} \\
\midrule
\multirow{3}{*}{TSG} 
  & 0.0 & 3 & 0.80 & 0.00 & 1.00 & 0.00 & \textbf{0.89} & 0.00 \\
  & 0.5 & 3 & 0.79 & 0.02 & 1.00 & 0.00 & 0.88 & 0.01 \\
  & 1.0 & 3 & 0.77 & 0.02 & 1.00 & 0.00 & 0.87 & 0.01 \\
\midrule
\multirow{3}{*}{V}   
  & 0.0 & 3 & 0.95 & 0.00 & 0.56 & 0.00 & 0.70 & 0.00 \\
  & 0.5 & 3 & 0.94 & 0.01 & 0.55 & 0.01 & 0.69 & 0.01 \\
  & 1.0 & 3 & 0.93 & 0.02 & 0.53 & 0.01 & 0.67 & 0.01 \\
\bottomrule
\end{tabular}}
\caption{Qwen3-8B-4bit Performance.}
\label{tab:qwen8b_4bit}
\end{table}

\section{VLM-as-judge}\label{appendix:sec:judge_alignment}

\begin{figure}
    \begin{center} \includegraphics[width=\linewidth]{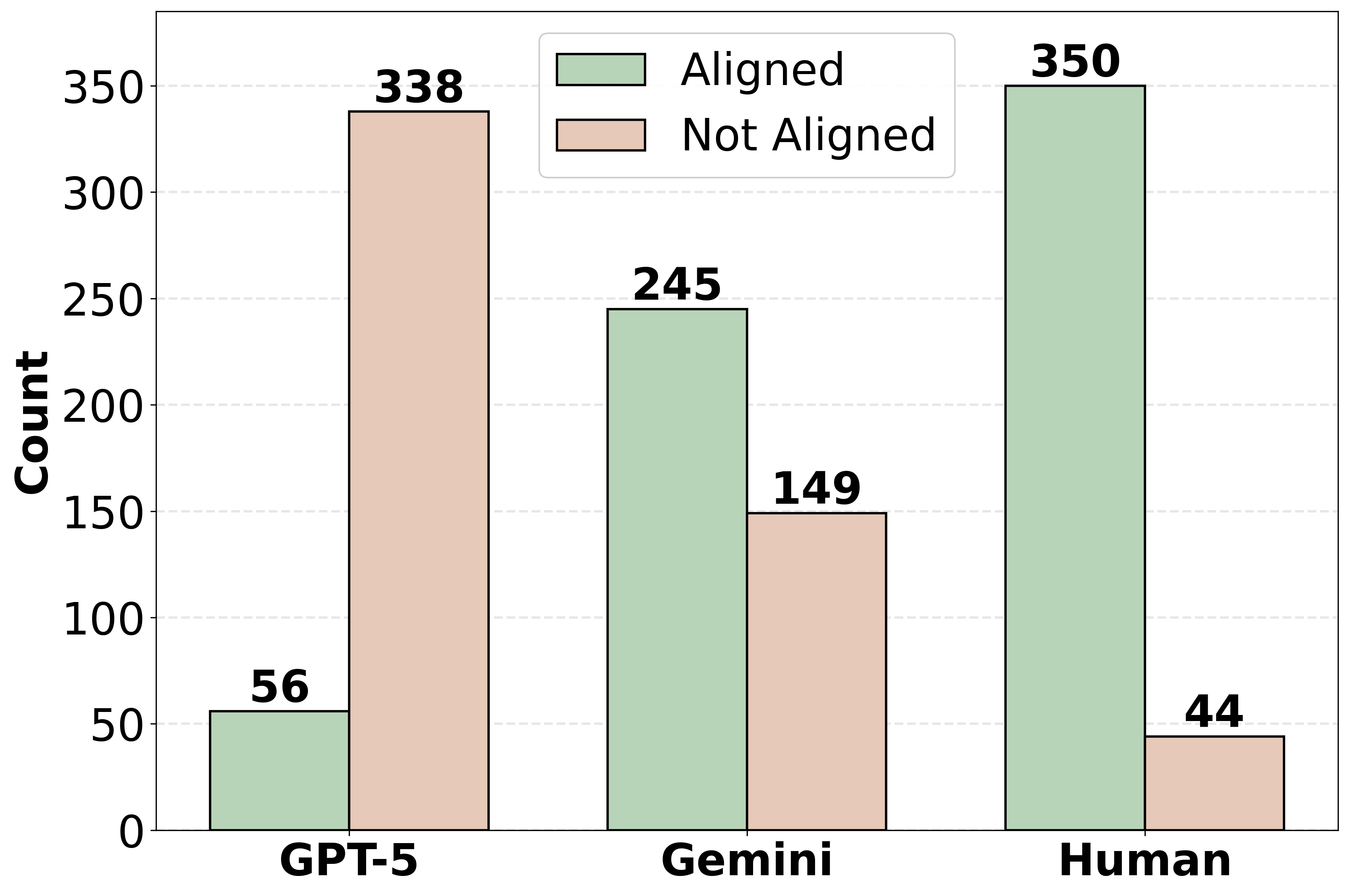}
    \end{center}
    \caption{Alignment outcomes when using VLMs as judges compared to human evaluation. Bars show the number of images classified as Aligned or Not Aligned with the corresponding scene graph and ground-truth hazards. Gemini exhibits substantially higher agreement with human judgments than  GPT-5.2-Pro, supporting its use as the VLM-as-judge for large-scale alignment filtering.}
    \label{fig:vlm-as-judge}
\end{figure}

\autoref{fig:vlm-as-judge} shows the alignment distribution of 394 scenarios, ground-truth, and corresponding images.

\section{Hyperparameters} \label{appendix:sec:hyperparameters}

\autoref{tab:hyperparameters} shows the hyperparameter list.

\renewcommand{\arraystretch}{1.5}
\begin{table}
\centering
\resizebox{\linewidth}{!}{  
\begin{tabular}{llc}
\toprule
\textbf{Category} & \textbf{Hyperparameter} & \textbf{Values} \\
\midrule
\midrule
\multirow{7}{*}{Optimizer}
  & Optimizer         & AdamW  \\
  & Learning rate     & $5 \times 10^{-6}$ \\
  & $\beta_1$         & 0.9 \\
  & $\beta_2$         & 0.99 \\
  & Weight decay      & 0.1\\
  & Warmup ratio      & 0.1 \\
  & LR scheduler      & Cosine \\
  & Gradient clip     & 0.1 \\
\midrule
\multirow{5}{*}{Training}
  & Batch size        & 2  \\
  & Grad. accum.\ steps & 4 \\
  & Rollouts $G$      & 8 \\
  & Max completion    & 2048 tokens \\
  & Importance sampling & Sequence-level \\
\midrule
\multirow{2}{*}{\shortstack[l]{Attention\\Guidance}}
  & Number of attention layers         & 4 (evenly spaced) \\
  & Image size                & $512 \times 512$ \\
  \midrule
  \multirow{2}{*}{Rewards}
  & Formatting $R_{\mathrm{format}}$ & $+1$ per correct tag (max $+2$) \\
  & Correctness $R_{\mathrm{correct}}$ & $+2$ if label matches, else $0$ \\
\bottomrule
\end{tabular}}
\caption{Hyperparameters for \sagashort.}
\label{tab:hyperparameters}
\end{table}
 
\section{Model Cards} \label{appendix:sec:model_cards}
\autoref{appendix:tab:model_cards} shows the model cards used in our 
experiments. For \saga{} training, all experiments were conducted on a single GPU, either an NVIDIA RTX A5000 or an NVIDIA RTX 6000 Ada, taking approximately 24 GPU-hours. We built on Unsloth, which provided memory-efficient training, combined with gradient checkpointing, to fit the training within GPU memory. All results are from a
single run per model; variance across sampling temperatures and
runs is reported for one model in Appendix ~\ref{appendix:sec:statistics}.

\begin{table}[H]

       \resizebox{\linewidth}{!}{  
\begin{tabular}{ccc}
\toprule
Model Name        & Complete Model ID                        & Hosting      \\ \midrule \midrule
llava-next-7B     & llava-hf/llava-v1.6-mistral-7b-hf        & Hugging Face \\
qwen3-8B          & Qwen/Qwen3-VL-8B-Instruct                & Hugging Face \\
llama-3.2-11B     & meta-llama/Llama-3.2-11B-Vision-Instruct & Hugging Face \\
 mistral-medium-3         & mistral-medium-3                         & GCP Vertex   \\
claude-sonnet-4.5 & claude-sonnet-4-5@20250929               & GCP Vertex   \\
gemini-3-flash    & gemini-3-flash-preview                   & GCP Vertex   \\
gemini-3-pro      & gemini-3-pro-preview                     & GCP Vertex  \\ 
qwen3.5-0.8B & unsloth/Qwen3.5-0.8B & Unsloth \\ 
qwen3.5-2B & unsloth/Qwen3.5-2B & Unsloth \\
gemma4-2B & unsloth/gemma-4-E2B-it & Unsloth \\
\bottomrule 
\end{tabular}}
\caption{Model cards used in our experiments}
\label{appendix:tab:model_cards}
\end{table}

\section{Human Annotation Guidelines}\label{app:annotation}
Annotators performed two tasks.
\textbf{(1) Ground-truth verification.} Given a scenario description, its
generated scene graph $\mathcal{G}$, and the extracted ground truth
$y=(\ell,k,\mathcal{H})$, annotators checked that $\mathcal{G}$ 
represents the description and that $\mathcal{H}$ lists exactly the hazards
that explicitly occur in it.
\textbf{(2) Image--scene alignment.} Given an image $I$, its scene graph
$\mathcal{G}$, and ground truth $y$, annotators labeled the triple
\texttt{ALIGNED} only if (i) all objects, attributes, and relations in
$\mathcal{G}$ are visibly supported in $I$, and (ii) the hazards visible in
$I$ exactly match $\mathcal{H}$; otherwise \texttt{NOT\_ALIGNED}.

\section{Prompts} \label{appendix:sec:prompts}
\begin{figure*}
\centering
\begin{tcolorbox}[title={Scene Graph Generation Prompt (Part 1)}]

You are an AI model whose core task is to generate a concise and accurate scene graph from a textual laboratory scenario, capturing all relevant objects, their attributes, and inter-object relationships for use by a text-to-image model in generating corresponding images.

A scene graph represents the scene as objects (nodes) with attributes and directed relationships (subject–predicate–object triplets) between them. Given a scenario description, your job is to identify only the key objects explicitly mentioned in the text and the most important relationships that are directly stated or unambiguously implied by the description. Do not infer, speculate, or add details that are not explicitly described. For each selected object, include an attributes dictionary with two keys: State and Hazard. State describes the object’s current physical properties and semantic conditions that influence how it behaves in the scene (e.g., broken, open, closed, tilted, transparent, nearby, in use, exposed, pressurized); Hazard describes any inherent or contextual risks associated with the object (e.g., flammable, toxic, reactive with air, UV-sensitive, breakable, spill risk, contamination risk, improper storage). If any attribute does not apply, use ``N/A''. Each relationship must be a directed triplet with ``subject'', ``predicate'', and ``object''.

Think step-by-step before writing the final output: first identify only the key objects essential to the scenario, then assign attributes to them using the fixed fields above, and finally construct subject–predicate–object triplets to represent the hazardous conditions described in the scenario.

The final output must be only the JSON scene graph, nothing else, in the following structure:
\begin{verbatim}
{
  "nodes": [
    {
      "object_name": "",
      "attributes": {
        "State": "",
        "Hazard": ""
      }
    }
  ],
  "relationships": [
    {
      "predicate": "",
      "subject": "",
      "object": ""
    }
  ]
}
\end{verbatim}

\textbf{Example:}

\emph{Input Scenario:}

``A chemical laboratory stores a bottle of diethyl ether in a transparent glass container. This container is mistakenly left on an open shelf near a window that receives direct sunlight during the day. While diethyl ether is stored in its original container, its placement near the window exposes it to both ultraviolet (UV) light and oxygen from the air whenever the container is opened for usage. The laboratory aims to maintain an organized environment with proper waste disposal facilities, but the current storage arrangement of diethyl ether raises concerns about proper safety practices.''

\end{tcolorbox}
\caption{Prompt used for scene graph generation from textual laboratory safety scenarios (Part 1).}
\label{fig:scene-graph-generation-prompt-part1}
\end{figure*}

\begin{figure*}
\centering
\begin{tcolorbox}[title={Scene Graph Generation Prompt (Part 2)}]

\emph{Expected Output:}
\begin{verbatim}
{
  "nodes": [
    {
      "object_name": "diethyl ether",
      "attributes": {
        "State": "stored in original container",
        "Hazard": "flammable, peroxide-forming, UV-sensitive"
      }
    },
    {
      "object_name": "glass container",
      "attributes": {
        "State": "transparent",
        "Hazard": "breakable"
      }
    },
    {
      "object_name": "open shelf",
      "attributes": {
        "State": "open",
        "Hazard": "N/A"
      }
    },
    {
      "object_name": "window",
      "attributes": {
        "State": "receives direct sunlight",
        "Hazard": "N/A"
      }
    },
    {
      "object_name": "sunlight",
      "attributes": {
        "State": "direct",
        "Hazard": "UV exposure"
      }
    },
    {
      "object_name": "air",
      "attributes": {
        "State": "present when container is opened",
        "Hazard": "oxidation risk"
      }
    }
  ]
}
\end{verbatim}

\end{tcolorbox}
\caption{Prompt used for scene graph generation from textual laboratory safety scenarios (Part 2).}
\label{fig:scene-graph-generation-prompt-part2}
\end{figure*}

\begin{figure*}
\centering
\begin{tcolorbox}[title={Scene Graph Generation Prompt (Part 3)}]

\emph{Expected Output (continued):}
\begin{verbatim}
{
  "relationships": [
    {
      "subject": "diethyl ether",
      "predicate": "stored_in",
      "object": "glass container"
    },
    {
      "subject": "glass container",
      "predicate": "placed_on",
      "object": "open shelf"
    },
    {
      "subject": "open shelf",
      "predicate": "near",
      "object": "window"
    },
    {
      "subject": "window",
      "predicate": "exposes_to",
      "object": "sunlight"
    },
    {
      "subject": "diethyl ether",
      "predicate": "exposed_to",
      "object": "sunlight"
    },
    {
      "subject": "diethyl ether",
      "predicate": "exposed_to",
      "object": "air"
    }
  ]
}
\end{verbatim}

\end{tcolorbox}
\caption{Prompt used for scene graph generation from textual laboratory safety scenarios (Part 3).}
\label{fig:scene-graph-generation-prompt-part3}
\end{figure*}

\begin{figure*}
\centering
\begin{tcolorbox}[title={Ground-Truth Hazard Classification Prompt }]
You are a laboratory safety scenario classification agent. Given the following input fields — subject, scenario\_description, lab\_safety\_related\_issues, and lab\_safety\_topic — determine whether the described laboratory scenario is hazardous or non-hazardous. Focus only on the events explicitly described in the scenario\_description when making your decision.

Classify the scenario as ``hazardous'' only if the description clearly indicates a definite safety failure, incident, or protocol violation (for example, torn gloves, chemical spill, missing PPE, or improper equipment handling). If the description shows that all safety measures are properly followed, classify it as ``non-hazardous.'' Use the lab\_safety\_related\_issues only for contextual understanding of potential risks, not as evidence of actual hazards. Some issues in that field may describe what could go wrong, but you must only classify hazards that definitely occur in the scenario\_description. Your output must include three fields in JSON format:
\begin{verbatim}
{"classification": "hazardous" or "non-hazardous",
  "hazards_count": <integer number of definite hazards found>,
  "existing_hazards": ["concise hazard name(s)" or [] if none]}
\end{verbatim}

If no hazards are confirmed, return:
\begin{verbatim}
{"classification": "non-hazardous",
  "hazards_count": 0,
  "existing_hazards": []}
\end{verbatim}

\textbf{Example 1:}

\emph{Input:}  
subject: biology  

Scenario\_description: During a microbiology experiment in a Level 2 biosafety laboratory, a researcher is transferring biohazardous bacterial cultures from a sealed container to growth plates. This procedure involves the use of PPE, including nitrile gloves, a lab coat, safety goggles, and a surgical mask. While placing a growth plate on the work surface, the researcher accidentally tears the glove on a sharp edge of a metallic instrument. Biohazard waste bins are available nearby for proper disposal of contaminated materials. A lab supervisor is present to oversee safety.

lab\_safety\_related\_issues: \{\ldots\}  

Lab\_safety\_topic: gloves  

\emph{Output:}
\begin{verbatim}
{ "classification": "hazardous",
  "hazards_count": 1,
  "existing_hazards": ["torn gloves"]}
\end{verbatim}

\vspace{0.5em}
\textbf{Example 2:}

\emph{Input:}  
subject: biology  

Scenario\_description: In a microbiology laboratory, a technician is preparing to centrifuge samples of biological fluids known to potentially contain infectious agents. The centrifuge is equipped with sealed rotors and safety cups designed to prevent the release of aerosols during operation. Biohazard waste containers are present for proper disposal of contaminated materials, and the technician is wearing appropriate personal protective equipment including a lab coat, gloves, and safety goggles.

lab\_safety\_related\_issues: \{\ldots\}  

Lab\_safety\_topic: centrifuge  

\emph{Output:}
\begin{verbatim}
{ "classification": "non-hazardous",
  "hazards_count": 0,
  "existing_hazards": []}
\end{verbatim}

\end{tcolorbox}
\caption{Prompt used to extract ground-truth hazard labels from textual laboratory scenarios }
\label{fig:gt-extraction-prompt-part1}
\end{figure*}

\begin{figure*}
\centering
\begin{tcolorbox}[title={Graph-Conditioned Image Generation Prompt (Hazardous)}]

Act as a laboratory safety monitoring agent and generate a realistic, full-color, surveillance-style image of a \texttt{\{subject\}} laboratory.

The image must be a single frame as if captured from a fixed, elevated ceiling viewpoint similar to a CCTV feed, but DO NOT depict the camera itself. The ceiling and walls must be free of any recording devices.

The laboratory should appear clean and professional while clearly depicting the safety hazard described in the scene graph below. Use the provided scene graph as the authoritative reference for all objects, attributes, and relationships, ensuring that their placement, materials, states, and hazardous conditions are visually consistent with the scene graph.

Depict the hazard only through natural physical or visual cues—such as object states, positions, and interactions—without overlays, highlights, or artificial annotations.

\textbf{Scene Graph:}
\begin{verbatim}
{scene_graph}
\end{verbatim}

\end{tcolorbox}
\caption{Prompt used for graph-conditioned image generation of hazardous laboratory scenarios.}
\label{fig:graph-conditioned-image-hazardous}
\end{figure*}

\begin{figure*}
\centering
\begin{tcolorbox}[title={Graph-Conditioned Image Generation Prompt (Non-Hazardous)}]

Act as a laboratory safety monitoring agent and generate a realistic, full-color, surveillance-style image of a \texttt{\{subject\}} laboratory.

The image must be a single frame as if captured from a fixed, elevated ceiling viewpoint similar to a CCTV feed, but DO NOT depict the camera itself. The ceiling and walls must be free of any recording devices.

The laboratory should appear clean, organized, and professional, accurately representing a safe and compliant working environment. Use the provided scene graph as the authoritative reference for all objects, attributes, and relationships, ensuring that their placement, materials,and states are visually consistent with the scene graph.

Depict the scene only through natural physical or visual cues—such as object states, positions, and interactions—without overlays, highlights, or artificial annotations.

\textbf{Scene Graph:}
\begin{verbatim}
{scene_graph}
\end{verbatim}

\end{tcolorbox}
\caption{Prompt used for graph-conditioned image generation of non-hazardous laboratory scenarios.}
\label{fig:graph-conditioned-image-nonhazardous}
\end{figure*}

\begin{figure*}
\centering
\begin{tcolorbox}[title={Hazard Detection Prompt (Image Only)}]

You are a laboratory safety monitoring agent. Analyze the provided \texttt{\{subject\}} laboratory image and determine whether the scene is hazardous or non-hazardous. Base your judgment only on what is directly and clearly visible in the image, and do not infer intentions, future actions, or hypothetical failures. Do not assume hazards based on missing information or general laboratory risks. A hazard must be a definite, observable safety failure or protocol violation that is visibly present in the image, such as a chemical spill, missing required personal protective equipment, exposed flames near flammable materials, or damaged laboratory equipment. Classify the scene as "hazardous" only if at least one such definite and visible safety hazard is present; otherwise, classify the scene as "non-hazardous" if no clear safety violations are visible.

Your output must be valid JSON and include exactly the following fields:

\begin{verbatim}
{
  "classification": "hazardous" or "non-hazardous",
  "hazards_count": <integer number of definite hazards found>,
  "existing_hazards": ["concise hazard name(s)"] or []
}
\end{verbatim}

If no hazards are confirmed, return exactly:

\begin{verbatim}
{
  "classification": "non-hazardous",
  "hazards_count": 0,
  "existing_hazards": []
}
\end{verbatim}
Do not include explanations, evidence descriptions, confidence scores, numbering, or any additional text.

\end{tcolorbox}
\caption{Prompt used for hazard detection from images only.}
\label{fig:prompt-image-only}
\end{figure*}

\begin{figure*}
\centering
\begin{tcolorbox}[title={Hazard Detection Prompt (Textual Scene Graph Only)}]

You are a laboratory safety monitoring agent. Analyze the provided \texttt{\{subject\}} laboratory scene described solely in the below attached textual scene graph and determine whether the scene is hazardous or non-hazardous. Base your judgment only on safety-relevant conditions that are explicitly and unambiguously stated in the scene graph. Do not infer intentions, future actions, or hypothetical failures. Do not assume hazards based on missing information, general laboratory risks, or domain knowledge beyond what is explicitly described in the scene graph. A hazard must be a definite and explicitly stated safety failure or protocol violation present in the scene graph, such as a chemical spill, missing required personal protective equipment, exposed flames near flammable materials, or damaged laboratory equipment. Classify the scene as "hazardous" only if at least one such definite hazard is explicitly described; otherwise, classify the scene as "non-hazardous" if no clear safety violations are stated.

Your output must be valid JSON and include exactly the following fields:

\begin{verbatim}
{
  "classification": "hazardous" or "non-hazardous",
  "hazards_count": <integer number of definite hazards found>,
  "existing_hazards": ["concise hazard name(s)"] or []
}
\end{verbatim}

If no hazards are confirmed, return exactly:

\begin{verbatim}
{
  "classification": "non-hazardous",
  "hazards_count": 0,
  "existing_hazards": []
}
\end{verbatim}
Do not include explanations, evidence descriptions, confidence scores, numbering, or any additional text.

Scene Graph:
\begin{verbatim}
{scene_graph}
\end{verbatim}

\end{tcolorbox}
\caption{Prompt used for hazard detection from textual scene graphs only.}
\label{fig:prompt-text-only}
\end{figure*}

\begin{figure*}
\centering
\begin{tcolorbox}[title={Hazard Detection Prompt (Image + Textual Scene Graph)}]

You are a laboratory safety monitoring agent. Analyze the provided \texttt{\{subject\}} laboratory image together with the below attached textual scene graph and determine whether the scene is hazardous or non-hazardous. Base your judgment only on safety-relevant conditions that are directly and clearly supported by the image, using the textual scene graph solely to clarify or confirm what is visually present. Do not infer intentions, future actions, or hypothetical failures. Do not assume hazards based on missing information, general laboratory risks, or scene-graph elements that are not visibly supported by the image. A hazard must be a definite, observable safety failure or protocol violation that is visibly present in the image, such as a chemical spill, missing required personal protective equipment, exposed flames near flammable materials, or damaged laboratory equipment. Classify the scene as "hazardous" only if at least one such definite and visible safety hazard is present; otherwise, classify the scene as "non-hazardous" if no clear safety violations are visible.

Your output must be valid JSON and include exactly the following fields:

\begin{verbatim}
{
  "classification": "hazardous" or "non-hazardous",
  "hazards_count": <integer number of definite hazards found>,
  "existing_hazards": ["concise hazard name(s)"] or []
}
\end{verbatim}

If no hazards are confirmed, return exactly:

\begin{verbatim}
{
  "classification": "non-hazardous",
  "hazards_count": 0,
  "existing_hazards": []
}
\end{verbatim}
Do not include explanations, evidence descriptions, confidence scores, numbering, or any additional text.

Scene Graph:
\begin{verbatim}
{scene_graph}
\end{verbatim}

\end{tcolorbox}
\caption{Prompt used for hazard detection from images with textual scene graph context.}
\label{fig:prompt-image-text}
\end{figure*}

\begin{figure*}
\centering
\begin{tcolorbox}[title={Hazard Detection Prompt (Scene Graph Guided, Part 1/2)}]

You are a laboratory safety monitoring agent that performs hazard identification in two strictly ordered stages:  (1) Scene Graph Generation  (2) Hazard Classification  

You must follow these stages internally in order, but output ONLY the final JSON object described below.  Do not include explanations, reasoning steps, or any text outside the final JSON.

STAGE 1: SCENE GRAPH GENERATION

Analyze the provided laboratory image and generate a concise and accurate scene graph based only on what is directly and clearly visible.

A scene graph represents the scene as objects (nodes) with attributes and directed relationships (subject–predicate–object triplets) between them. Do not infer, speculate, or add details that are not explicitly described. For each selected object, include an attributes dictionary with two keys: State and Hazard. State describes the object's current physical properties and semantic conditions that influence how it behaves in the scene (e.g., broken, open, closed, tilted, transparent, nearby, in use, exposed, pressurized); Hazard describes any inherent or contextual risks associated with the object (e.g., flammable, toxic, reactive with air, UV-sensitive, breakable, spill risk, contamination risk, improper storage). If any attribute does not apply, use "N/A". Each relationship must be a directed triplet with "subject", "predicate", and "object".

Think step-by-step before writing the final output: first identify only the key objects essential to the scenario, then assign attributes to them using the fixed fields above, and finally construct subject–predicate–object triplets to represent the conditions described in the scenario.

\end{tcolorbox}
\caption{Prompt used for scene graph guided hazard detection (generate a scene graph, then classify hazards), Part 1/2.}
\label{fig:prompt-sg-guided-part1}
\end{figure*}

\begin{figure*}
\centering
\begin{tcolorbox}[title={Hazard Detection Prompt (Scene Graph Guided, Part 2/2)}]

STAGE 2: HAZARD CLASSIFICATION

Analyze the \texttt{\{subject\}} laboratory scene using only the scene graph you generated in Stage 1 and determine whether the scene is hazardous or non-hazardous. Base your judgment only on safety-relevant conditions that are explicitly and unambiguously stated in the scene graph. Do not infer intentions, future actions, or hypothetical failures. Do not assume hazards based on missing information, general laboratory risks, or domain knowledge beyond what is explicitly described in the scene graph. A hazard must be a definite and explicitly stated safety failure or protocol violation present in the scene graph, such as a chemical spill, missing required personal protective equipment, exposed flames near flammable materials, or damaged laboratory equipment. Classify the scene as "hazardous" only if at least one such definite hazard is explicitly described; otherwise, classify the scene as "non-hazardous" if no clear safety violations are stated.

FINAL OUTPUT FORMAT (JSON ONLY):

Return a single valid JSON object with exactly the following structure:

\begin{verbatim}
{
  "scene_graph": {
    "nodes": [
      {
        "object_name": "",
        "attributes": {
          "State": "",
          "Hazard": ""
        }
      }
    ],
    "relationships": [
      {
        "subject": "",
        "predicate": "",
        "object": ""
      }
    ]
  },
  "hazard_assessment": {
    "classification": "hazardous" or "non-hazardous",
    "hazards_count": <integer>,
    "existing_hazards": ["concise hazard name(s)"] or []
  }
}
\end{verbatim}

If no definite hazards are confirmed, return:

\begin{verbatim}
{
  "scene_graph": { ... },
  "hazard_assessment": {
    "classification": "non-hazardous",
    "hazards_count": 0,
    "existing_hazards": []
  }
}
\end{verbatim}

Do not include explanations, evidence descriptions, confidence scores, numbering, or any additional text.

\end{tcolorbox}
\caption{Prompt used for scene graph guided hazard detection (generate a scene graph, then classify hazards), Part 2/2.}
\label{fig:prompt-sg-guided-part2}
\end{figure*}

\begin{figure*}
\centering
\begin{tcolorbox}[title={Mask Generation Prompt}]

For hazardous samples with annotated ground-truth hazards, we use the following prompt:

\begin{verbatim}
Generate mask for the {existing_hazards}, and {mask_topic}.
Output a binary segmentation mask image where the masked region is white
and the background is black. Do not add any text, labels, or annotations.
\end{verbatim}

For non-hazardous samples, we use:

\begin{verbatim}
Generate mask for the {mask_topic}.
Output a binary segmentation mask image where the masked region is white
and the background is black. Do not add any text, labels, or annotations.
\end{verbatim}

Here, \texttt{\{existing\_hazards\}} denotes the annotated ground-truth hazard names, and \texttt{\{mask\_topic\}} denotes the corresponding LabSafetyBench scenario topic.

\end{tcolorbox}
\caption{Prompt used to generate binary hazard relevance masks.}
\label{fig:prompt-mask-generation}
\end{figure*}



\end{document}